\documentclass{article}
\usepackage[a4paper,left=1in,top=1in,right=1in,bottom=1in,nohead]{geometry}
\usepackage[english]{babel}
\usepackage{lmodern}
\usepackage[T1]{fontenc}  
\usepackage[utf8]{inputenc}
\usepackage{amsmath}
\usepackage{amssymb}
\usepackage{amsfonts}
\usepackage{physics}
\usepackage{mathtools}
\usepackage{graphicx}
\usepackage{url}
\usepackage[export]{adjustbox}
\usepackage{xcolor}
\usepackage[numbers]{natbib}
\usepackage[pdftex, pdftitle={Article}, pdfauthor={Author}]{hyperref}

\title{Category learning in deep neural networks: Information content and geometry of internal representations}
\author{Laurent Bonnasse-Gahot$^{1}$ and Jean-Pierre Nadal$^{1,2,*}$}
\date{
\normalsize
(1) Centre d'Analyse et de Math\'ematique Sociales (CAMS)\\
EHESS, CNRS \\
\'Ecole des Hautes \'Etudes en Sciences Sociales \\
54 bd. Raspail, 75006 Paris, France \\
(2) Laboratoire de Physique de l’\'Ecole Normale Sup\'erieure (LPENS),\\ ENS, Universit\'e PSL, CNRS, Sorbonne Universit\'e, Université Paris Cit\'e\\
\'Ecole Normale Sup\'erieure \\
24 rue Lhomond,  75005 Paris, France \\
$\,$\\
(*) Corresponding author (\texttt{jean-pierre.nadal@phys.ens.fr})\\
$\,$\\
This is an author-produced version of an article published in \textit{Physical Review E}\\
\url{https://doi.org/10.1103/mp35-bdx5}
}

\def\beq {\begin{equation}}
\def\eeq {\end{equation}}
\def\beqa {\begin{eqnarray}}
\def\eeqa {\end{eqnarray}}

\usepackage{comment}

\begin{document}
\maketitle
\thispagestyle{empty}

\noindent\rule{\textwidth}{0.7pt}
\begin{abstract}
In humans and other animals, category learning is associated with a better ability to discriminate between stimuli that are close to the category boundary, compared with stimuli well within a category. This perceptual within-category compression and between-category separation, called categorical perception, was also empirically observed in artificial neural networks trained on classification tasks. In previous modeling works based on empirical neuroscience data, we took an information-theoretic approach that shows that this expansion-compression is a necessary outcome of efficient learning. As a result, the impact of input or neuronal noise is reduced where it is most detrimental, namely at the boundary between categories.\\
Here we extend our theoretical framework to artificial feedforward networks. The Bayes cost that we consider is an average over the data distribution of the standard cross-entropy loss function. We show that minimizing this cost implies maximizing the mutual information between the set of categories and the neural activities prior to the decision layer. We then consider structured data, formalized by the assumption of an underlying feature space of small dimension. We show that, for wide networks, and more generally in situations of high signal-to-noise ratio, maximizing the mutual information implies (i) finding an appropriate projection space, and, (ii) building a neural representation with the appropriate metric. The latter is based on a Fisher information matrix measuring the sensitivity of the neural activity to changes in the projection space. Optimal learning makes this neural Fisher information follow a category-specific Fisher information, measuring the sensitivity of the category membership to changes in the projection space. One consequence is that category learning induces the main neural correlate of categorical perception, an expansion of neural space near decision boundaries.\\
To make this statement more precise we characterize the properties of the categorical Fisher information. We show that its eigenvectors give the most discriminant directions at each point of the projection space. We find that, unexpectedly, its maxima are in general not exactly at, but near, the class boundaries. Considering toy models and the MNIST handwritten digits dataset, we numerically illustrate how after learning the two Fisher information matrices match, and essentially align with the boundaries between categories. Finally, we provide a variety of supplemental analyses, in particular we relate our approach to that of the Information Bottleneck, and we exhibit a bias-variance decomposition of the Bayes cost, which is of interest on its own.\\

\end{abstract}
\noindent\rule{\textwidth}{0.8pt}

\maketitle

\clearpage
\tableofcontents 
\clearpage

\section{Introduction} 
\label{sec:intro} 

In neuropsychology, a large amount of experimental works has been done on the study of perceptual decision making with stimuli whose level of ambiguity is (one of) the control parameter(s) -- see, e.g., Refs. \citep{Liberman_etal_1957,cross1965identification,nelson1989categorical,kuhl1991human,Beale_Keil_1995,iverson1996influences,freedman2001categorical,GoldShadlen2001decisionmaking,gold2007neuralbasis,caves2018categorical,okazawa2021representational}. A main and general qualitative outcome of these experiments is to exhibit common behavioral properties leading to the notion of a Categorical Perception (CP) phenomenon~\citep{harnad1987psychophysical}. The psychophysics of CP is characterized first by a sharp transition between classes, where the categorical identity changes abruptly near the boundary -- the continuous inputs are clearly segregated in discrete outputs. Second, this categorization comes with a within-category compression and a between-category separation. That is, two stimuli, close in input space, are perceived closer if they belong to a same category than if they belong to different categories. Near the boundary between categories (where the categorical ambiguity is the greatest), discriminability $d'$ and reaction times are greater~\citep{Harnad_1987}. Various works have addressed the issue of explaining the CP phenomenon (see, e.g., \citep{anderson1977distinctive,padgett1998simple,Damper_Harnad_2000,Guenther_Bohland_2002}). In our previous works~\citep{LBG_JPN_2008,LBG_JPN_2012} we show that CP is a necessary outcome of efficient coding of the stimuli when the goal is to optimize the identification of the associated categories (and not to efficiently encode the stimuli themselves). Our analysis implies that the neural correlates of CP are characterized by a neural geometry in which the space is expanded near class boundaries and contracted away from the boundaries. The few neurophysiological experiments which give some hints on the neural geometry confirm these predictions: see in particular Ref. \citep{Koida_Komatsu_2007}, Figs. 2 and 6, showing that more neural resources are allocated to the class boundary, and Ref. \citep{okazawa2021representational}, in which an analysis of the activity of a pool of recorded neurons exhibits the expansion/compression effect, as illustrated in their Fig. 5, panel H.\\
In contrast, in machine learning, for categorization tasks, the focus is mainly on finding a decision boundary. Authors have addressed the issue of finding the best possible margin for linear separation by a perceptron or with kernel methods (see, e.g., Refs. \citep{boser1992optimalmargin,scholkopf2018learning,AmariWu1999improving}). However, few works consider issues specifically related to the data (stimuli) ambiguity. Yet, previous computational work has shown that the perceptual warping typical of categorical perception also happens in artificial neural networks \citep{harnad1991categorical,tijsseling1997warping,theriault2018learning,LBG_JPN_CPDeepLearning_2022,zavatone2023neural}. With protocols inspired by typical cognitive experiments, in Ref.~\citep{LBG_JPN_CPDeepLearning_2022} we show with extensive numerical experiments that the neural geometry, from layer to layer, gradually acquires the characteristics of the geometry underlying CP, where space is magnified near category boundaries. These numerical studies confirm the expectations from our theoretical analysis of CP in previous computational neuroscience works~\citep{LBG_JPN_2008,LBG_JPN_2012}. In these studies of categorization tasks, we adopt a Bayesian and information-theoretic viewpoint that is at the basis of many works in the context of the modeling of sensory coding in the brain (see, e.g.,  Ref.~\cite{DayanAbbott}). In line with these previous works, here we show that, for analyzing artificial multilayer neural networks, one can actually adopt, and adapt, the Bayesian viewpoint we considered in the neuroscience context for the modeling of the neural basis of categorical perception in human or other animals.

In the neuroscience context, modeling takes advantage of empirical results on the type of neural architectures, neural codes and decision dynamics which are found ubiquitous in perceptual decision making. In particular, a variety of empirical results reveal an encoding layer with a distributed representation of stimulus-specific cells -- coding, e.g., for orientation, or movement, or more complex features-- followed by a decoding (decision) layer with cells specific to each one of the possible categories (see, e.g.,  Refs.~\citep{freedman2001categorical,kreiman2000category,GoldShadlen2001decisionmaking}). In the modeling of categorical perception, the feature space is thus considered as known, and the processing leading to this encoding feature space is not considered. The modeling of categorical perception then amounts to considering a neural architecture with, on one hand, the feature space of small dimension, essentially identified with the stimulus space, and on the other hand, the neural representation, the neural activity giving, in a distributed and generally noisy way, the localization in this space. The readout is obtained by a decoding (or decision) layer with category-specific cells. It is for such an architecture that we obtained analytical results.\\
In the context of machine learning, one has to deal with a high-dimensional input and the learning of the multilayer processing able to produce a neural representation that can be linearly decoded. Typical layers have a large number of neurons, and the possible existence of an underlying feature space of small dimension is not necessarily discussed. In the present paper, by formalizing the notion of underlying feature spaces, we further extend our Bayesian approach to the case of feature spaces of small dimension for each layer. This allows us to adapt to the machine learning context the analytical results we obtained in the neuroscience context. We also derive new ones as briefly described below. We analyze the outcomes of our analysis in terms of geometry of the neural representations. In doing so, we provide a better theoretical understanding of the numerical results previously obtained in Ref.~\cite{LBG_JPN_CPDeepLearning_2022}, that is more generally of CP in shallow and deep networks. We also perform additional numerical experiments illustrating the new theoretical results. It is important to notice that we study properties based on the adaptation of the network to the stimulus (data) {\em distribution}, and not as the result of learning from a finite sample of examples. This might seem as taking a step backward from the core goal of machine learning. However, as we will see, this is what allows us to characterize the geometry of internal representations of a (natural or artificial) neural network which has learned a categorization task.\\

The organization of the paper is as follows. 
First, in Sec. \ref{sec:BayesToInfomax}, we extend the Bayesian formalism introduced in Ref.~\cite{LBG_JPN_2012} to the case of multilayer networks, allowing us to cast our approach and results within the machine learning framework. We consider multilayer feedforward networks for which the goal is to learn a categorization task. We do not specify the type of neural activation functions. We develop a statistical approach: data (stimuli) are characterized by probability distributions, and their category membership is also a random variable. For the network, we consider weakly noisy neural activities. Technically, the statistical framework and the neural stochasticity allow us to make use of Bayesian and information-theoretic formalisms and tools. More fundamentally, as stressed in Refs.~ \cite{tishby99information,schwartz-ziv2017opening}, neural noise plays an important role by revealing the data and network complexity. In addition, we note that standard regularization techniques in machine learning, such as the dropout one \cite{bouthillier2015dropout}, consist in adding neural noise during learning. 

Within this statistical framework, we introduce the mean Bayes risk (expressed in terms of a Kullback-Leibler divergence) adapted to a categorization task. We show that the minimization of this cost amounts to dealing with two issues: optimizing the decision stage in order to provide the best possible estimator of the category given the neural activities; and optimizing the stimulus encoding (through the multilayer processing) by maximizing the mutual information between categories and neural code. We discuss the links and differences with the information bottleneck approach~\citep{tishby99information}. 

Next, in Sec. \ref{sec:featurespaces}, we formalize the hypothesis that structured data lives in a manifold of small dimension as compared with the data (network input) dimension. Through the feedforward processing, the network transforms this underlying manifold into a new version which, through learning, will be adapted to the task. We show how the mean Bayes cost can be re-written in terms of these underlying manifolds. We also discuss a bias-variance type decomposition of the Bayes cost, of interest on its own. In addition, in the appendix, we derive bounds on the cost based on this decomposition.

In Sec. \ref{sec:geometry} we characterize the mutual information between the discrete categories and the neural code in a regime of high signal-to-noise ratio (focusing mainly on the limit of wide networks, that is for a large number of neurons in the considered layer). The analysis is an extension of the main result in Ref.~\citep{LBG_JPN_2008}. This particular asymptotic limit allows us to reveal the neural metrics relevant for the categorization task. It shows that maximizing the mutual information leads to finding the feature space most relevant for the classification (and amenable to easy decoding), and to probe this space with a particular metric: the space should be expanded near a class boundary, and contracted far from a boundary. This implies a better ability to discriminate between nearby inputs in the vicinity of a class boundary, than far from such boundary, that is, the categorical perception effect.

Formally, the maximization of the mutual information implies the matching of two Fisher information matrices. One, that we refer to as the categorical Fisher information, characterizes the sensitivity of the probability of the class (considered as ``responsible'' for the occurrence of the stimulus), to small displacements in the feature space. Along a path in feature space which goes from an item of one category to an item of another category, this categorical Fisher information will be the largest near the class boundary. The other Fisher information, that we refer to as the neural Fisher information, characterizes the sensitivity of the neural representation to small displacements in the underlying feature space. This Fisher information is the one usually encountered in neuroscience, related to the behavioral discriminability measured in experiments \citep{DayanAbbott,seung1993simple}. Matching of the neural Fisher information with the categorical Fisher information thus leads to the categorical perception effect mentioned above.

To better understand the consequences of the maximization of the mutual information, and of the matching between the two Fisher information matrices, in Sec. \ref{sec:Fcat} we then characterize the categorical Fisher information -- an analysis not done in our previous works, except for the scalar case, that is for a one-dimensional (1D) feature space. We show that the eigenvectors of this matrix give, {\em at each point} of the feature space, the most relevant discriminant directions, which we call the {\em principal discriminant directions}. We provide numerical illustrations of our results for the simple case of Gaussian categories. We also study the location of the maxima of the categorical Fisher information. One might expect that the maximum is reached exactly when crossing the category boundary. This is the case for distributions with the same (co)variance matrices. However, we show here that otherwise the location of the maxima of the categorical Fisher information is actually displaced away from the class boundary. Characterizing this displacement, we show that it is typically small, so that the qualitative conclusions concerning the categorical perception effect remain valid. 

In Sec. \ref{sec:illustrations}, we provide numerical illustrations with multilayer networks trained on either Gaussian data or on the MNIST database of handwritten digits. We go beyond the numerical analysis we did in the related work, Ref.~\cite{LBG_JPN_CPDeepLearning_2022}, making here precise links with the new analyses of the present paper. In particular, in the simplest cases, we represent the categorical Fisher information, and the matching between the categorical and neural Fisher information matrices. We also validate the use in Ref.~\cite{LBG_JPN_CPDeepLearning_2022} of a proxy for the Fisher information, which cannot be easily computed in deep networks.

Finally, Sec. \ref{sec:Discussion} we discuss the significance of the results. We give details and supplementary information in a set of appendixes.

\section{Category learning: from Bayes to Infomax}
\label{sec:BayesToInfomax}
This Sec. extends to multilayer networks the approach we introduced in Ref.~\citep{LBG_JPN_2012}. The formulation given here, although very close to the one in this previous work, makes explicit the decoupling between coding and decoding tasks that results from the analysis of the Bayes cost function. This leads to the infomax criterion for the coding part, and the optimality of having the output estimating the probability of the category given the neural activity for the decoding part. 

This Sec. is quite general, there is no hypothesis on the data structure, apart from the fact that they belong to a finite set of categories. In the following Sec. \ref{sec:featurespaces}, we consider structured data. Given the dual context of neuroscience and data science, in all this paper we interchangeably make use of the terms ``stimulus'' and ``input data'' (or simply ``data'' when there is no ambiguity). 

\subsection{General framework}
\label{sec:general_framework}

\subsubsection{Sensory or data space} 
To model the input data, we assume given a discrete set of categories (classes), $y=1,\ldots,M $ with probabilities of occurrence $P_{y} \geq 0$, so that $\textstyle{\sum}_{y} P_{y} = 1$. Each category is characterized by a density distribution $P(\mathbf{s}|y)$ over the input (sensory or data) space.

We assume that every probability density function (pdf) is as regular as needed. If the support of the pdf of the stimuli is not connected, the categorization task decomposes into independent categorization tasks associated with each one of the connected components. Hence, without loss of generality, we can assume that the support of the pdf of the stimuli is connected. Since the focus of the paper is on the neural geometry induced by the categorization of possibly ambiguous stimuli, we assume that the supports of the pdf of the stimuli given the categories are not disjoint.

\subsubsection{Feedforward network}
We consider a multilayer feedforward (shallow or deep) network. A sensory input $\mathbf{s} =\{s_1,\ldots,s_{N_s} \}$ elicits a cascade of noisy neural responses, up to the last coding layer with neural activities $\mathbf{r}=\{r_1,\ldots,r_{N} \}$. As concerns the read-out, there is $M$ output cells. Each output activity is a deterministic function $g_{y}(\mathbf{r})$ of the neural activity in the last coding layer. We consider these outputs as estimators of the posterior probability $P(y|\mathbf{s})$, where $\mathbf{s}$ is the (true) stimulus that elicited the neural activity $\mathbf{r}$. Throughout this paper we interchangeably note the output as either a function, $g_{y}(\mathbf{r})$, or as a probability, $g(y|\mathbf{r})$. Finally, the category corresponding to the largest output $g_{y}(\mathbf{r})$ provides the estimate of the true category. 

We denote with capital letters the random variables, e.g., $Y, \mathbf{S}, \mathbf{R}$, and with small letters particular realizations, such as $y, \mathbf{s}, \mathbf{r}$.

\subsection{Bayesian approach}
\label{sec:meanBayescost}

\subsubsection{Mean Bayes cost}
\label{sec:meancost}
For a given stimulus $\mathbf{s}$ and a neural activity $\mathbf{r}$ in the last coding layer, the relevant Bayesian quality criterion is given by the discrepancy $\mathcal{C}(\mathbf{s},\mathbf{r})$ between the true probabilities $\{P(y|\mathbf{s}), y=1,...,M\}$ and the estimator $\{g_{y}(\mathbf{r}), y=1,...,M\}$, defined as a Kullback-Leibler divergence (or relative entropy) \citep{Cover_Thomas_2006}:
\beq
\mathcal{C}(\mathbf{s},\mathbf{r}) = D_{\text{KL}}(P(Y|\mathbf{s}) \| g(Y|\mathbf{r})),
\label{eq:cost}
\eeq
with
\beq
D_{\text{KL}}(P(Y|\mathbf{s}) \| g(Y|\mathbf{r})) \equiv \sum_{y=1}^{M} P(y|\mathbf{s}) \ln \frac{P(y|\mathbf{s})}{g(y|\mathbf{r}) }
\label{eq:DKLpg}
\eeq
(in this paper we  make use of this common notation for Kullback-Leibler divergences). Averaging over $\mathbf{r}$ given $\mathbf{s}$, and then over $\mathbf{s}$, the mean cost induced by the estimation can be written:
\beq
\mathcal{\overline{C}}[Y, \mathbf{S}, \mathbf{R}] = - \mathcal{H}[Y|\mathbf{S}] + \textstyle{\iint} \, \left( -\;\textstyle{\sum}_y P(y|\mathbf{s}) \ln g(y|\mathbf{r})\right) P(\mathbf{r}|\mathbf{s})\,P(\mathbf{s}) \,d^{N_s}\mathbf{s} \;d^{N}\mathbf{r},
\label{eq:cm1_bis}
\eeq
where 
\beq
\mathcal{H}[Y|\mathbf{S}] = \textstyle{\int} \left(-\,\textstyle{\sum}_{y=1}^{M} P(y|\mathbf{s}) \ln P(y|\mathbf{s}) \right)\, P(\mathbf{s})\,d^{N_s}\mathbf{s} 
\eeq
is the conditional entropy of the category membership given the stimulus. 

\subsubsection{Link with the cross-entropy loss function} 
\label{sec:algo}
As discussed in Refs.~\citep{LBG_These,LBG_JPN_2012}, the above cost function $\mathcal{\overline{C}}$ is directly related to the cross-entropy loss commonly used in supervised learning (see, e.g., Ref.~\citep{chollet2017book}). For completeness, we restate this result in the present context. For a given stimulus $\mathbf{s}$, the target (teacher) output is $t_y(\mathbf{s})=1$ if $\mathbf{s}$ belongs to category $y$, and $t_y(\mathbf{s})=0$ otherwise. The cross-entropy loss characterizing the discrepancy between the target and the network output $g_{y}(\mathbf{r})$ is given by $\mathcal{C}_{\text{CE}}(\mathbf{s},\mathbf{r})=-\textstyle{\sum}_{y=1}^M t_y(\mathbf{s}) \ln g_{y}(\mathbf{r})$. Note that, since $t_y(\mathbf{s})$ is $0$ or $1$, this is also the Kullback-Leibler (KL) divergence $\textstyle{\sum}_{y=1}^M t_y(\mathbf{s}) \ln \frac{t_y(\mathbf{s})}{g_{y}(\mathbf{r})}$. Its average $\mathcal{C}_{\text{CE}}(\mathbf{s})$ over all possible neural activities given the stimulus is $\mathcal{C}_{\text{CE}}(\mathbf{s}) =-\; \textstyle{\sum}_{y} \textstyle{\int} \, P(\mathbf{r}|\mathbf{s})\, t_y(\mathbf{s}) \ln g_{y}(\mathbf{r}) \; d^{N}\mathbf{r}$. In the limit of a very large training set, according to the law of large numbers, the sum of the costs $\mathcal{C}_{\text{CE}}(\mathbf{s})$ over the examples $\mathbf{s}$ converges toward the statistical mean of the cross-entropy loss:
\beq
\overline{\mathcal{C}_{\text{CE}}}[Y, \mathbf{S}, \mathbf{R}] = \;-\; \textstyle{\sum}_{y} P_{y} \textstyle{\iint} \, P(\mathbf{r}|\mathbf{s}) P(\mathbf{s}|y) \ln g_{y}(\mathbf{r}) \; d^{N}\mathbf{r} \, d^{N_s}\mathbf{s}.
\label{eq:algo4}
\eeq
Making use of the Bayes rule $ P(\mathbf{s}|y)P_{y}=P(y|\mathbf{s}) P(\mathbf{s})$, one gets
\beq
\overline{\mathcal{C}_{\text{CE}}}[Y, \mathbf{S}, \mathbf{R}] = \;-\; \textstyle{\sum}_{y} \textstyle{\iint} \, P(y|\mathbf{s}) \ln g_{y}(\mathbf{r}) \; P(\mathbf{r}|\mathbf{s}) \, P(\mathbf{s})\, d^{N_s}\mathbf{s} \, d^{N}\mathbf{r} .
\label{eq:algo5}
\eeq
This is the same expression as the one for $\overline{\mathcal{C}}[Y, \mathbf{S}, \mathbf{R}]$, Eq. (\ref{eq:cm1_bis}), except for the term $\mathcal{H}[Y|\mathbf{S} ]$, that is
\beq
\overline{\mathcal{C}_{\text{CE}}}[Y, \mathbf{S}, \mathbf{R}] = \overline{\mathcal{C}}[Y, \mathbf{S}, \mathbf{R}]\;+\; \mathcal{H}[Y|\mathbf{S} ].
\label{eq:algo6}
\eeq
Since $\mathcal{H}[Y|\mathbf{S}]$ is a constant -- it only depends on the statistical links between categories and stimuli --, the minimization of $\overline{\mathcal{C}_{\text{CE}}}[Y, \mathbf{S}, \mathbf{R}]$ is equivalent to the one of $\overline{\mathcal{C}}[Y, \mathbf{S}, \mathbf{R}]$. In other words, the use of the cross-entropy loss in supervised learning is equivalent to a stochastic gradient descent for the mean cost $\overline{\mathcal{C}}$. 

\subsection{Decoupling into coding and decoding tasks} 
\label{sec:decoupling}
In the expression (\ref{eq:algo4}) of $\overline{\mathcal{C}_{\text{CE}}}$ (which is thus equal to $\overline{\mathcal{C}}+\mathcal{H}[Y|\mathbf{S}]\,$), we perform the integration over $\mathbf{s}$, $ \textstyle{\int} \, P(\mathbf{r}|\mathbf{s}) P(\mathbf{s}|y) d^{N_s}\mathbf{s}= P(\mathbf{r}|y)$, and with $P(\mathbf{r}|y)P_{y}=P(y|\mathbf{r})P(\mathbf{r})$, one has
\beq
\overline{\mathcal{C}}[Y, \mathbf{S}, \mathbf{R}]=-\mathcal{H}[Y|\mathbf{S}] - \textstyle{\sum}_{y} \textstyle{\int} \,P(y|\mathbf{r}) \,\ln g_{y}(\mathbf{r}) \;P(\mathbf{r}) d^N\mathbf{r}.
\eeq
We add and subtract $\mathcal{H}[Y|\mathbf{R}]$ to the right hand side of this equation. We have $\mathcal{H}[Y|\mathbf{R}]- \mathcal{H}[Y|\mathbf{S}]= I[Y,\mathbf{S}] - I[Y,\mathbf{R}]$, where $I[.,.]$ denotes the mutual information between two random variables, e.g.,
\beq
I[Y, \mathbf{S}]=\mathcal{H}[Y]-\mathcal{H}[Y|\mathbf{S}].
\eeq
Hence one gets that one can rewrite the mean Bayes cost as
\beq
\mathcal{\overline{C}} [Y, \mathbf{S}, \mathbf{R}]
= \mathcal{\overline{C}}_{\text{coding}}[Y, \mathbf{S}, \mathbf{R}] + \mathcal{\overline{C}}_{\text{decoding}}[Y, \mathbf{R}],
\label{eq:cost_main_expression}
\eeq
with
\beq
\mathcal{\overline{C}}_{\text{coding}}[Y, \mathbf{S}, \mathbf{R}] \,=\, I[Y,\mathbf{S}] - I[Y,\mathbf{R}],
\label{eq:Ccod}
\eeq
and 
\beq
\mathcal{\overline{C}}_{\text{decoding}}[Y, \mathbf{R}] \,=\, \textstyle{\int} \;
D_{\text{KL}}(P(Y|\mathbf{r}) \| g(Y|\mathbf{r}))\;P(\mathbf{r}) \,d^{N}\mathbf{r}.
\label{eq:Cdec} 
\eeq
The latter is the average over the neural activity of the Kullback-Leibler divergence of $P(Y|\mathbf{r})$ from the network output $g_{Y}(\mathbf{r})$.

As a consequence of this decomposition, Eq. (\ref{eq:cost_main_expression}), one can study separately the decoding and coding tasks, as discussed below, Secs. \ref{sec:optdecod} and \ref{sec:optcod} respectively.\\

We also mention here that, in Sec. \ref{sec:biasvar}, we discuss an alternative decomposition of the mean cost, analogous to a bias-variance decomposition.

\subsection{Optimal decoding} 
\label{sec:optdecod}
The decoding cost $\mathcal{\overline{C}}_{\text{decoding}}$, Eq. (\ref{eq:Cdec}), is the average relative entropy between the true probability of the category given the neural activity, and the output function $g$. It is the only term in the total cost $\mathcal{\overline{C}}$ depending on $g$, hence the function $g$ minimizing $\mathcal{\overline{C}}$ is the one minimizing $\mathcal{\overline{C}}_{\text{decoding}}$, that is (if it can be realized),
\beq
g_{y}(\mathbf{r}) = P(y|\mathbf{r}).
\label{eq:optdecod}
\eeq
Given our choice of cost function, the goal of the categorization task is to approximate the probability of the category given the input. However, in practice, one is interested in finding the most likely category given the stimulus. Learning with the cross-entropy loss may provide good performance for this task before the more demanding estimation task of the probabilities is fully achieved.

In Ref.~\citep{LBG_JPN_2012}, we considered the biologically motivated simplified case where the stimulus space is identified with a feature space of small dimension. Then, in an asymptotic limit of a very large number of coding cells, this estimator (\ref{eq:optdecod}) of $P(y|\mathbf{s})$ is efficient: it is unbiased and saturates the associated Cramér-Rao bound. In the present context of multilayer networks, we reconsider the efficiency of decoding below, Sec. \ref{sec:efficientdecod}. We do this by formalizing the hypothesis of structured data -- making explicit the existence of a feature (latent) space of small dimension, different from the stimulus space of large dimension -- and a similar hypothesis for the neural activity. 
 
\subsection{Optimal coding: Infomax}
\label{sec:optcod}
The coding cost (\ref{eq:Ccod}) is the difference between the information content of the signal, and the mutual information between category membership and neural activity. Since processing cannot increase information (``data processing inequality'', see, e.g., Ref.~\citep{Blahut_1987}), the information $I[Y,\mathbf{R}]$ conveyed by the neural activity about the category is at most equal to the one conveyed by the sensory input. That is,
\beq
I[Y,\mathbf{R}] \leq I[Y, \mathbf{S}].
\eeq
\label{eq:data_proc_inequ}
Note that, if one considers the succession of layers $l=1,...,L$, with neural activities 
$\mathbf{r}^1=\{r^1_1,\ldots,r^1_{N_1} \}$, ..., $\mathbf{r}^L=\{r^L_1,\ldots,r^L_{N_L} \}$ ($\mathbf{r}=\mathbf{r}^L$, $N_L=N$), 
the data processing inequality implies
\beq
I[Y,\mathbf{R}^{L}] \leq ...\leq I[Y, \mathbf{R}^{l+1}] \leq I[Y, \mathbf{R}^l] \leq ...\leq I[Y,\mathbf{R}^1] \leq I[Y, \mathbf{S}].
\label{eq:data_proc_inequ_L}
\eeq
The number of categories being finite, note also that $I[Y, \mathbf{S}]$ is itself at most equal to the entropy $\mathcal{H}[Y]$ of the category distribution:
\beq
 I[Y, \mathbf{S}] \leq \mathcal{H}[Y] \;\leq\; \ln M.
\label{eq:Imax}
\eeq

Since $\mathcal{\overline{C}}_{\text{coding}} = I[Y, \mathbf{S}] - I[Y, \mathbf{R}] \geq 0$, its minimization is equivalent to the maximization of the mutual information between neural activity and category membership:
\beq
\min \mathcal{\overline{C}}_{\text{coding}}[Y, \mathbf{S}, \mathbf{R}] \equiv	\max I[Y, \mathbf{R}].
\eeq
Hence the infomax principle~\citep{linsker1988selforganization} is here an outcome of the global Bayesian optimization problem.

Note that, if it is possible to find parameters such that the optimal estimator is reached, that is (\ref{eq:optdecod}) is realized, then the full average cost function (\ref{eq:cost_main_expression}), $\mathcal{\overline{C}} = \mathcal{\overline{C}}_{\text{coding}} + \mathcal{\overline{C}}_{\text{decoding}}$, reduces to $\mathcal{\overline{C}}_{\text{coding}} = I[Y, \mathbf{S}] - I[Y, \mathbf{R}]$.\\

The decomposition of the cost function in coding and decoding parts shows that each problem can be dealt with separately. The coding part of the network has to maximize the mutual information between neural activity and category, without taking into account what the decoding part is doing. The decoding part of the network must built the best estimator given what is fed into it from the coding layers -- even if this coding part is not optimized: in Ref.~\citep{LBG_JPN_2012} we made use of this property for the interpretation of experimental data from a psycholinguistic experiment.

\subsection{Link with the Information Bottleneck approach} 
\label{sec:IB}
Tishby, Pereira and Bialek introduced the Information Bottleneck (IB) approach \cite{tishby99information}, 
which can be formulated as a rate distortion problem. The considered learning cost is a distortion function that measures how well the category $y$ is predicted from the compressed noisy neural representation $\mathbf{r}$, compared with its prediction from the stimulus $\mathbf{s}$. Tishby and collaborators developed this framework, theoretically and algorithmically, first in the computational neuroscience context, then in the deep learning context, see, e.g., Refs.~\citep{tishby-zaslavsky2015} and \citep{schwartz-ziv2017opening}. Authors have challenged the genericity of some of their numerical results, finding in particular that they may actually depend on the choice of transfer function (sigmoidal vs. ReLU)~\citep{saxe2018ontheIB}. For efficient implementation, Alemi {\em et al}~\cite{alemi2017}  proposed the variational information bottleneck (VIB), an approximation scheme to handle the IB cost function for learning in deep networks.

The qualitative idea of the IB approach is that the neural activity should convey as little information as possible about the stimulus provided the information about the category is preserved. Thus, with our notation, the goal is to minimize $I[\mathbf{S},\mathbf{R} ] - \beta I[Y,\mathbf{R} ]$ where $\beta$ is a Lagrange multiplier. Analysing this optimization principle, \citet{tishby99information} 
show that the Kullback-Leibler divergence $D_{\text{KL}}(P(Y|\mathbf{s})\|(P(Y|\mathbf{r}))$ ``emerges'' as the relevant effective distortion measure. This divergence corresponds to our cost function once the decoding stage is optimized, that is $g_{y}(\mathbf{r}) = P(y|\mathbf{r})$. Then one sees that our approach is somewhat dual to the IB one. We start from the Kullback-Leibler divergence, and the infomax criterion emerges from the cost function. There are however two differences. First, the full cost function that we consider includes the decoding part, and second, the correspondence is with the IB cost in the $\beta \rightarrow \infty$ limit (see below).\\
An alternative way to see this correspondence is to consider, from a distortion measure viewpoint, the IB cost associated with the Bayes cost (\ref{eq:cm1_bis}):
\beq
\mathcal{\overline{C}}_{\text{IB}}(\beta) = I[\mathbf{S},\mathbf{R} ] + \beta \, \mathcal{\overline{C}}.
\label{eq:cost_ib}
\eeq
Making use of the decomposition (\ref{eq:cost_main_expression}) in coding and decoding parts for $\mathcal{\overline{C}}$, we can write
\beq
\mathcal{\overline{C}}_{\text{IB}}(\beta) 
= \mathcal{\overline{C}}_{\text{IB,coding}}(\beta) + \beta \, \mathcal{\overline{C}}_{\text{decoding}},
\label{eq:cost_main_expression_bis}
\eeq
where $\mathcal{\overline{C}}_{\text{decoding}}$ is given by (\ref{eq:Cdec}), and 
\beq
\mathcal{\overline{C}}_{\text{IB,coding}}(\beta) \,=\, I[\mathbf{S},\mathbf{R} ] + \beta\, \left( I[Y,\mathbf{S}] - I[Y,\mathbf{R} ] \right).
\label{eq:Ccod_IB}
\eeq
Since $I[Y,\mathbf{S}]$ is a constant, (\ref{eq:Ccod_IB}) is the usual information bottleneck cost function, and the large $\beta$ limit means maximizing the mutual information $I[Y,\mathbf{R} ]$.\\
As concerns the analysis and results in the present paper, we found that working at finite $\beta$ is not relevant. In Appendix \ref{app:optFisher}, however, we consider a finite $\beta$ as a regularization parameter to find the optimal relationship between the neural and the categorical Fisher information quantities resulting from the minimization of the cost in the large signal-to-noise ratio regime. In this appendix, we also briefly mention the possible link between large $\beta$ and large signal-to-noise ratio limits (large number of cells and/or large time limit in the context of neuroscience), with the occurrence of bifurcations at finite $\beta$ or non large times. From now on, in the main body of this paper, we stick to the Bayes cost function, Eq. (\ref{eq:cm1_bis}).

\section{Data and neural underlying feature spaces} 
\label{sec:featurespaces}
In this section we formalize the notion of structured data and of underlying space for the neural processing. We then derive results specific to data and neural activities characterized by underlying feature spaces of small dimensions, making use of the general framework introduced in the previous section.

\subsection{Manifold-structured data} 
\label{sec:underlyingman}
It is generally believed that structured data, such as natural images, lie on an underlying manifold of dimension typically small compared with the input dimension space. As nicely put forward in Ref. \citet{goldt2020modeling}, ``this manifold (...) constitutes the actual input space, or the ``world,'' of our problem. While the manifold is not easily defined, it is tangible: for example, its dimension can be estimated''. 

We assume the existence of such an underlying space for the input data, but we are mainly interested in the part that is relevant for the category membership. We denote this underlying feature space $X^*$, of dimension $K^*$ much smaller than the dimension of the data. 
We assume a sufficiency property:
\beq
P(y | \mathbf{s})= P(y | \mathbf{x}^*).
\label{eq:bayesiansufficiency}
\eeq
An obvious but important consequence of this property is that the signal information content satisfies
\beq
I[Y, \mathbf{S}]= I[Y, \mathbf{X}^*].
\label{eq:siginfo}
\eeq
We note also that, given (\ref{eq:bayesiansufficiency}), one can write the mean cost function (\ref{eq:cm1_bis}) in term of $\mathbf{X}^*$:
\beqa
\mathcal{\overline{C}}[Y, \mathbf{X}^*, \mathbf{R}] &=&
\iint P(\mathbf{r}|\mathbf{x}^*) P(\mathbf{x}^*) \sum_y P(y|\mathbf{x}^*) \ln \frac{P(y|\mathbf{x}^*)}{g_{y}(\mathbf{r})} \,d^{K^*}\mathbf{x}^* d^{N}\mathbf{r} .
\label{eq:cm1_x*}
\eeqa 
In the decomposition (\ref{eq:cost_main_expression}) in the coding and decoding parts, $\mathcal{\overline{C}} = \mathcal{\overline{C}}_{\text{coding}} + \mathcal{\overline{C}}_{\text{decoding}}$, the decoding part is unchanged, that is it is the same as in Eq. (\ref{eq:Cdec}), and for the coding part we have $\mathcal{\overline{C}}_{\text{coding}}[Y, \mathbf{S}, \mathbf{R}] = I[Y, \mathbf{S}] - I[Y,\mathbf{R} ] =I[Y, \mathbf{X}^*] - I[Y,\mathbf{R} ] = \mathcal{\overline{C}}_{\text{coding}}[Y, \mathbf{X}^*, \mathbf{R}]$.

In psychology and neuroscience, some protocols provide by design the control of the stimulus feature space. Typical examples are those where, by controlling a relevant feature, the experimentalist builds a series of morphs interpolating between stimuli (see, e.g., all the references mentioned in the first phrase of the introduction). But in machine learning, the data scientist has only access to the data. In such case, the underlying space $\mathbf{X}^*$ cannot be uniquely identified. Any (smooth) reversible transformation (change of representation) gives an equivalent space, for which the quantities of interest are invariant. 

\subsection{Feature space underlying neural activity}
\label{sec:networkfeaturespaces}
\subsubsection{Low-dimensional manifold} 
Similarly to, and coherently with, the hypothesis of structured data, many recent works show how both biological and artificial neural activities can be understood as acting on a manifold of lower dimension than the one of the input space and the one of the neural layer or pool involved in the task. For works in neuroscience, see, e.g., Refs.~\citep{sussillo2013opening,archer2014low,sadtler2014neural,cunningham2014dimensionality,gallego2017neural,mastrogiuseppe2018linking,chung2021neural,jazayeri2021interpreting}, and for the machine learning literature, see, e.g., Refs.~\citep{ma2018dimensionality, ansuini2019intrinsic, pope2021intrinsic}. In machine learning, the hypothesis of structured data is explicitly used for the design of neural architectures, as for autoencoders~\citep{hinton2006reducing}, the goal being to capture the data underlying manifold of possibly small dimension. 

The underlying manifold is not necessarily straightforwardly expressed in terms of the neural activities, exactly as the input data underlying space is not easily obtained from the data themselves. Here is a simple example borrowed from neuroscience. A network projects onto a manifold $X$ of small dimension $K$ in $\mathbb{R}^N$, such as, e.g., a two-dimensional manifold, and the neural activities give the coordinates in $\mathbb{R}^N$ of the stimulus location in this space. A particular case is the one of neural activities given by radial basis functions covering this space -- the analogous of a population code studied in neuroscience (see, e.g., Ref.~\cite{DayanAbbott}), with feature specific cells such as place cells~\cite{okeefe1971TheHippocampus} or head direction cells~\cite{taube1998HeadDirectionCells}. In these typical models of biological neural networks, the stochastic neural activity is parameterized by its mean and variance, in which case one may write, for each given neuron $i$, 
\begin{eqnarray}
E[R_i \vert \mathbf{s}] & = & f_i(X(\mathbf{s})),\\
Var[R_i \vert \mathbf{s}] & = & v_i(X(\mathbf{s})),
\end{eqnarray}
with for instance $f_i(x)=R_i^{\text{max}} f(\frac{x-x_i}{a_i})$ ($f_i$ is the tuning curve of neuron $i$), and for Poisson noise, $v_i= f_i$. The function $f$ decreases from $1$ to $0$ as its argument goes from $0$ to $\pm \infty$, $R_i^{\text{max}}$ is the maximum rate that $i$ can achieve, $x_i$ is the preferred stimulus for $i$ (the center of the radial function), and $a_i$ is the width of the tuning curve (the radius of the radial function). As an artificial network example, in Ref.~\cite{LBG2023interpolation} the author generates artificial high-dimensional data from a two-dimensional (2D) space $X^{*}$. A one hidden layer network learns to identify ten categories. Then using an autoencoder allows to reveal the low-dimensional space $X$ underlying the high-dimensional neural activity in the hidden layer (see Fig. 2 in Ref.\cite{LBG2023interpolation}). Furthermore, the analysis shows that through learning the network selects a space $X$ with the same dimension as the one of $X^{*}$.

\subsubsection{Markov chain}   

We formalize the hypothesis of the existence of an underlying projection space $X$ associated with a neural coding activity as follows. We assume that the network \emph{implicitly} realizes a deterministic non-linear transformation of the data underlying manifold through the transformation of the input. In the following we refer to this manifold $X$ as the underlying feature or projection space (or for short projection space), associated with the neural activity of the coding layer. As discussed below, in the course of learning we expect $X$ to become more and more category-specific, possibly becoming a non-linear transformation of the data category-specific underlying manifold, $X^*$.
 
Focusing on the coding layer with neural activity $\mathbf{r}$, the network processing chain is
\beq
\mathbf{s} \rightarrow  \mathbf{r} \rightarrow \mathbf{g}.
\label{eq:processing-chain} 
\eeq
The multilayer feedforward processing is here decomposed into the coding of the stimulus $\mathbf{s}$ by the neural activity $\mathbf{r}$, followed by the decoding of the category given by the output of the network $\mathbf{g}$. Our hypotheses on the data and neural underlying spaces can be summarized by the following Markov chain
\beq
y \rightarrow \mathbf{x}^* \rightarrow \mathbf{s} \rightarrow \mathbf{x}
\rightarrow \mathbf{r} \rightarrow \mathbf{g}.
\label{eq:markov-chain-X} 
\eeq
In the following, some analysis are specific to the projection space, $\mathbf{X}$, with no explicit dependency on the data space $\mathbf{X}^*$. In such cases, the relevant Markov (sub)chain to consider is simply 
\beq
y \rightarrow \mathbf{s} \rightarrow \mathbf{x}
\rightarrow \mathbf{r}.
\label{eq:markov-chain-small} 
\eeq
We note that this formalization applies as well to any intermediate layer. For a given layer, in the above chains (\ref{eq:markov-chain-X}), (\ref{eq:markov-chain-small}), the neural activity $\mathbf{r}$ is then that of this layer, and $\mathbf{X}$ is the associated underlying space. We also note that our formalization allows to coherently combine the generative and projection viewpoints. The Markov chain (\ref{eq:markov-chain-X}) corresponds to a generative viewpoint. For instance, the stimulus is a (deterministic or stochastic) function of a point in the manifold $X^*$. We can also adopt a projection viewpoint, considering for instance $\mathbf{x}^*$, or $\mathbf{x}$, as some deterministic function of the data, $\mathbf{x}^*=X^*(\mathbf{s})$, $\mathbf{x}=X(\mathbf{s})$. 

\subsubsection{Information content of the projection space}
\label{sec:qualprojspace}

The quality of the projection $X$ is given by how much the probability of the category given the stimulus is well approximated by the probability of the category given the projection $X(\mathbf{s})$. This is measured by the mean Bayes cost
\begin{eqnarray}
\overline{C}_X
&\equiv &\textstyle{\int} D_{\text{KL}}(P(Y|\mathbf{s}) \| P(Y|X(\mathbf{s}) )) \, P(\mathbf{s}) d^{N_s}\mathbf{s} \nonumber \\
&=& \int \sum_{y=1}^{M} P(y|\mathbf{s}) \ln \frac{P(y|\mathbf{s})}{P(y|X(\mathbf{s})) }\, P(\mathbf{s}) d^{N_s}\mathbf{s}.
\label{eq:costX}
\end{eqnarray}
We can write the above cost as
\beq
\overline{C}_X = -\mathcal{H}[Y|\mathbf{S}]\;+\;\mathcal{H}[Y|\mathbf{X}], 
\eeq
that is, adding and subtracting the category entropy $\mathcal{H}[Y]$,
\beq
\overline{C}_X = I[Y, \mathbf{S}]\;-\; I[Y, \mathbf{X}].
\label{eq:LX}
\eeq
Hence minimizing the mean Bayes cost (\ref{eq:costX}) is equivalent to maximizing the mutual information between the categories and the projection space. From the analysis in Sec. \ref{sec:general_framework}, we have
\beq
\mathcal{\overline{C}}_{\text{coding}} = I[Y,\mathbf{S}] - I[Y,\mathbf{R}]. 
\label{eq:Ccod_bis} 
 \eeq
Given the Markov chain (\ref{eq:markov-chain-X}), from the data processing theorem, we have
\beq
I[Y,\mathbf{R}] \leq I[Y, \mathbf{X}] \leq I[Y,\mathbf{S}], 
\eeq
so that we can expect that maximizing $I[Y,\mathbf{R}]$ will tend to increase $I[Y, \mathbf{X}]$.

Under the hypothesis of an underlying manifold $X^*$, if the network can find a projection space $X$ equivalent with respect to the categories to $X^*$, that is such that $P(y | X(\mathbf{s}))= P(y | X^*(\mathbf{s}))=P(y | \mathbf{s})$, then optimal coding is achieved with $\overline{C}_X=0$.

\subsection{Efficient decoding in a large signal-to-noise limit}
\label{sec:efficientdecod}
As we have seen Sec. \ref{sec:optdecod}, $\mathbf{r}$ being the neural activity of the last coding layer, the optimal decoder is obtained for having as output activities $g_{y}(\mathbf{r})=P(y|\mathbf{r})$, as best estimator of $P(y|\mathbf{s})$. 

This estimator is unbiased if
\beq
 \textstyle{\int} P(y|\mathbf{r}) P(\mathbf{r}|\mathbf{s}) d^{N}\mathbf{r} = P(y|\mathbf{s})
 \label{eq:optdecod_unbiased_est_bis}
 \eeq
(see Ref.~\cite{LBG_JPN_2012}, Appendix A.1). Given the Markov chain (\ref{eq:markov-chain-X}), at best decoding can extract $P(y|x)$. From Ref.~\citep{LBG_JPN_2012}, we have that, in a regime of high signal to noise ratio (large $N$ limit, with $\mathbf{x}$ of small dimension) $g_{y}(\mathbf{r})=P(y|\mathbf{r})$ is an unbiased, efficient, estimator of $P(y|\mathbf{x})$. In particular we have, at leading order in the number $N$ of neurons,
 \beq
 \textstyle{\int} P(y|\mathbf{r}) P(\mathbf{r}|\mathbf{x}) \,d^{N}\mathbf{r} = P(y|\mathbf{x}).
 \label{eq:optdecod_unbiased_est}
 \eeq
Note that (\ref{eq:optdecod_unbiased_est}) implies
\beq
 \textstyle{\int} P(y|\mathbf{r}) P(\mathbf{r}|\mathbf{x}=X(\mathbf{s})) \,d^{N}\mathbf{r} = P(y|\mathbf{x}=X(\mathbf{s})).
\eeq
But this does not necessarily imply (\ref{eq:optdecod_unbiased_est_bis}). It will be the case if the knowledge of $X$ is sufficient for estimating $y$, that is if for every $y$
\beq
P(y | X(\mathbf{s}))=P(y | X^*(\mathbf{s}))=P(y | \mathbf{s}),
\label{eq:sufficiency}
\eeq
in which case the network has found $X$ for which the cost $\overline{C}_X$, Eq. (\ref{eq:LX}), is exactly zero. Otherwise, one has a bias which corresponds to the nonzero value of the Kullback-Leibler divergence $\overline{C}_X$. 

The fact that the estimator is efficient means that it saturates the associated Cramér-Rao bound. This is the Cramér-Rao bound for the estimation of a function of the unknown parameter, which can be understood as the bound for a biased estimator of the parameter -- see, e.g., Ref. \citep{Cover_Thomas_2006}. 
In the $K=1$ case, this bound reads:
\beq
\int P(\mathbf{r}|x) \big(g_{y}(\mathbf{r})-P(y|x)\big)^2\, d^N\mathbf{r}\,
 = \frac{\big(P'(y|x)\big)^2}{F_\text{code}(x)},
 \label{eq:optim_var}
 \eeq
 where $P'(y |x)$ denotes the derivative of $P(y|x)$ with respect to $x$, and $F_{\text{code}}(x)$ is the Fisher information defined by
\beq
F_{\text{code}}(x) = - \int \, \frac{\partial^2 \ln P(\mathbf{r}|x) }{\partial x^2} \; P(\mathbf{r}|x) \,d^N\mathbf{r}.
\label{eq:fisher_code_1}
\eeq 
In Appendix \ref{app:fisher-s-x}, we briefly discuss what can be said for the Cramér-Rao bound in terms of the dependency of the output with respect to the stimulus $\mathbf{s}$ instead of $\mathbf{x}$.

As we will see, the above Fisher information plays an important role in the analysis of the coding part, see Sec. \ref{sec:geometry}.

\subsection{Bias-variance decompositions of the mean Bayes cost}
\label{sec:biasvar}
Initially introduced for quadratic error cost functions~\cite{Geman1992BVDilemma}, the bias-variance decomposition has been generalized to a variety of loss functions \cite{Domingos2000AUnified,Buja2005LossFns,pfau2013generalized}. The loss function that we consider in the present paper is based on a Kullback-Leibler divergence, which belongs to the family of Bregman divergences~\cite{Bregman1967}, for which bias-variance decompositions have been studied~\cite{Buja2005LossFns,pfau2013generalized,Yang2020RethinkingBVT,adlam2022understanding}. Bias-variance decompositions are typically discussed in the context of learning from a finite number of examples, allowing to highlight a learning dilemma~\cite{Geman1992BVDilemma}. In that context, the training set is considered as a random sampling of the data distribution, so that the output of the network is a random variable. In the present paper we work with the full distribution of the data. However, we consider processing noise, so that the network output $g_{y}(\mathbf{r}), y=1,..., M$ is as well a random variable. We can then consider bias-variance type decompositions as shown in Ref.~\cite{pfau2013generalized} for Bregman divergences within a general setting (see ``Theorem 0.1'' in this paper).

Here we make explicit the relevant bias-variance decompositions specific to our framework. The motivation is to search for relations giving insights in the spirit of the Cramér-Rao bound. Given an estimator, the bias corresponds to the discrepancy between the target and the mean of the estimator. Introducing this mean into the expression of the total mean cost leads to a bias-variance type decomposition, as we show now.

We first consider the processing $y \rightarrow \mathbf{s} \rightarrow \mathbf{r} \rightarrow \mathbf{g}$. We remind that the network outputs, $g_{y}(\mathbf{r}), y=1,..., M$, are considered as estimators of the probabilities $P(y| \mathbf{s})$. For a given set of normalized positive functions $g_{y}$ ($\textstyle{\sum}_{y} g_{y}(\mathbf{r})=1$ for any $\mathbf{r}$),
the mean is
\beq
 \overline{g_{y}}(\mathbf{s})\equiv E[g_{y}|\mathbf{s}] = \textstyle{\int} g_{y}(\mathbf{r}) P(\mathbf{r}|\mathbf{s}) d^{N}\mathbf{r}. 
 \label{eq:bias_main}
\eeq
Note that the normalization is preserved: for any $\mathbf{s}$, $\textstyle{\sum}_{y} \overline{g_{y}}(\mathbf{s})= 1$. In case the estimator would be unbiased, one would have $\overline{g_{y}}(\mathbf{s})= P(y|\mathbf{s})$. 

The total mean cost $\mathcal{\overline{C}}$, Eq.~(\ref{eq:cm1_bis}), can be written as
\beq
\mathcal{\overline{C}} = \textstyle{\int} \, \mathcal{\overline{C}}(\mathbf{s})\,P(\mathbf{s})\,d^{N_s}\mathbf{s},
\label{eq:cbar_main}
\eeq
with
\beq
\mathcal{\overline{C}}(\mathbf{s}) = \textstyle{\int} \, D_{\text{KL}}(P(Y|\mathbf{s})\|g(Y|\mathbf{r}))\,P(\mathbf{r}|\mathbf{s})\, d^{N}\mathbf{r}.
\label{eq:cbarx_main}
\eeq
One can write
\beq
D_{\text{KL}}(P(Y|\mathbf{s})\|g(Y|\mathbf{r})) \,=\, 
D_{\text{KL}}(P(Y|\mathbf{s})\|\overline{g}(Y|\mathbf{s}))
\,+\, \sum_{y=1}^{M} P(y|\mathbf{s}) \ln \frac{\overline{g_{y}}(\mathbf{s})}{g_{y}(\mathbf{r})},
\label{eq:cost2_main}
\eeq
and the mean cost for a given input can then be written
\beqa
	\mathcal{\overline{C}}(\mathbf{s}) &=& 
	D_{\text{KL}}(P(Y|\mathbf{s})\|\overline{g}(Y|\mathbf{s})) \\ &+&
	\sum_{y=1}^{M} P(y|\mathbf{s}) \left\{ \ln \overline{g_{y}}(\mathbf{s}) - \int \ln g_{y}(\mathbf{r})\,P(\mathbf{r}|\mathbf{s}) \,d^{N}\mathbf{r} \right\}.
	\label{eq:cbarx2_main}
\eeqa
The first term quantifies the cost for having a bias, it is positive or zero and goes to zero as the bias cancels. The second term is also positive or zero. Indeed, for each $\mathbf{s}$ and each $y$, the term within $\left\{...\right\} $ is the log of an average minus the average of the log, which is a positive quantity by convexity of the logarithm (Jensen's inequality \cite{Jensen_1905,Jensen1906}). This quantity is small if the variance of the estimator is small (since in that case $g_{y}(\mathbf{r})$ is typically close to its mean $\overline{g_{y}}(\mathbf{s})$). Eq.~(\ref{eq:cbarx2_main}) is thus a bias-variance decomposition of the mean cost. 

We now write a similar decomposition taking into account the discrepancy between the underlying manifolds $X^*$ and $X$. After some manipulations analogous to those above, we get for the global mean cost a sum of three positive terms:
\begin{eqnarray}
\mathcal{\overline{C}} &=& 
\textstyle{\int}\, D_{\text{KL}}(P(Y|\mathbf{x}^*)\|P(Y|\mathbf{x})) \;P(\mathbf{x}^*,\mathbf{x}) \,d^{K^*}\mathbf{x}^*\,d^K\mathbf{x}
\label{eq:cbarx3-a_main} \\
& + & 
\textstyle{\int}\, D_{\text{KL}}(P(Y|\mathbf{x})\|\overline{g}(Y|\mathbf{x})\,)
\,P(\mathbf{x})\,d^K\mathbf{x} 
\label{eq:cbarx3-b_main} \\
& + & \textstyle{\int} \textstyle{\sum}_{y=1}^{M} P(y|\mathbf{x}) \left\{ \ln \overline{g_{y}}(\mathbf{x}) - \textstyle{\int} \ln g_{y}(\mathbf{r})\,P(\mathbf{r}|\mathbf{x})\,d^{N}\mathbf{r} \right\} P(\mathbf{x})d^K\mathbf{x}.
\label{eq:cbarx3-c_main}
\end{eqnarray}
Here $P(\mathbf{x}^*,\mathbf{x})$ is the joint distribution $P(\mathbf{x}^*,\mathbf{x}) = \textstyle{\int} \, \delta(\mathbf{x}^*-\mathbf{X}^*(\mathbf{s})) \, \delta(\mathbf{x}-\mathbf{X}(\mathbf{s}))\,P(\mathbf{s})\,d^{N_s}\mathbf{s}$, and $\overline{g_{y}}(\mathbf{x}) = \textstyle{\int} \, \delta(\mathbf{x}-\mathbf{X}(\mathbf{s}))\, \overline{g_{y}}(\mathbf{s})\, P(\mathbf{s}) d^{N_s}\mathbf{s}$, $P(y|\mathbf{x}) = \textstyle{\int} \delta(\mathbf{x}-\mathbf{X}(\mathbf{s}))\, P(y|\mathbf{s})\, P(\mathbf{s})d^{N_s}\mathbf{s}$. The first term (\ref{eq:cbarx3-a_main}) measures how much the manifold $\mathbf{X}$ differs from $\mathbf{X}^*$ as discriminant space, and the two other terms give the bias-variance decomposition for $g_{y}$ as estimator of $P(y|\mathbf{x})$. This decomposition highlights the dilemma that learning is facing when trying to minimize the mean cost: finding the best network hidden feature space, and the best bias-variance compromise for the decoding. 

In Appendix \ref{app:biasvar} we show that, for an estimator close to efficiency, this bias-variance decomposition reduces to a quadratic type trade-off. In addition, we make use of this analysis to derive a bound on the cost. We also make use of known bounds for the Jensen gap to get bounds for the variance part, Eq. (\ref{eq:cbarx3-c_main}). 

\section{Revealing the geometry of internal representations} 
\label{sec:geometry}

\subsection{The mutual information in the limit of wide networks}
\label{sec:Nlarge}
During the course of learning the network will adapt both the manifold $X$ and its $N$-dimensional neural representation $\mathbf{r}$. Here we characterize the mutual information $I[Y,\mathbf{R}]$ for a given space $X$ of dimension $K$, when $N$ is large (wide network). The projection space $X$ is thus not necessarily (fully) optimized with respect to the categorization task. However, we assume that (i) the dimension $K$ of this space is small compared with $N$, (ii) given $\mathbf{r}$, the probability of what is the associated $\mathbf{x}$ is sharply peaked around the most probable value, $\mathbf{x}_m(\mathbf{r})$. Qualitatively, $K$ being fixed, the larger $N$, the more detailed the sampling of the $\mathbf{x}$ distribution. As a particular example, one may consider the neural units in the last coding layer as radial basis functions covering the $\mathbf{X}$-space. Below we extend to the present setting results obtained in Ref.~\cite{LBG_JPN_2008} for a model which corresponds here to the case $N_s=1$, $X(s)=s$ and $L=1$ (a single hidden layer with a large number of coding cells). Under the above hypothesis, in the infinite $N$ limit the full information content of the signal as seen by the layer, that is the stimulus projected onto $X$, is recovered:
\beq
\lim_{N \rightarrow \infty} I[Y,\mathbf{R}] = I[Y, \mathbf{X}].
\label{eq:mutu_infty}
\eeq

Now we compute the first correction in $1/N$. In the one-dimensional case ($K=1$), we get:
\beq
I[Y,\mathbf{R}]= I[Y,X] - \frac{1}{2} \int \frac{F_{\text{cat}}(x)}{F_{\text{code}}(x)}\; P(x)\,dx.
\label{eq:Delta_general_1d}
\eeq
Here $F_{\text{cat}}(x)$ and $F_{\text{code}}(x)$ are two Fisher information quantities whose definitions and meaning are as follows.\\
$F_{\text{cat}}(x)$, which we refer to as the categorical Fisher information, characterizes the sensitivity of the category membership with respect to small variations of $x$:
\beq
F_{\text{cat}}(x) = -\sum_{y=1}^M \, \frac{\partial^2 \ln P(y|x)}{\partial x^2}\; P(y|x),
\label{eq:fisher_cat}
\eeq
which can also be written as
\beq
F_{\text{cat}}(x) = \sum_{y=1}^M \, \frac{P'(y |x)^2}{P(y |x)},
\label{eq:fisher_cat_sum}
\eeq
where $P'(y |x) = \partial P(y |x)/\partial x$. 
As discussed in Ref.~\cite{LBG_JPN_2008}, $F_{\text{cat}}(x)$ is large at locations $x$ near a boundary between categories, and small if $x$ is well within a category.\\
The quantity $F_{\text{code}}(x)$, which we refer to as the neural Fisher information, characterizes the sensitivity of the neural activity $\mathbf{r}$ with respect to small variations of $x$. 
We have seen this Fisher information, Sec. \ref{sec:efficientdecod}, Eq. (\ref{eq:fisher_code_1}), as it also enters in the characterization of the decoding part. We recall its definition here:
\beq
F_{\text{code}}(x) = - \int \, \frac{\partial^2 \ln P(\mathbf{r}|x) }{\partial x^2} \; P(\mathbf{r}|x) \,d^N\mathbf{r}.
\label{eq:fisher_code}
\eeq
It corresponds to the ``usual'' Fisher information considered in neuroscience, and it is related to the discriminability measured in psychophysics \citep{MacmillanCreelman91}. We remind that the inverse of the Fisher information $F_{\text{code}}(x)$ is an optimal lower bound on the variance $\sigma_x^2 $ of any unbiased estimator $\widehat{x}(\mathbf{r})$ of $x$ (Cramér-Rao bound, see, e.g., Ref.~\citep{Cover_Thomas_2006}): 
\beq
\sigma_x^2 \equiv \int \, \big(\widehat{x}(\mathbf{r}) - x\big)^2 \;P(\mathbf{r}|x)\,d^N\mathbf{r}\;\geq \; \frac{1}{F_{\text{code}}(x)}.
\eeq
Note that $F_{\text{cat}}$ is independent of the neural code of the considered layer, and that, for $N$ coding cells, $F_{\text{code}}$ is of order $N$ (except for some particular families of correlations), so that the right-hand side of (\ref{eq:Delta_general_1d}) is of order $1/N$ (higher order terms are neglected). \\

In the more general case of a $K$-dimensional space, we get for $N \gg 1$ and $K \ll N$ (Appendix \ref{app:eqDelta_gen}):
\beq
I[Y,\mathbf{R}] \;=\; I[Y,\mathbf{X}] - \frac{1}{2} \int \operatorname{tr} \left( \mathbf{F}_{\text{cat}}^\mathsf{T} (\mathbf{x})\,\mathbf{F}_{\text{code}}^{\,-1}(\mathbf{x}) \right)\; P(\mathbf{x}) \,d^K\mathbf{x} 
\label{eq:Delta_general}
\eeq
where $\mathbf{F}_{\text{code}}(\mathbf{x})$ is the $K\times K$ Fisher information matrix of the neuronal population:
\beq
\big[\mathbf{F}_{\text{code}}(\mathbf{x})\big]_{ij} \; = - \int \, \frac{\partial^2 \ln P(\mathbf{r}|\mathbf{x})}{\partial x_i \partial x_j}\;P(\mathbf{r}|\mathbf{x}) \,d^N\mathbf{r} \, ,
\label{eq:fisher_code_matrix}
\eeq
$\mathbf{F}_{\text{cat}}(\mathbf{x})$ is the $K\times K$ Fisher information matrix of the categories:
\beq
\big[\mathbf{F}_{\text{cat}}(\mathbf{x})\big]_{ij} \; = -\sum_{y=1}^M \, \, \frac{\partial^2 \ln P(y|\mathbf{x})}{\partial x_i \partial x_j} \; P(y|\mathbf{x})\, ,
\label{eq:fisher_cat_matrix}
\eeq
and $\operatorname{tr}$, the superscripts $\mathsf{T}$ and $-1$, respectively denote the trace, and the matrix transpose and inverse. Although the Fisher information matrices are symmetric, in Eq.~(\ref{eq:Delta_general}) we keep the transpose sign on $\mathbf{F}_{\text{cat}}$ to better see the structure of the formulas (making the Frobenius product more obvious).\\

The above asymptotic formulas assume that the probability density functions are smooth enough so that the neural Fisher information exists (it is finite), and is invertible.\\
If the Fisher is not defined (infinite), the mutual information is still well defined, but there is no general expression for the asymptotic regime -- as in the case of the mutual information between the neural activity and a continuous parameter, for which there only exists scaling properties depending on the type of non smoothness, see Ref.~\cite{HausslerOpper1997}. However, as discussed in Ref.~\cite{LBG_JPN_2008}, for boxcar activation functions that are not differentiable everywhere, one can derive an analogous expression where, in place of the neural Fisher information, appears a quantity which characterizes, for the considered model, the smallest possible variance for an estimator of $x$.\\
The invertibility property assumes that we restrict the analysis to a space $X$ on which the neural Fisher information has no null eigenvalues.

An important remark is that, the mutual information being invariant under any reversible transformation on $\mathbf{x}$, this has also to be the case for the right-hand side of (\ref{eq:Delta_general}). In Appendix (\ref{app:invariance}) we show that (\ref{eq:Delta_general}) is indeed invariant under such a transformation.

\subsection{Discriminant spaces and geometry of internal representations}
\label{sec:Optlearning}
\subsubsection{Summary of the main results so far}
\label{sec:wherewestand}
Let us first summarize where we stand. We have first shown, Sec. \ref{sec:optcod}, that the minimization of the mean Bayes cost implies the minimization of the coding part, $ \mathcal{\overline{C}}_{\text{coding}}= I[Y,\mathbf{S}] - I[Y,\mathbf{R}]$, hence the maximization of the mutual information $I[Y,\mathbf{R}]$ between the categories and the neural representation provided by the network prior to decoding. Given the Markov chain (\ref{eq:markov-chain-X}), that is $y \rightarrow \mathbf{x}^* \rightarrow \mathbf{s} \rightarrow \mathbf{x} \rightarrow \mathbf{r}$, 
 we have
\beq
I[Y,\mathbf{R}] \leq I[Y, \mathbf{X}] \leq I[Y,\mathbf{S}]=I[Y,\mathbf{X}^*].
 \eeq
Thus, for a given projection space $X$, at best $I[Y,\mathbf{R}]= I[Y, \mathbf{X}]$, and optimization with respect to the choice of the space $X$ gives optimally $I[Y, \mathbf{X}]=I[Y,\mathbf{S}] =I[Y,\mathbf{X}^*]$.

Then, Sec. \ref{sec:Nlarge}, considering wide (and possibly deep) networks, within a specific asymptotic regime we have seen that the mutual information $I[Y, \mathbf{R}]$ takes the form 
\beq
I[Y,\mathbf{R}]= I[Y,\mathbf{X}] - \frac{1}{2} \int \frac{F_{\text{cat}}(x)}{F_{\text{code}}(x)}\; P(x)\,dx.
\label{eq:Delta_general_1d_bis}
\eeq
We reproduce here Eq. (\ref{eq:Delta_general_1d}), which is for the 1D case, but Eq. (\ref{eq:Delta_general}) gives the general $K$ dimensional case.

The interpretation of these asymptotic expressions is remarkably simple and intuitive, as we discuss now.

\subsubsection{Finding a proper discriminant space}
\label{sec:firsterm}
The first term, $I[Y,\mathbf{X}]$, characterizes the correlation between the categories and the underlying projection space $X$. Maximizing this term means finding a discriminant space, an appropriate space from the point of view of the categorization task. Efficient learning should lead to a space $X$ which contains a (possibly non-linear) transformation of the data category-specific underlying space, $X^*$. 
More precisely, in that case one has a sufficiency statistics property, $P(y|\mathbf{x})=P(y|\mathbf{x}^*)$. 

\subsubsection{The geometry of internal representations}
\label{sec:secondterm}

The second term tells us what should be the metrics of the neural representation, how this space $X$ should be probed: the Fisher information $F_{\text{code}}$ should be large where the categorical Fisher information $F_{\text{cat}}$ is large in order to minimize the second term. Thus for a given space $X$, minimization of the second term in the mutual information (\ref{eq:Delta_general_1d_bis}) leads to a neural code such that $F_{\text{code}}(x)$ is some increasing function of $F_{\text{cat}}(x)$ -- for, e.g., an information-theoretic constraint, in the vein of the Information Bottleneck approach~\citep{tishby99information}, one gets $F_{\text{code}}(x)=F_{\text{cat}}(x)$ as optimum (see Appendix \ref{app:optFisher}), but other constraints may lead to other relationships -- see Refs.~\cite{LBG_JPN_2008, KB_JPN_2020}. Efficient coding in view of optimal classification is thus obtained by essentially matching the two metrics. Since $F_{\text{cat}}$ is larger near a class boundary, this should also be the case for $F_{\text{code}}(x)$. A larger $F_{\text{code}}(x)$ around a certain value of $x$ means that the neural representation is stretched at that location (the neural representation tiles the space $x$ more finely near than far from the class boundaries). Thus, category learning implies better cross-category than within-category discrimination, hence the so-called \textit{categorical perception}.\\

Irrelevant dimensions will lead to Fisher information matrices with some null eigenvalues. The analytical results we give on the mutual information imply the inverse of the neural Fisher information $F_\text{code}$. We can expect that, during learning, it will be the case that the Fisher information is invertible. As we will see with numerical simulations, efficient learning will locally lead to zero or very small eigenvalues through alignment with the categorical information.\\

To conclude this section, minimizing the mean Bayes cost thus implies maximizing the mutual information $I[Y,\mathbf{R}]$, which leads to, on one hand, finding an appropriate projection space, and, on the other hand, building a neural representation with the appropriate metrics on this space. Given the intuitive character of the above conclusions, we expect their validity to be wider than for the cases for which the asymptotic formulas (\ref{eq:Delta_general_1d}) and (\ref{eq:Delta_general}) of the mutual information apply. Indeed, these formulas have been obtained under specific hypothesis, in particular assuming that asymptotically the probability of what is the associated $\mathbf{x}$ is sharply peaked around the most probable value $\mathbf{x}_m(\mathbf{r})$. This strong hypothesis might not be valid for any $\mathbf{x}$, however we expect these results to be valid under weaker hypotheses. Moreover, in the considered asymptotic limit, the noise is vanishing and the distribution becomes Gaussian. In Appendix \ref{app:singlecell} we discuss the simple case of a non-wide network, that of a single coding cell with small, non Gaussian, multiplicative noise. The resulting formula for the mutual information is the same as (\ref{eq:Delta_general_1d}), except for non Gaussian noise. In such case, the term depending on the Fisher information quantities is multiplied by a global factor, and thus the main qualitative results are not affected. 

We also compute, in Appendix \ref{app:Fisher_KN}, the neural Fisher information for the multidimensional case with additive noise of arbitrary distribution. The main conclusion is the same in the case of uncorrelated noise. However, for correlated noise, one gets that the neural Fisher information mixes three components: the noise amplitude, the shape of the noise distribution, and the local changes of metrics due to the transfer functions. The adaptation of the local neural metrics can thus be obtained in different ways through the combination of these components. We note this might have important consequences if one wants to uncover the underlying feature space. If one were to reconstruct it from the activity of a population of neurons in response to a set of stimuli, one would be faced with the potential issue that this space is not unique, due to the invariance of the mutual information under any invertible transformation. For instance, given a series of morphs that go from one category to another, equally spaced in stimulus space, one could find an $\mathbf{X}$-space where these stimuli are also equally spaced, but for which the neural activity on top of it is more sensitive at the boundary between categories. One could instead find an $\mathbf{X}$-space that itself carries the deformation, \textit{ie} where these stimuli are further away near the boundary, but now with the neurons responding more equally to the whole set. In the end, the geometry is of course the same with respect to the stimulus space, that is with greater sensitivity between categories.

\section{Categorical Fisher information: Discriminant directions and location of the maxima}
\label{sec:Fcat}
\subsection{The categorical Fisher information}
As just seen, learning should lead to the matching between the neural Fisher information and the categorical Fisher information. In particular the resulting neural geometry will show expansion of the space where the categorical Fisher information is the highest. Before considering the neural geometry after learning --which we do in the next section through numerical illustration--  we thus need to characterize the categorical Fisher information matrix, which is the goal of this section. Our previous work~\citep{LBG_JPN_2008} only qualitatively discussed the properties of the categorical Fisher information in the simplest case of a one-dimensional stimulus. Here we consider the multidimensional case, first for arbitrary distributions, then with detailed illustrations for Gaussian distributions. In particular, we study the eigenvectors and eigenvalues of the categorical Fisher information matrix, and the location of the maxima with respect to the location of the class boundaries.

We study the properties of the categorical Fisher information matrix $\mathbf{F}_{\text{cat}}(\mathbf{x})$ for $\mathbf{x}$ in a $K$-dimensional space and $M$ categories, assuming that the probabilities $P(y|\mathbf{x})$ are everywhere differentiable with respect to each one of the $K$ components $x_i$. In all this section, $\mathbf{x}$ is not specific. 
That is, it could be the location in the underlying space to the stimulus (in which case $\mathbf{x}=\mathbf{x}^*$), or associated with a given neural layer (not necessarily optimized), or it could be the stimulus $\mathbf{s}$ itself (see however Appendix \ref{app:fisher-s-x} for that case). Depending on the context, the dimension $K$ of this space might be small or large.

We recall that, for $i, j = 1,...,K$,
\beq
\big[\mathbf{F}_{\text{cat}}(\mathbf{x})\big]_{ij} \; = \;- \sum_{y=1}^M \, \, \frac{\partial^2 \ln P(y|\mathbf{x})}{\partial x_i \partial x_j} \; P(y|\mathbf{x}), 
\eeq
or, equivalently, 
\begin{eqnarray}
\big[\mathbf{F}_{\text{cat}}(\mathbf{x})\big]_{ij} &= &\; \sum_{y=1}^M \; \frac{\partial \ln P(y|\mathbf{x})}{\partial x_i} \, \frac{\partial \ln P(y|\mathbf{x})}{\partial x_j} \; P(y|\mathbf{x}) \nonumber \\
& =& \; \sum_{y=1}^M \; \frac{\partial_i P(y|\mathbf{x}) \; \partial_j P(y|\mathbf{x})}{P(y|\mathbf{x})}
\label{eq:fisher_cat_matrix_bis}
\end{eqnarray}
where $\partial_i$ stands for $\frac{\partial}{\partial x_i}$.

\subsection{The case of two categories}
\label{sec:fcat2cat}
\subsubsection{Principal (local) discriminant directions}
\label{sec:pdc}
We consider here the case of two categories, $M=2$, $y=\pm$, in $K$ dimensions. Since $\textstyle{\sum}_{y} P(y|\mathbf{x})=1 $, one has $\partial_i P(-|\mathbf{x}) = - \partial_i P(+|\mathbf{x})$. Hence we can write 
\begin{eqnarray}
\big[\mathbf{F}_{\text{cat}}(\mathbf{x})\big]_{ij} & = & \frac{\partial_i P(+|\mathbf{x}) \; \partial_j P(+|\mathbf{x})}{P(+|\mathbf{x})}\; + \;\frac{\partial_i P(-|\mathbf{x}) \; \partial_j P(-|\mathbf{x})}{P(-|\mathbf{x})} \nonumber \\
& = & \partial_i P(+|\mathbf{x}) \; \partial_j P(+|\mathbf{x}) \; \left( \frac{1}{P(+|\mathbf{x})} \,+\, \frac{1}{P(-|\mathbf{x})} \right) \nonumber \\
& = & \frac{\partial_i P(+|\mathbf{x}) \; \partial_j P(+|\mathbf{x})}{P(+|\mathbf{x})\;P(-|\mathbf{x})}.
\end{eqnarray}
In matrix form,
\beq
\mathbf{F}_{\text{cat}}(\mathbf{x}) = \frac{1}{P(+|\mathbf{x})\;P(-|\mathbf{x})}\grad P(+|\mathbf{x}) \grad P(+|\mathbf{x})^\mathsf{T}.
\eeq
From this expression one sees that $\grad P(+|\mathbf{x})$ is eigenvector of $\mathbf{F}_{\text{cat}}$ for the eigenvalue
\beq
f_{\text{cat}}(\mathbf{x}) = \operatorname{tr}\big[\mathbf{F}_{\text{cat}}(\mathbf{x})\big] = \frac{1}{P(+|\mathbf{x})\;P(-|\mathbf{x})}\, \textstyle{\sum}_j [\partial_j P(+|\mathbf{x})]^2.
 \label{eq:fcat}
\eeq
This is the unique nonzero eigenvalue, the null eigenspace being the space of dimension $K-1$ orthogonal to the eigenvector $\grad P(+|\mathbf{x})$. We  
call {\em principal discriminant direction} (PDD) at location $\mathbf{x}$, the (local) direction of the eigenvector $\grad P(+|\mathbf{x})$.

If we denote by $L(\mathbf{x})$ the log odds ratio,
\beq
L(\mathbf{x}) = \ln \frac{P(+|\mathbf{x})}{P(-|\mathbf{x})},
\eeq
we can also write 
\beq
P(\pm|\mathbf{x}) = \frac{1}{1+ \exp \mp L(\mathbf{x})},
\eeq
and we have
\beq
\mathbf{F}_{\text{cat}}(\mathbf{x}) = P(+|\mathbf{x})\;P(-|\mathbf{x})\;\grad L(\mathbf{x}) \grad L(\mathbf{x})^\mathsf{T}\,.
\eeq
The vector $\grad L(\mathbf{x})$, parallel to the vector $\grad P(+|\mathbf{x})$, is an eigenvector for the nonzero eigenvalue, and we have
\beq
f_{\text{cat}}(\mathbf{x}) = P(+|\mathbf{x})\;P(-|\mathbf{x})\, \|\grad L(\mathbf{x})\|^2.
\label{eq:fcat2}
\eeq
Note that the factor $P(+|\mathbf{x})\;P(-|\mathbf{x})$ can be written as
\beq 
P(+|\mathbf{x})\;P(-|\mathbf{x})=\frac{1}{4}(1-m(\mathbf{x})^2),
\eeq 
where $m(\mathbf{x})$ is the local difference between the posterior probabilities,
\beq
m(\mathbf{x}) \equiv P(+|\mathbf{x}) - P(-|\mathbf{x}) = 2 P(+|\mathbf{x}) -1.
\label{eq:bias}
\eeq
The class boundary is defined by the set of stimuli for which 
\beq
m(\mathbf{x}) =0,
\label{eq:Boundary1}
\eeq
or equivalently is given by the level set $\mathcal{L}_{0}$ of null log odds ratio:
\beq
\mathcal{L}_0 = \{ \mathbf{x}: L(\mathbf{x}) = 0\}.
\label{eq:Boundary2}
\eeq
If we consider the level set $\mathcal{L}_{\theta}$ of a given log odds ratio value $\theta$, that is
\beq
\mathcal{L}_{\theta} =\{ \mathbf{x}: L(\mathbf{x}) = \theta\},
\eeq
we get the important, yet expected, result that the principal discriminant direction, being given by $\grad L(\mathbf{x})$, is at each point orthogonal to the level set going through that point. For each location $\mathbf{x}$, there is thus a one-dimensional discriminant space, a (curved) line going through $\mathbf{x}$, with tangent vector $\grad L(\mathbf{x}')$ at every point $\mathbf{x}'$ along the line. We call {\em Principal Discriminant Curve} (PDC) the curve that at each point is tangent to the local PDD. We give numerical examples below.

We note that any PDC crossing the boundary is a possible 1D curve sufficient for performing the discrimination task. Along the curve, the information on the probability of belonging to a category is given by the length of the eigenvector. Any point $\mathbf{x}$ can be projected onto the chosen PDC in a way preserving the information on the membership probability, by following a curve remaining inside the level set to which the point $\mathbf{x}$ belongs to.

\subsubsection{Maxima of the categorical Fisher information}
\label{sec:maxfcat}

From the expression (\ref{eq:fcat2}) of the eigenvalue $f_{\text{cat}}(\mathbf{x})$, one sees that the categorical information is maximum on the boundary only if $\grad L(\mathbf{x})$ does not depend on the location $\mathbf{x}$. This is the case for two Gaussian categories with same covariance matrices. Otherwise, that is for different covariant matrices, the maximum of sensitivity does not coincide with the class boundary. We discuss these different cases in this section. 

We consider two equiprobable categories in dimension $K$. We want to know where the maximum of $f_{\text{cat}}(\mathbf{x})$ is located along a PDC as compared with the class boundary. Given a location $\mathbf{x}_0$, we can parametrize the PPD going through that point by
\beq
\frac{d\mathbf{x}(t)}{dt}= \pm \grad L(\mathbf{x}),
\eeq
with initial condition $\mathbf{x}(t=0)=\mathbf{x}_0$. 
In practice, the sign $\pm$ (independent of $\mathbf{x}$) is chosen so that the curve so generated crosses the category boundary. The extrema of $f_{\text{cat}}(\mathbf{x})$ along this curve satisfy
\beq
\frac{df_{\text{cat}}(\mathbf{x}(t))}{dt}=0,
\eeq
that is
\beq
\grad f_{\text{cat}}(\mathbf{x}) \cdot \grad L(\mathbf{x}) =0. 
\eeq
Now
\begin{eqnarray}
\grad f_{\text{cat}}(\mathbf{x}) &=&\grad L(\mathbf{x}) \, \frac{\partial}{\partial L} \left( \frac{1}{1+\exp L} \,\frac{1}{1+\exp - L} \right)\,\|\grad L(\mathbf{x})\,\|^2 \nonumber \\
&\;&\; +\;
 \frac{1}{1+\exp L} \,\frac{1}{1+\exp - L}\, \grad \|\grad L(\mathbf{x})\|^2,
\end{eqnarray} 
and
\beq 
\grad\|\grad L(\mathbf{x})\|^2\;=\;2 \mathbf{H}(\mathbf{x}) \grad L(\mathbf{x}),
\eeq 
where $\mathbf{H}$ is the Hessian matrix of $L$.
We have then
\beq 
\frac{1- \exp L(\mathbf{x})}{1+ \exp L(\mathbf{x})}\; \| \grad L(\mathbf{x})\|^4 \;+\; 2\,\grad L(\mathbf{x})^\mathsf{T} \mathbf{H}(\mathbf{x}) \grad L(\mathbf{x})\;=\;0.
\label{eq:maxfcat}
\eeq 
Note that $\grad L(\mathbf{x})=0$ gives a solution, but which corresponds to a minimum ($f_{\text{cat}}=0$). Hence for the maxima we search for solutions with $\grad L(\mathbf{x})\neq0$. 

The first term in the left-hand size of this equation is positive (negative) if $L(\mathbf{x})$ is negative (positive), that is if the category ``$-$'' (``$+$'') is the most probable at this location, and solutions can exist only if the second term is negative (positive). It is not clear if one can establish general and useful statements on the sign of this term. In the case of Gaussian categories, for which the Hessian is independent of the location, we consider below simple cases for which all the eigenvalues of the Hessian matrix have a same sign.

\subsection{The case of two Gaussian categories}
\label{sec:FcatGauss}

\subsubsection{Two Gaussian categories with the same covariance matrix}
\label{sec:FcatGauss_samecov}
We consider first the simple case of two Gaussian categories with the same covariance matrix, in a space of arbitrary $K$ dimensions:
\beq
P(\mathbf{x}|y)= \frac{1}{\sqrt{(2 \pi)^K \det \mathbf{\Sigma}}}\,\exp{-\textstyle{\frac{1}{2}} (\mathbf{x}-\mathbf{c}_{y})^\mathsf{T} \mathbf{\Sigma}^{-1} (\mathbf{x}-\mathbf{c}_{y})},
\label{eq:KdGsameC}
\eeq
with $y=\pm$, without loss of generality one can assume $\mathbf{c}_{\pm}=\pm\, \mathbf{c}$. We also assume a same frequency of occurrence, $P_{y} =1/2$.

Since the covariance matrices are identical, the quadratic terms in the probabilities of $\mathbf{x}$ given $y$ are identical, which leads to a simple equation for the boundary, Eq. (\ref{eq:Boundary1}). One gets:
\beq
\widetilde{\mathbf{c}} \cdot \mathbf{x} = 0,
\label{eq:ClassBound_KdsameG}
\eeq
where $\widetilde{\mathbf{c}}$ is the vector defined by
\beq
\widetilde{\mathbf{c}} = \mathbf{\Sigma}^{-1} \mathbf{c}.
\eeq
This is the equation of a hyperplane (a straight line in 2D) going through the origin, orthogonal to the direction of $\widetilde{\mathbf{c}}$. Note that the probability of, say, the category $+$, for equiprobable categories, is given by
\beq
P(+ | \mathbf{x})= 1 /(1 + \exp(-2\, \widetilde{\mathbf{c}} \cdot \mathbf{x})).
\eeq
The categorical Fisher information matrix takes the simple form
\beq
\mathbf{F}_{\text{cat}}(\mathbf{x}) = (1-m(\mathbf{x})^2)\, \widetilde{\mathbf{c}}\, \widetilde{\mathbf{c}}^\mathsf{T},
\label{eq:FcatGauss2cat}
\eeq
that is $[\mathbf{F}_{\text{cat}}(\mathbf{x})]_{i,j} = (1-m(\mathbf{x})^2)\, \widetilde{\mathbf{c}}_i \widetilde{\mathbf{c}}_j$. 
The nonzero eigenvalue is here 
\beq
f_{\text{cat}}(\mathbf{x}) = (1-m(\mathbf{x})^2)\, \|\widetilde{\mathbf{c}}\|^2.
\eeq
One sees that the categorical Fisher information is equal to $f_{\text{cat}}(\mathbf{x})$ along the direction parallel to the eigenvector $\widetilde{\mathbf{c}}$, and null along any orthogonal direction to this vector. In agreement with the general result shown above, the principal discriminant direction, $\widetilde{\mathbf{c}}$, is also the direction orthogonal to the boundary hyperplane. 

Since $1-m(\mathbf{x})^2$ is between $0$ and $1$, $f_{\text{cat}}$ is maximum at the boundary. The norm of $\widetilde{\mathbf{c}}$ is a measure of how much the two Gaussian distributions are well separated. In one dimension, $\widetilde{c}$ is a scalar, the distance between the means divided by the common standard deviation: it measures a global discriminability between the two categories. In psychophysics, the behavioral discriminability, $d'$, measures the ability to discriminate between $\mathbf{s}$ and $\mathbf{s}+\delta \mathbf{s}$ where $\delta \mathbf{s}$ results from a small modification $\delta x$ of a control parameter $x$ in the stimulus space. If this parameter corresponds to a relevant feature, efficient neural coding implies $d'=\delta x \sqrt{F_{\text{code}}(x)}$~\cite{seung1993simple}. Within our framework, efficient coding implies the matching of $F_{\text{code}}$ and $F_{\text{cat}}$, hence $d'$ is some monotonic increasing function of the product $(1-m(\mathbf{x})^2)\, \|\widetilde{\mathbf{c}}\|^2$, which is the product of a measure of how much one category is more probable than the other, by the global discriminability of the two distributions.

For this particular case of two Gaussian categories with the same variance, the principal discriminant direction is independent of the location, being everywhere given with the direction of the vector joining the centers of the categories. The principal discriminant curves are straight lines. An efficient learning could be obtained by a projection onto this direction.

\subsubsection{Two Gaussian categories with different covariance matrices}
\label{sec:FcatGauss_diffcov}
In the case of Gaussian categories with different covariance matrices, $\mathbf{\Sigma}_{\pm}$, the distribution of $\mathbf{x}$ given the category writes
\beq
P(\mathbf{x}|y)= \frac{1}{\sqrt{(2 \pi)^K \det \mathbf{\Sigma}_{y}}}\,\exp{-\textstyle{\frac{1}{2}} (\mathbf{x}-\mathbf{c}_{y})^\mathsf{T} \mathbf{\Sigma}_{y}^{-1} (\mathbf{x}-\mathbf{c}_{y})},
\label{eq:2dGdiffC}
\eeq
with $y=\pm$, $\mathbf{c}_{\pm}=\pm \mathbf{c}$. In that case the log odds ratio is
\beq
L(\mathbf{x}) = \frac{1}{2}\mathbf{x}^\mathsf{T} (\mathbf{\Sigma}_{-}^{-1}-\mathbf{\Sigma}_{+}^{-1}) \,\mathbf{x} 
\,+\, \mathbf{c}^\mathsf{T} (\mathbf{\Sigma}_{-}^{-1}+\mathbf{\Sigma}_{+}^{-1})\, \mathbf{x} 
\,+\, \frac{1}{2} \mathbf{c}^\mathsf{T}(\mathbf{\Sigma}_{-}^{-1}-\mathbf{\Sigma}_{+}^{-1}) \,\mathbf{c} \,+\, \frac{1}{2} \ln \frac{\det \mathbf{\Sigma}_{-}}{\det \mathbf{\Sigma}_{+}},
\label{eq:LGauss}
\eeq
and the class boundary is thus given by the quadratic manifold $ L(\mathbf{x})=0$.

The principal discriminant directions are obtained by taking the gradient of $L(\mathbf{x})$, that is
\beq
\grad L(\mathbf{x} ) \; = \; (\mathbf{\Sigma}_{-}^{-1}-\mathbf{\Sigma}_{+}^{-1})\, \mathbf{x} 
+\, (\mathbf{\Sigma}_{-}^{-1}+\mathbf{\Sigma}_{+}^{-1})\, \mathbf{c}.
\eeq
Below we give examples of the resulting Principal Discriminant Curves in the case of $K=2$ dimensions.

The location of the maximum of $f_{\text{cat}}(\mathbf{x})$ is different from that of the class boundary, since $\grad L(\mathbf{x} )$ is not constant. The Hessian matrix is independent of the location,
\beq
\mathbf{H} \; = \; \mathbf{\Sigma}_{-}^{-1}-\mathbf{\Sigma}_{+}^{-1}.
\eeq
If $\mathbf{\Sigma}_{+}$ is larger (smaller) than $\mathbf{\Sigma}_{-}$, that is if all the eigenvalues of $\mathbf{H}$ are positive (negative), then $ \grad L(\mathbf{x})^\mathsf{T} \mathbf{H} \grad L(\mathbf{x})$ is positive (negative), so that the maximum of the categorical Fisher information lies in the domain where the $+$ ($-$) category is the most probable. The simplest example is that of covariance matrices that are diagonalizable in a same basis with, on each eigen axis, e.g., the variance of category $+$ larger than the variance of category $-$ (or vice versa). In the general case, a sufficient condition for having either $\mathbf{H} \succeq 0$ or $\mathbf{H} \preceq 0$ is that the smallest eigenvalue of one of the covariance matrix is larger than the largest eigenvalue of the other covariance matrix (see Appendix \ref{app:maxfcat} for details). 
If not all eigenvalues of $\mathbf{H}$ have the same sign, it is not clear if a general statement can be given.

In the following subsection, we give numerical illustrations. In Appendix \ref{app:maxfcat}, we provide more details for distributions with different covariance matrices, together with additional numerical illustrations for the 1D (hence scalar) case and for the 2D case. The main qualitative results are that (i) the maximum is displaced in direction of the category with the largest variance, and (ii) for reasonably concentrated distributions, this location remains very close from the class boundary (see numerical illustrations and Appendix \ref{sec:maxfactapprox}). Otherwise, that is if, e.g., one of the distributions has a very large variance compared with the other one, the maximum of $f_{\text{cat}}$ can be far from the class boundary. 

\subsubsection{Numerical illustrations}
\label{sec:2cat_numeric}
\begin{figure}
\centering
\textbf{a.}\hspace{-0.1cm}
\includegraphics[height=4.7cm, valign=t]{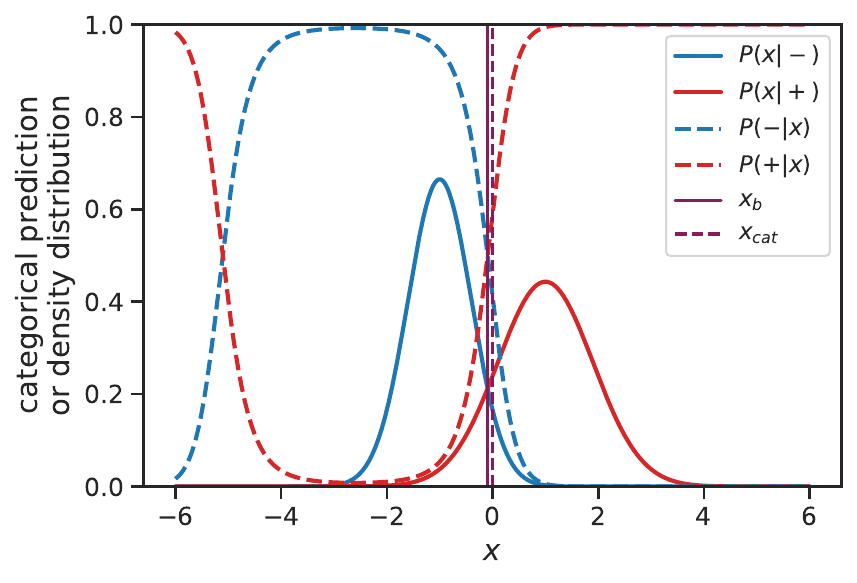}
\hfill
\textbf{b.}\hspace{-0.1cm}
\includegraphics[height=4.7cm, valign=t]{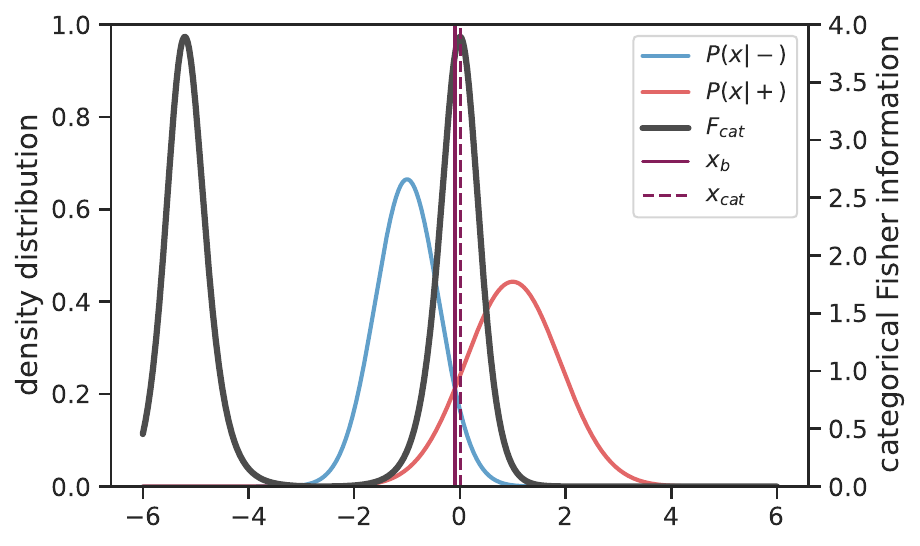}\\[5pt]
\textbf{c.}\hspace{-0.1cm}
\includegraphics[height=4.7cm, valign=t]{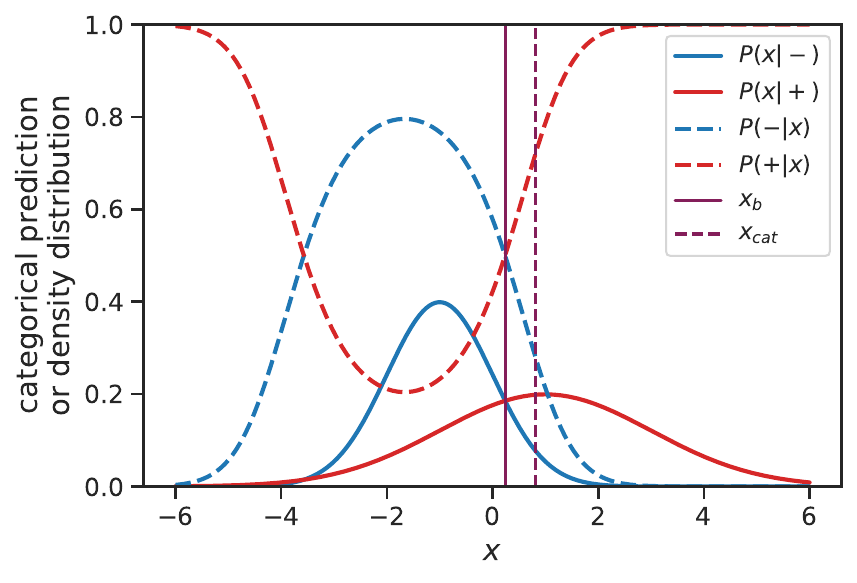}
\hfill
\textbf{d.}\hspace{-0.1cm}
\includegraphics[height=4.7cm, valign=t]{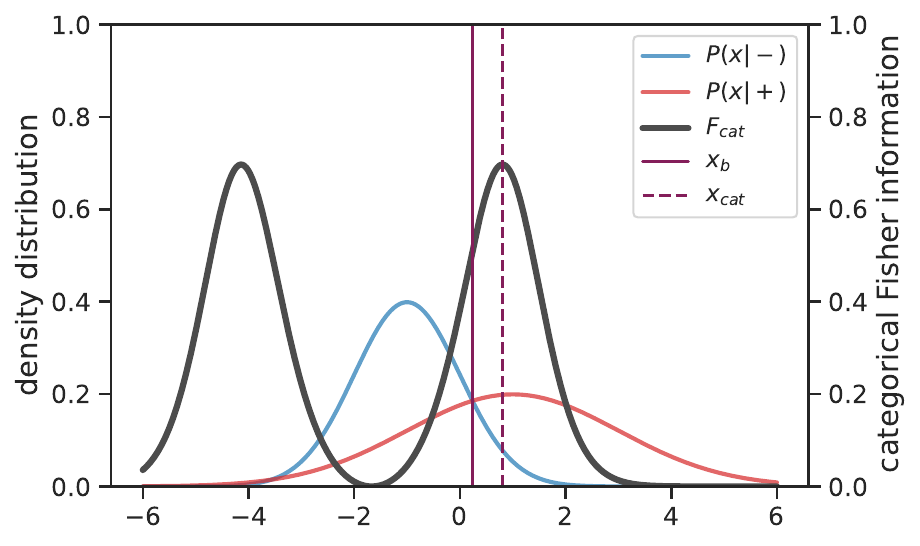}
\caption{\textbf{One-dimensional examples with two Gaussian categories: category boundary and categorical Fisher information}. Top, (a) and (b): $a=1.5$, $\sigma=0.6$. Bottom, (c) and (d): $a=2$, $\sigma=1$. 
Left, (a) and (c): Density distribution of the two classes, their corresponding posterior probabilities, along with the location of the boundary $x_b$ and the location of the relevant maximum of the categorical Fisher information $F_{\text{cat}}(x)$. 
Right, (b) and (d): Density distribution of the two classes and categorical Fisher information $F_{\text{cat}}(x)$.}
\label{fig:gaussian1d_xb_xcat_example}
\end{figure}

\begin{figure}
	\centering
	\includegraphics[width=0.5\linewidth]{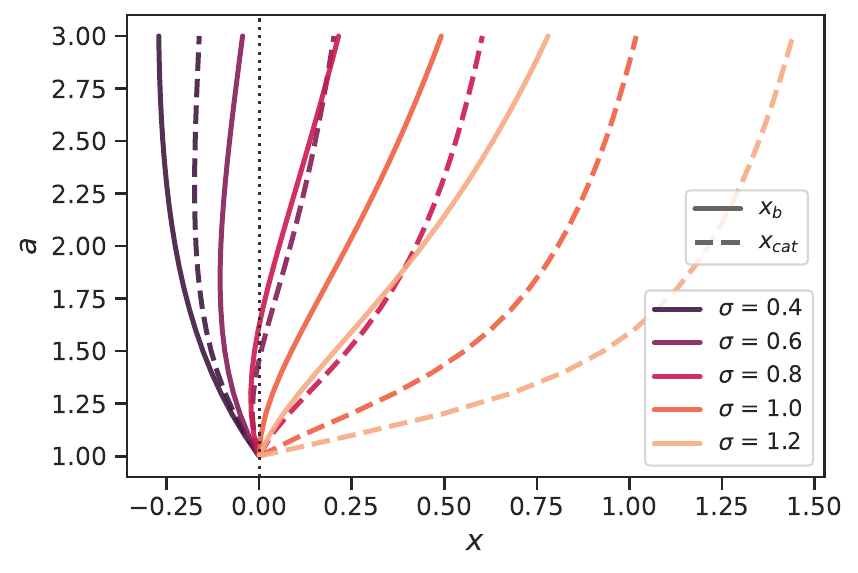}
	\caption{\textbf{One-dimensional example with two Gaussian categories: category boundary vs. argmax of categorical Fisher information}. Location $x_b$ of the boundary and location $x_\text{cat}$ of the relevant maximum of the categorical Fisher information for various values of $a$ and $\sigma$.}
	\label{fig:gaussian1d_xb_xcat}
\end{figure}

In Figs. \ref{fig:gaussian1d_xb_xcat_example} and \ref{fig:gaussian1d_xb_xcat} we present results for 1D-Gaussian categories. The $+$ and $-$ Gaussian distributions are centered at $\pm c$, $c=1$, with standard deviations $\sigma_{-}=\sigma$, $\sigma_{+}= a\,\sigma$, $a\geq 1$. We present results for different values of the parameters $\sigma, a$. In Appendix \ref{app:maxfcat} we derive the formulas giving the class boundary $x_b$ and the location $x_\text{cat}$ of the maximum of the categorical Fisher information. Figure~\ref{fig:gaussian1d_xb_xcat_example} presents the results for two particular parameter choices. Note that there are actually two class boundaries (and two associated maxima of $f_{\text{cat}}$), but only one matters, the other one being in a part of the space $x$ where there is essentially no data [the probability $P(x)$ is extremely small]. In Fig.~\ref{fig:gaussian1d_xb_xcat} we plot the locations of $x_\text{cat}$ and $x_b$ for various choices of $a$ and $\sigma$. When the two categories have the same variance ($a=1.0$), the boundary $x_b$ and the location $x_{\text{cat}}$ of the maximum of $f_{\text{cat}}$ are both $x=0$. But as $a$ increases, that is, as the relative width of the category on the right increases, these two quantities differ. Looking at the position of the curves relative to the line $x=0$, the behavior as $a$ increases is not intuitive. Actually, from the figure and from inspection of the formulas, one sees that: for large $\sigma$ values, both quantities always lie on the right side of $x=0$; for small $\sigma$ values, from $a=1$ the curves start on the left side, but eventually cross the $x=0$ line at a value of $a$ which is greater the smaller the variance -- for the smallest $\sigma$ values, this occurs outside the range of $a$ values shown in the figure. There is a range of intermediate $\sigma$ values for which at small $a$, the curve $x_b$ starts on the left side at small $a$ values whereas $x_\text{cat}$ still fully lies on the right side. Finally, note that at a given value of $a$ the difference between $x_b$ and $x_\text{cat}$ increases with $\sigma$.

\begin{figure}[htb]
\centering
\includegraphics[width=7cm,height=6.5cm,valign=t]{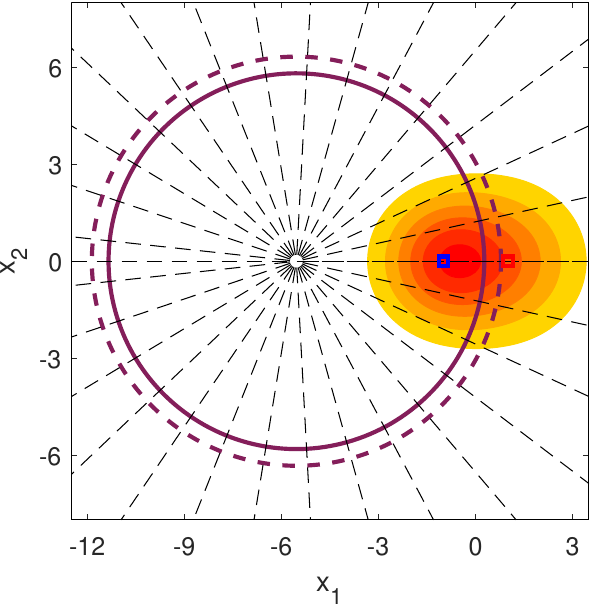}
\caption{\textbf{Two-dimensional example with two Gaussian categories: category boundary, maxima of the categorical Fisher information, and Principal discriminant curves}. 
For this simple example, the covariance matrices are $\mathbf{\Sigma}_{-}= \sigma^2 \,\mathbb{I}, \;\mathbf{\Sigma}_{+}= a^2 \,\mathbf{\Sigma}_{-}$, with $a=1.2,\; \sigma=1.3$. In the $(x_1, x_2)$ plane, the blue and red squares localize the centers of the two categories. The circle in continuous line gives the category boundary. The $-$ category is the most probable inside the red circle. The circle in dashed line gives the location of the maxima of the categorical Fisher information. The dashed black lines are principal discrimination curves. The color map gives the density distribution of $\mathbf{x}$.} 
\label{fig:gaussian2d_pdc}
\end{figure}

In Fig.~\ref{fig:gaussian2d_pdc} we give for $K=2$ an illustration of the case of Gaussian categories with diagonal covariance matrices. See Appendix \ref{app:diagcov} for the mathematical details. The two concentric circles are the class boundary (continuous line) and the location of the maxima of the categorical information (dashed line). We show a sample of Principal Discrimination Curves (which are here the rays of these circles). We also represent the density distribution of $\mathbf{x}=(x_1, x_2)$, showing that actually only a small part of the plane is relevant for the categorization task. In Appendix \ref{app:diagcov}, we show in Fig.~\ref{fig:ppdcurves} other 2D-examples of PDCs, for cases where these are not straight but curved lines.

\subsection{Multiclass \texorpdfstring{($M > 2$)}{M>2} case}
\subsubsection{Discriminant directions: nonzero eigenvalues}
The result on the number of nonzero eigenvalues, seen above Sec. \ref{sec:pdc}, generalizes to an arbitrary number $M$ of categories in dimension $K$ in the following way. If $K < M-1$, the categorical Fisher information matrix has, obviously, at most $K$ nonzero eigenvalue. If $K \geq M-1$, the categorical Fisher information matrix has (at most) $M-1$ nonzero eigenvalues. The proof is as follows. 

We look for the eigenvectors of $\mathbf{F}_{\text{cat}}(\mathbf{x})$. Let $\mathbf{u}$ be an eigenvector for the eigenvalue $f$, that is
\beq
 \mathbf{F}_{\text{cat}}(\mathbf{x}) \cdot \mathbf{u} = f \, \mathbf{u}.
\eeq
One can write this equation as:
\beq
\text{for any}\;i \in \{1,\ldots,K\},\;\;\left[\sum_{y=1}^{M}\,\frac{\partial_i P(y|\mathbf{x})}{P(y|\mathbf{x})}\; \grad P(y|\mathbf{x}) \right] \cdot \mathbf{u} = f \,u_i.
\label{eq:eigen}
\eeq
Since $\textstyle{\sum}_{y} P(y|\mathbf{x})=1 $, $\textstyle{\sum}_{y=1}^{M} \grad P(y|\mathbf{x})=0$, the $M$ vectors $\grad P(y|\mathbf{x})$ are not linearly independent, they span a space of dimension at most $M-1$. The $K$ linear combinations of these vectors, $\textstyle{\sum}_{y=1}^{M}\,\frac{\partial_i P(y|\mathbf{x})}{P(y|\mathbf{x})} \grad P(y|\mathbf{x}), i=1,...,K$, belong to this space. Hence any $\mathbf{u}$ orthogonal to this space gives a null value. We have thus (at most) $M-1$ nonzero eigenvalues, associated with eigenvectors which we called the principal discriminant directions (PPD). 

\subsubsection{General expectations}
From the above analytical and numerical results we get the following general picture: In $K$ dimension with $M$ categories, the categorical Fisher information, at any location $\mathbf{x}$, has at most a number of $\min\{M-1,K\}$ nonzero eigenvalues. On the boundary between two categories (far from other categories), the direction associated with the largest eigenvalue is orthogonal to the boundary, and points toward this boundary for locations near the boundary. One can define a principal discriminant curve which, at each location, is tangent to the principal eigenvector, and cross the boundary. The other eigenvectors with nonzero eigenvalues define a hyperplane orthogonal to the principal direction, hence tangent to the boundary for locations on the boundary. We provide a numerical illustration with three categories in Sec. \ref{sec:2d_gauss}, comparing the categorical Fisher information with the neural Fisher information after learning.

\section{Neural geometry: Numerical illustrations} 
\label{sec:illustrations}

In this section, we study numerically the neural geometry underlying categorical perception induced by learning in artificial feedforward networks. We go beyond the numerical analysis we did in the related work Ref.~\citep{LBG_JPN_CPDeepLearning_2022}, making precise links with the analysis in the previous sections. In addition, in Ref.~\citep{LBG_JPN_CPDeepLearning_2022} we used a proxy for the Fisher information, since it is difficult to compute in deep networks. Here, in one of the examples we discuss, we can numerically exactly compute the neural Fisher information. The analysis actually validates {\em a posteriori} the use of the proxy considered in Ref.~\citep{LBG_JPN_CPDeepLearning_2022}.

We consider both simple Gaussian categories in 2D, and numerical experiments on the MNIST database with networks of various number of hidden layers. In all cases, in order to test the formal analysis, we consider the neural geometry near the boundary between pairs of categories, or along a path crossing the boundary between two categories, whatever the total number of categories learned by the network (three in the case of the simple Gaussian example, ten in the case of the MNIST database).

\subsection{Two-dimensional example with Gaussian categories}
\label{sec:2d_gauss}

\begin{figure}
\centering
\textbf{a.}\hspace{-0.3cm}
\includegraphics[width=0.47\linewidth,valign=t]{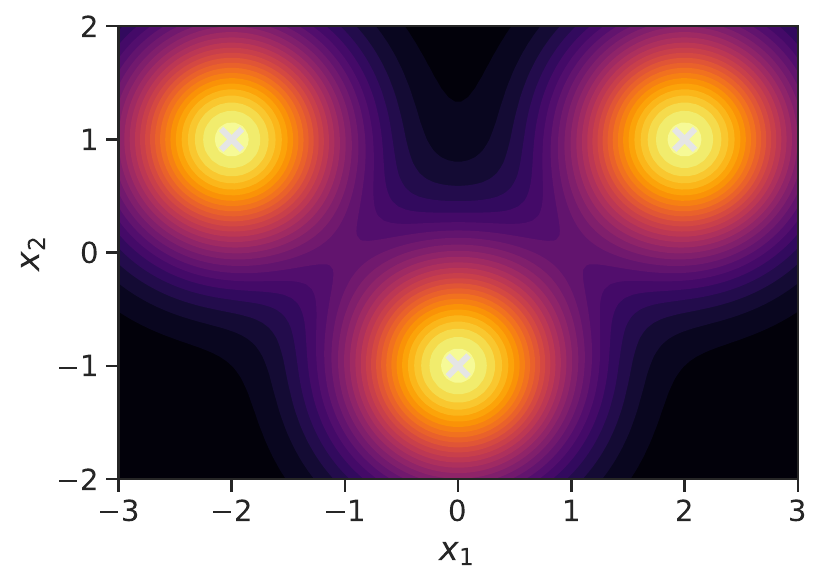}
\hfill
\textbf{b.}\hspace{-0.3cm}
\includegraphics[width=0.47\linewidth,valign=t]{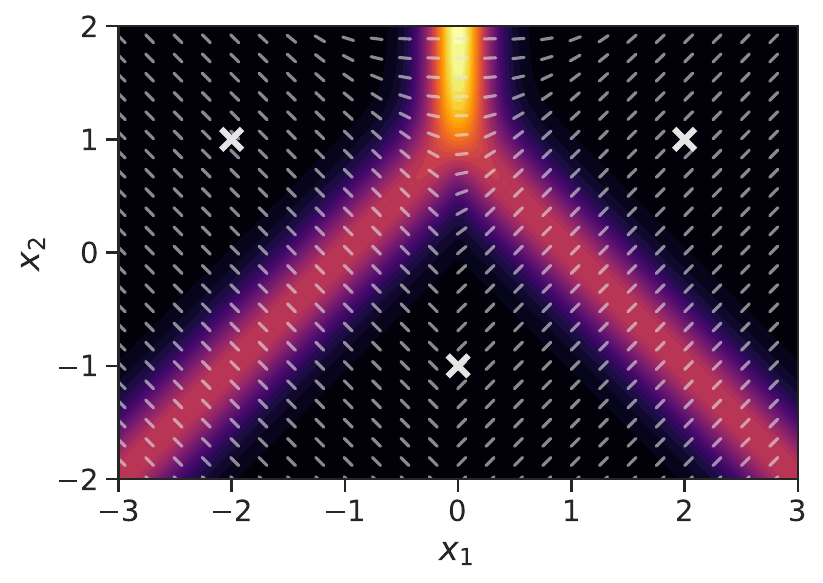}\\
\textbf{c.}\hspace{-0.3cm}
\includegraphics[width=0.47\linewidth,valign=t]{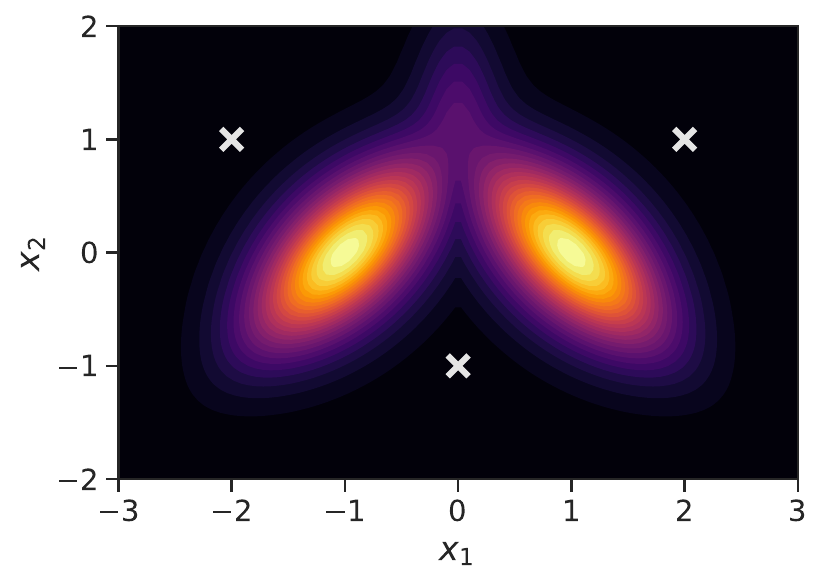}
\hfill
\textbf{d.}\hspace{-0.3cm}
\includegraphics[width=0.47\linewidth,valign=t]{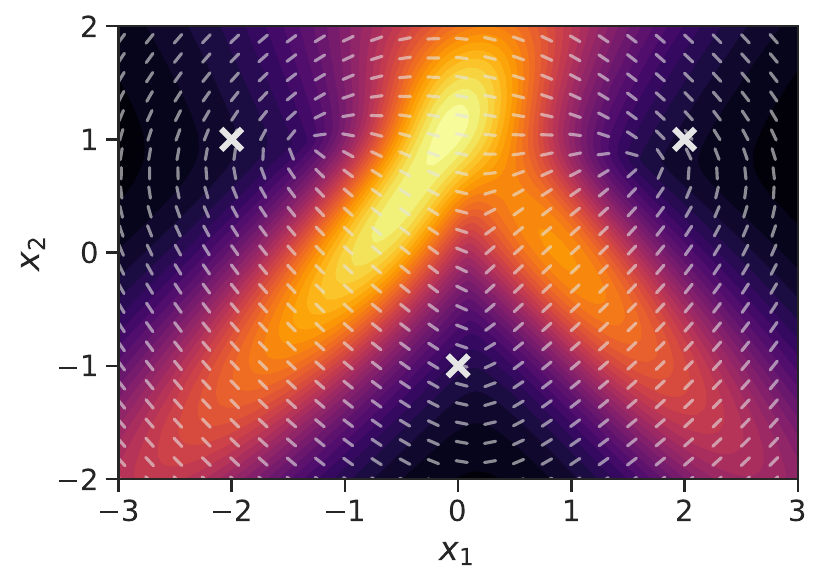}
\caption{\textbf{Two-dimensional example with three Gaussian categories: Fisher information quantities}. 
(a) Probability $P(\mathbf{x})$
(b) Visualization of the categorical Fisher information matrix $\mathbf{F}_{\text{cat}}(\mathbf{x})$ at each point on the $\mathbf{x} = (x_1, x_2)$ plane. The small line represents the direction at this point of the eigenvector of the Fisher information matrix associated with the largest eigenvalue $f_{\text{cat}}(\mathbf{x})$. The magnitude of this largest eigenvalue is represented by the color, the lighter the greater.
(c) The quantity $P(\mathbf{x}) f_{\text{cat}}(\mathbf{x})$, quantifying the source of the potential classification errors in the $\mathbf{x}$-plane.
(d) Visualization of the neural Fisher information matrix $\mathbf{F}_{\text{code}}(\mathbf{x})$, at each point on the $(x_1, x_2)$ plane, after learning. The graphic convention is the same as in (b).}
\label{fig:gaussian2d}
\end{figure}

\begin{figure}
	\centering
	\textbf{a.}\hspace{-0.3cm}
	\includegraphics[width=0.445\linewidth,valign=t]{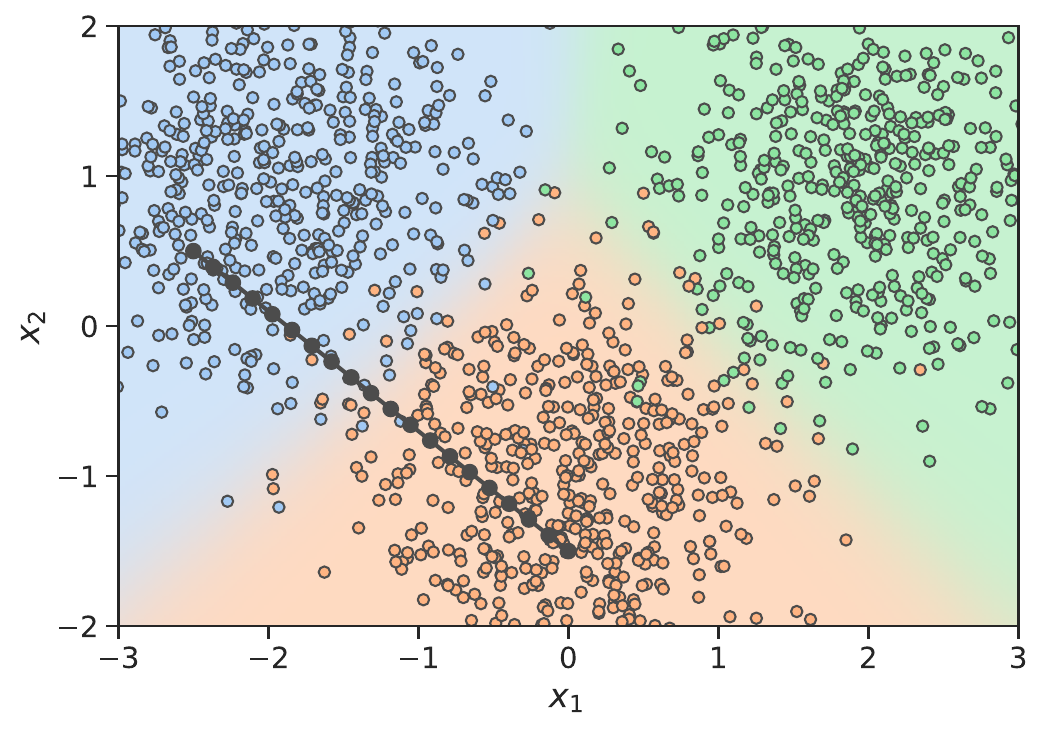}
	\hfill
	\textbf{b.}
	\includegraphics[width=0.493\linewidth,valign=t]{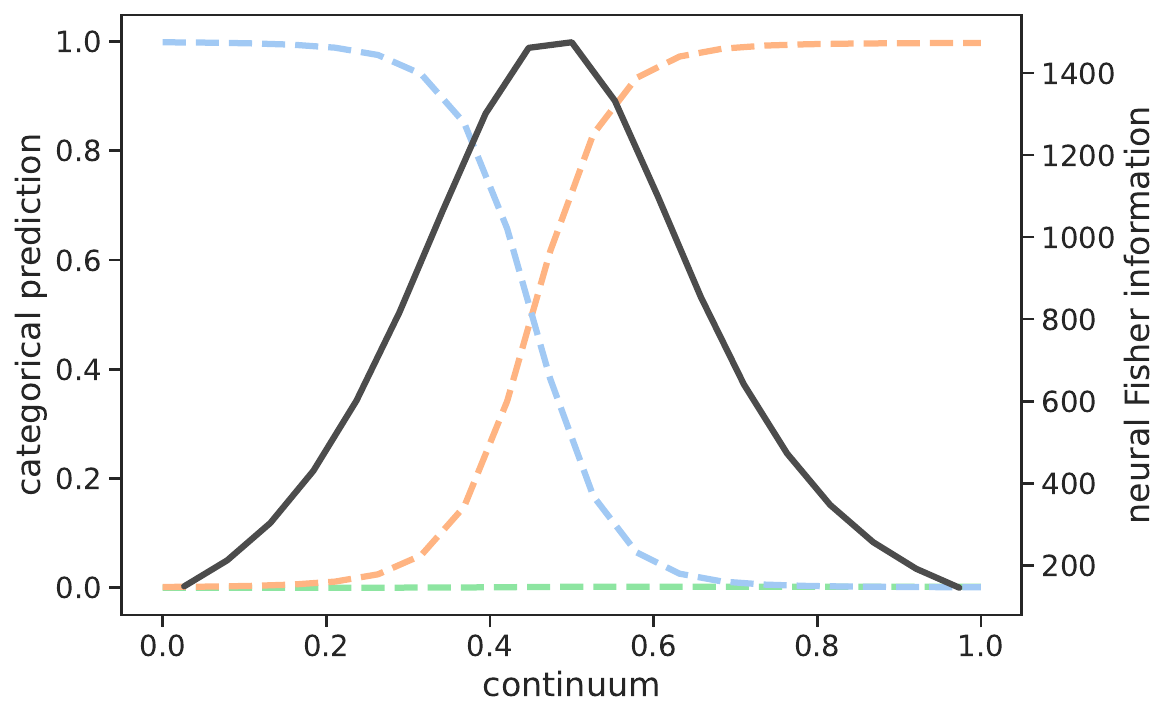}
	\caption{\textbf{Two-dimensional example with three Gaussian categories: Fisher information along a one-diemnsional path}. 
		(a) Colored dots: training set, random samples from each of the categories. Background color: mix between the colors that correspond to each of three categories, proportionally to the posterior probabilities $P(y|\mathbf{x})$ as estimated by the neural network. Dark dots: a path interpolating between two samples from the blue and the red categories. 
		(b) The dashed colored lines indicate the posterior probabilities, as found by the network, each color representing its respective category. The solid line is the scalar Fisher information along the one-dimensional path shown in (a).}
	\label{fig:gaussian2d_path}
\end{figure}

We consider the case of three Gaussian categories in 2D; see Figs.~\ref{fig:gaussian2d}(a) and~\ref{fig:gaussian2d_path}(a). The neural network is a multilayer perceptron with two hidden layers of 32 cells with sigmoidal activation. In the last hidden layer, each cell $i$ has a noisy neural activity given by $r_i(\mathbf{x}) = f_i(\mathbf{x}) + \sigma \sqrt{g_i(\mathbf{x})} z_i$, where $f_i$ is a sigmoidal activation function, $z_i$ is a normal unit random variable, and $\sigma=0.3$. Here we take $g_i(\mathbf{x}) = f_i(\mathbf{x})$. In the context of machine learning, this neural noise may be correlated with the one injected during learning under the name of dropout~\citep{srivastava2014dropout}, a commonly used heuristic aimed at improving learning efficiency. In the original work \citep{srivastava2014dropout}, dropout consists of multiplicative noise (in each layer) in the form of Bernoulli or Gaussian noise, with $g_i(\mathbf{x}) = f_i(\mathbf{x})^2$. Other types of noise distribution can be considered. Our choice here, $g_i(\mathbf{x}) = f_i(\mathbf{x})$, yields a Poisson-like noise, as commonly found in biological neural networks (see, e.g., Refs.~\citep{tolhurst1983statistical} and \citep{softky1993highly}). We assume that the noise is not correlated between neurons given a stimulus $\mathbf{x}$, which means that we can write $P(r|\mathbf{x}) = \prod P(r_i|\mathbf{x})$, which in turn implies that the Fisher information can be written as $\mathbf{F}_{\text{code}}(\mathbf{x}) = \textstyle{\sum}_i \mathbf{F}_{\text{code}, i}(\mathbf{x})$, where $\mathbf{F}_{\text{code}, i}(\mathbf{x})$ is the Fisher information of neuron $i$.

Figure~\ref{fig:gaussian2d}(b) proposes a representation of the $2\times2$ Fisher information matrix $\mathbf{F}_{\text{cat}}(\mathbf{x})$ at each point on the $\mathbf{x}=(x_1, x_2)$ plane. As expected from our analysis, the largest associated eigenvalue is strongest at the boundary between categories, with the associated eigenvector being orthogonal to the boundary. Figure~\ref{fig:gaussian2d_si} in Appendix~\ref{app:2d_fcat_fcod} shows the same representation but for the second eigenvalue (the smallest one). One can see that it is very small everywhere compared with the first eigenvalue, except at the location where the three categories overlap a little bit more. In practice, however, the important part of the space is where the quantity $P(\mathbf{x}) f_{\text{cat}}(\mathbf{x})$ is important, as can be seen from the asymptotic expressions, Eqs.~(\ref{eq:Delta_general_1d}) and ~(\ref{eq:Delta_general}), which relates the mutual information to the Fisher information. This can be visualized in Fig.~\ref{fig:gaussian2d}(c): the salient regions are the boundaries between categories where something can happen
(\textit{i.e.}, a region with a nonzero probability).

After learning, as expected, the network has learned to estimate the posterior probabilities $P(y|\mathbf{x})$, correctly partitioning the three categories into their respective regions, see Fig.~\ref{fig:gaussian2d_path}(a). Regarding the matching between the categorical and neural Fisher information quantities, we can see in Fig.~\ref{fig:gaussian2d}(d) that, after learning, the Fisher information matrix $\mathbf{F}_{\text{code}}(\mathbf{x})$ qualitatively follows $\mathbf{F}_{\text{cat}}(\mathbf{x})$: the largest eigenvalue is the greatest at the boundary between categories, illustrating the categorical perception phenomenon. Furthermore, at each point on a boundary, the eigenvector associated with the largest eigenvalue is orthogonal to the class boundary, and points toward the boundary at a location away from it. Figure~\ref{fig:gaussian2d_si} in Appendix~\ref{app:2d_fcat_fcod} shows how, during the course of training, the second eigenvalue gets smaller and smaller compared with the first eigenvalue, except again at the overlapping location of the three categories, which aligns with the local dimensionality of the categorical Fisher information.

Finally, we consider a one-dimensional path in input space, for which the Fisher information quantities are nonzero, depicted by the dark dots in Fig.~\ref{fig:gaussian2d_path}(a), interpolating between two items drawn from two different categories. In doing so, we mimic the use of morphed continua in psychophysics and cognitive neuroscience experiments. We compute the (scalar) Fisher information of the neural code along this line. We show the results in Fig.~\ref{fig:gaussian2d_path}(b) together with the categorical prediction output by the neural network. As expected, the neural Fisher information is the greatest at the boundary between categories.

\subsection{Images of handwritten digits}
\label{sec:mnist}
\begin{figure}
	\centering
	\textbf{a.}\hspace{-0.3cm}
	\includegraphics[width=0.47\linewidth,valign=t]{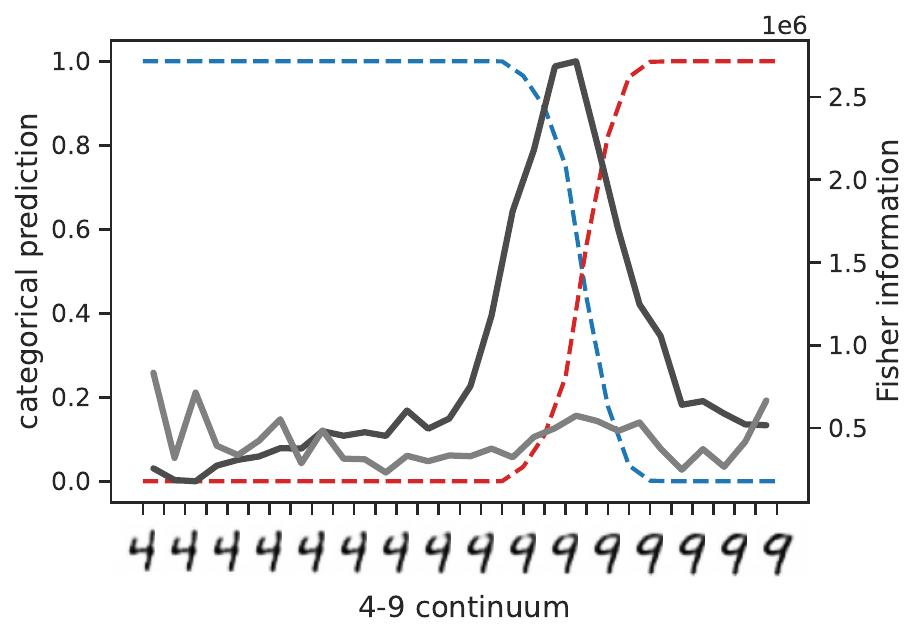}
	\hfill
	\textbf{b.}\hspace{-0.3cm}
	\includegraphics[width=0.47\linewidth,valign=t]{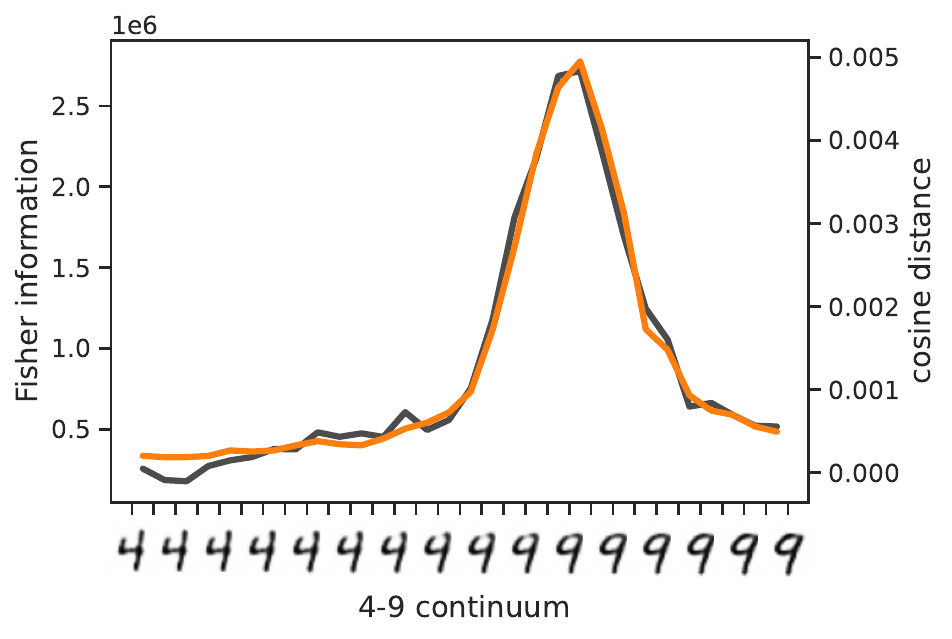}
	\caption{\textbf{Categorical perception along a 4 to 9 continuum}. (a) Scalar neural Fisher information $F_{\text{code}}$ along the 4-9 continuum (average over 10 training runs of the model), before (light gray) and after (dark gray) training. The dashed colored lines indicate the posterior probabilities, as found by the network, blue corresponding to category 4 and red to category 9. (b) Comparison between Fisher information (dark gray, left y axis) and cosine distance (orange, right y axis) between neural activities evoked by contiguous items along the continuum.}
	\label{fig:mnist}
\end{figure}

Here we consider the MNIST dataset \citep{lecun1998gradient}, a dataset of $28\times28$ handwritten digits (hence, the stimulus $\mathbf{s}$ lives in a 784 dimensional space). The neural network is a multilayer perceptron with two hidden layers, each made of 256 cells with ReLU activation. Poisson like neuronal noise affects the last hidden layer, just as in the previous example, with $\sigma=0.1$. The neural network is trained on the full MNIST training set. A continuum between two images taken from the MNIST test set is created by interpolating between them in a latent space discovered by training an autoencoder to reconstruct digits from the MNIST training set, as done in Ref.~\citep{LBG_JPN_CPDeepLearning_2022}. Here, we consider a continuum between an item from the 4 category and an item from the 9 category (two categories that are among the most confusable ones). 
Each image along the continuum lies in the relevant manifold of digits. The labels on the abscissa of Fig.~\ref{fig:mnist}(a) pictures a few samples from the continuum, which is made of 31 images.

This continuum can be viewed as a 1D ``$x$'' in the previous discussions. One can then compute the categorical predictions outputted by the neural network together with the scalar Fisher information of the last hidden layer of neurons. Once again, Fig.~\ref{fig:mnist}(a) shows that learning induces categorical perception, with larger Fisher information at the boundary between the two categories. In our previous work \citep{LBG_JPN_CPDeepLearning_2022}, the cosine distance between the neural activities $\mathbf{r}(x)$ and $\mathbf{r}(x+\delta x)$ was used as a proxy for Fisher information $F_{\text{code}}(x)$, because it is much easier to compute. Figure~\ref{fig:mnist}(b) shows that these two quantities indeed behave quite similarly -- actually these two quantities appear to be quantitatively almost the same up to a global scale factor (the alignment is here performed by minimizing the mean absolute error between a linear transformation of the cosine distance and the neural Fisher information).

In Appendix~\ref{app:additionalMNIST}, Fig.~\ref{fig:mnist_extra} reproduces the results presented in Fig.~\ref{fig:mnist} with the same neural network probed on another continuum, going from category 1 to category 7. 
This supplementary figure also plots the tuning curves of an arbitrarily chosen set of neurons in the last hidden layer. Following the empirical approach of neuroscience, these tuning curves are defined as the mean response of the neurons to the images along the continuum. First, we see that many neurons have a smooth tuning curve along the continuum, despite having a ReLU activation function. Second, as expected from our analysis, the steepest slopes of these tuning curves are roughly located in the transition region between categories. This is what, collectively, results in a greater neural Fisher information at this location. We also note that some neurons do not activate at all in this part of the input space. Finally, Fig.~\ref{fig:mnist_extra_deeper} replicates all these findings but considering a deeper multilayer perceptron with four hidden layers.

\section{Discussion}

\label{sec:Discussion} 
In this paper, for the study of artificial neural networks performing a categorization task, we extend and develop a Bayesian and information-theoretic approach we initially introduced in the context of computational neuroscience. We are thus making use of methods and results obtained in neuroscience to open artificial networks ``black boxes''. The Bayes cost that we consider is the average, over the data distribution, of the entropy loss commonly used in machine learning. A formal analysis gives two interesting decompositions of this Bayes cost. One shows that one can separately deal with the neural coding and decoding tasks. The other one is a bias-variance type decomposition -- but not related to the sources of errors that would result from the learning of a finite number of examples. We show that minimizing the coding cost notably implies maximizing the mutual information between category membership and neural activity. 

Within this general setting, we consider structured data, characterized by an underlying feature space of dimension much smaller that the one of the coding layer. We derive in that case an asymptotic formulas for the mutual information between the neural activity and its underlying feature space. It allows to make explicit the two tasks jointly solved through learning: (i) finding an appropriate projection (feature) space, and, (ii) building a projection with the appropriate metrics on this space. This metrics is characterized by the matching of two Fisher information matrices. One, the categorical Fisher information, characterizes the geometry of the categories in the feature space. The other one, the neural Fisher information, characterizes the sensitivity of the neural activity to change in the feature space. The matching of these two Fisher information matrices results in a magnification of the space near category boundaries, characteristic of the categorical perception effect. We make more precise this statement, thanks to a detailed analysis of the properties of the categorical Fisher information. We show the non intuitive result that the largest expansion of the neural space is not necessarily exactly at, although very near, the class boundaries. Our predictions about the categorical perception phenomenon are well supported by the various numerical results presented in our related paper~\citep{LBG_JPN_CPDeepLearning_2022}, an empirical work that present results based on a great diversity of architectures and datasets, including both multilayer perceptrons and convolutional neural networks of various depths, many different continua tested in the case of MNIST, and a different dataset with complex images involving a cat/dog classification. In the present paper, working with both toy examples and the MNIST handwritten digits dataset, we present new simulations that make precise links with the analytical results. In particular, we illustrate how, after learning, the two Fisher information matrices essentially align with the boundaries between categories.

Future works should address several issues. On the theoretical side, our main analytical result for the mutual information is based on restrictive hypotheses. However, the predictions that results from its optimization, and the numerical simulations, strongly suggest a wider range of validity. It would be interesting to further explore its domain of validity or at least to get exact bounds on the mean Bayes cost -- and in this paper we provide several analysis going in this direction, notably by considering bounds on the Jensen gap appearing in the bias-variance decomposition of the cost. Another essential point is that our results are based on the use of the exact probability density function of the data. Obviously, they should be reconsidered in the context of learning with a finite set of examples. Note however that the numerical illustrations clearly indicate that the main results hold in such a learning context. 

In the neuroscience context --but also in the machine learning context-- one should study the effect of (possibly strong) noise at any stage of processing, also implying noise correlations in the subsequent layers. For this, one issue is to numerically estimate the neural Fisher information quantity. Here, in our numerical illustrations, for uncorrelated noise the cosine distance between neural activities appears to be a remarkably good proxy for the Fisher information. To see this, the latter is computed numerically exactly, taking advantage of the decomposition of the Fisher information in a sum of separate contributions from each neuron. But such decomposition does not exist in the case of correlations, making difficult the computation of the Fisher information -- hence also difficult to test the validity of any proxy easier to compute. Another related issue is to understand the effect of noise correlations on the geometry of the neural space --e.g., in the spirit of Ref.~\citep{franke2016structures}, but for the case of category learning.

\section*{Acknowledgments} 
We are grateful to the two anonymous reviewers for their constructive comments that helped improving the manuscript. A short version of this work in a preliminary stage was selected for an Oral presentation at the Information-Theoretic Principles in Cognitive Systems Workshop at the 37th Conference on Neural Information Processing Systems (NeurIPS 2023). We also thank the two reviewers of this workshop for their valuable comments.

\appendix 

\renewcommand{\theequation}{\thesection\arabic{equation}}

\section*{Appendices} 

\section{Bias-variance decomposition of the mean Bayes cost}
\label{app:biasvar}
\setcounter{equation}{0}
In this appendix we extend the analysis based on the decomposition of the cost seen Sec. \ref{sec:biasvar}, Eqs. (\ref{eq:cbarx3-a_main}), (\ref{eq:cbarx3-b_main}), (\ref{eq:cbarx3-c_main}), that is,
\begin{eqnarray}
\mathcal{\overline{C}} &=& 
\textstyle{\int}\, D_{\text{KL}}(P(Y|\mathbf{x}^*)\|P(Y|\mathbf{x})) \;P(\mathbf{x}^*,\mathbf{x}) \,d^{K^*}\mathbf{x}^*\,d^K\mathbf{x}
\label{eq:cbarx3-a_app} \\
& + & 
\textstyle{\int}\, D_{\text{KL}}(P(Y|\mathbf{x})\|\overline{g}(Y|\mathbf{x})\,)
\,P(\mathbf{x})\,d^K\mathbf{x} 
\label{eq:cbarx3-b_app} \\
& + & \textstyle{\int} \textstyle{\sum}_{y=1}^{M} P(y|\mathbf{x}) \left\{ \ln \overline{g_{y}}(\mathbf{x}) - \textstyle{\int} \ln g_{y}(\mathbf{r})\,P(\mathbf{r}|\mathbf{x})\,d^{N}\mathbf{r} \right\} P(\mathbf{x})d^K\mathbf{x}.
\label{eq:cbarx3-c_app}
\end{eqnarray}
First, Sec. \ref{app:biasvarefficient}, we relate this decomposition to the classical bias-variance trade-off for quadratic error loss functions by considering the vicinity of an efficient estimator. In addition, we use this analysis to derive bounds on the mean Bayes cost, making a connection with the asymptotic formula we obtained for the mutual information. Then, Sec. \ref{app:jensengap}, we derive bounds for the variance part (\ref{eq:cbarx3-c_app}), making use of relationships between the Jensen gap and the variance.

\subsection{Vicinity of an efficient estimator}
\label{app:biasvarefficient}
An efficient estimator has no bias and a variance as small as possible (saturating the Cramér-Rao bound). We consider the case where the estimator is close to be efficient: small bias and small variance, assuming that the variance can be small. 

\subsubsection{If the bias is small} 
The bias $b(y|\mathbf{x})$ is defined by
\beq
\overline{g_{y}}(\mathbf{x})= P(y|\mathbf{x})+b(y|\mathbf{x}).
\eeq
It satisfies $\textstyle{\sum}_{y} b(y|\mathbf{x})= 0$. If the bias is small, we expand the term (\ref{eq:cbarx3-b_app}) in the mean cost:
\begin{eqnarray}
D_{\text{KL}}(P(Y|\mathbf{x})\|\overline{g}(Y|\mathbf{x})) &=& \sum_{y} P(y|\mathbf{x}) \left(-\frac{b(y|\mathbf{x})}{P(y|\mathbf{x})} + \frac{b(y|\mathbf{x})^2}{2\,P(y|\mathbf{x})^2} \right) \nonumber \\
&=& \frac{1}{2} \, \sum_{y} \frac{b(y|\mathbf{x})^2}{P(y|\mathbf{x})},
\end{eqnarray}
that is
\beqa
D_{\text{KL}}(P(Y|\mathbf{x})\|\overline{g}(Y|\mathbf{x})) &=& \frac{1}{2} \, \sum_{y} P(y|\mathbf{x})\, \left(\frac{\overline{g_{y}}(\mathbf{x})-P(y|\mathbf{x})}{P(y|\mathbf{x})} \right)^2 \nonumber \\
&(+& \mbox{higher order terms}),
\label{eq:expansion-bias}
\eeqa
which is thus a (normalized) standard quadratic bias term. 

\subsubsection{If the variance is small}
We expand the last term, (\ref{eq:cbarx3-c_app}), assuming that the variance is small. As was done in Ref.~\cite{LBG_JPN_2012}, for the typical values of r given a stimulus x, we write that $g_{y}(\mathbf{r})$ is a good approximation of $\overline{g_{y}}$:
\begin{eqnarray}
\ln \frac{\overline{g_{y}}(\mathbf{x})}{g_{y}(\mathbf{r})} &=& - \ln\left( 1+\frac{g_{y}(\mathbf{r})-\overline{g_{y}}(\mathbf{x})}{\overline{g_{y}}(\mathbf{x})} \right) \\
&=& - \frac{g_{y}(\mathbf{r})-\overline{g_{y}}(\mathbf{x})}{\overline{g_{y}}(\mathbf{x})} \;+\;\frac{1}{2} \left( \frac{(g_{y}(\mathbf{r})-\overline{g_{y}}(\mathbf{x}))^2}{\overline{g_{y}}(\mathbf{x})^2} \right)\nonumber \\
 &( +& \mbox{higher order terms}).
 \label{eq:smallvar}
\end{eqnarray}
Performing the integral over $\mathbf{r}$, the first term in (\ref{eq:smallvar}) gives zero (by definition of $\overline{g_{y}}$), and one gets
\beqa
\int \, \ln \frac{\overline{g_{y}}(\mathbf{x})}{P(y|\mathbf{r})} \,P(\mathbf{r}|\mathbf{x})\,d^{N}\mathbf{r} &=& \frac{1}{2} \, \int \, \frac{(g_{y}(\mathbf{r})-\overline{g_{y}}(\mathbf{x}))^2}{\overline{g_{y}}(\mathbf{x})^2} \,P(\mathbf{r}|\mathbf{x})\, d^{N}\mathbf{r} \; \nonumber \\ 
&(\,+ &\;\mbox{higher order terms}\,).
\label{eq:expansion-var}
\eeqa
Thus this variance part of the cost precisely reduces to the variance of the estimator, normalized by the (square of the) mean.

\subsubsection{Bounding the cost}
\label{app:bound}
From the above expansions it is tempting to try to get a bound on the cost. At the order of the expansion (\ref{eq:expansion-var}), making use of the Cramér-Rao bound, for the 1D case ($K=1$), we have
\beq
	\int \, \ln \frac{\overline{g_{y}}(x)}{g_{y}(\mathbf{r})} \,P(\mathbf{r}|x)\, d^{N}\mathbf{r} \;\geq \;\frac{1}{2\overline{g_{y}}(x)^2}
	\frac{1}{F_{\text{code}}(x)} \left( \frac{d}{dx} \overline{g_{y}}(x)\right)^2,
	\label{eq:inequ_1}
\eeq
and thus, at this order,
\beq
	\mathcal{\overline{C}}(x) \,\geq\, D_{\text{KL}}(P(Y|x)\|\overline{g}(Y|x)) \;+\;\frac{1}{2} \;\frac{\widetilde{F_{\text{cat}}}(x) }{F_{\text{code}}(x)},
\eeq
with
\beq
\widetilde{F_{\text{cat}}}(x) \equiv \sum_{y=1}^{M} P(y|x) \;
	\left( \frac{d}{dx\,}\, \ln \overline{g_{y}}(x) \right)^2.
\eeq
Hence
\beq
	\mathcal{\overline{C}} \,\geq\, \int\, D_{\text{KL}}(P(Y|x)\|\overline{g}(Y|x)) \; P(x)\,dx\;+\;\frac{1}{2} \int\;\frac{\widetilde{F_{\text{cat}}}(x) }{F_{\text{code}}(x)}\,\; P(x)\,dx.
\eeq

In the limit of zero bias, the KL divergence goes to zero, $\widetilde{F_{\text{cat}}}(x)=F_{\text{cat}}(x)$, and thus
\beq
\mathcal{\overline{C}} \; \geq \; \frac{1}{2}\int \frac{F_{\text{cat}}(x)}{F_{\text{code}}(x)}\; P(x)\,dx.
\label{eq:ineq}
\eeq
In the multidimensional case ($K>1$), the inequality (\ref{eq:inequ_1}) becomes
\beq
	\int \, \ln \frac{\overline{g_{y}}(\mathbf{x})}{g_{y}(\mathbf{r})} \,P(\mathbf{r}|\mathbf{x})\, d^{N}\mathbf{r} \;\geq \; \frac{1}{2}
	[\grad \ln \overline{g_{y}}(\mathbf{x})]^\mathsf{T}
	\mathbf{F}_{\text{code}}(\mathbf{x})^{-1} \grad \ln \overline{g_{y}}(\mathbf{x}),
	\label{eq:inequ_K}
\eeq
with $\grad$ the vector of components $\partial / \partial x_i$. Multiplying by $P(y|\mathbf{x} )$ and summing over $y$ and $\mathbf{x}$, one gets
\beq
	\mathcal{\overline{C}} \,\geq \, \textstyle{\int}\, D_{\text{KL}}(P(Y|\mathbf{x})\|\overline{g}(Y|\mathbf{x})) \, P(\mathbf{x})\, d^K\mathbf{x}\;+\;
	\textstyle{\frac{1}{2}} \; \textstyle{\int} \operatorname{tr}
	[\widetilde{F_{\text{cat}}}(\mathbf{x})^\mathsf{T}
	\mathbf{F}_{\text{code}}(\mathbf{x})^{-1} ]\;P(\mathbf{x})\, d^K\mathbf{x},
 \label{eq:tildeineqK}
\eeq
with
\beq
[\widetilde{\mathbf{F}_{\text{cat}}}(\mathbf{x})]_{i,j} \equiv \sum_{y} P(y|\mathbf{x})\,\frac{\partial \ln \overline{g_{y}}}{ \partial x_i}\,\frac{\partial \ln \overline{g_{y}}}{ \partial x_{j}}.
\label{eq:tildeFcat}
\eeq 
The quantity $\widetilde{\mathbf{F}_{\text{cat}}}(\mathbf{x})$ is the analogous of the true $\mathbf{F}_{\text{cat}}(\mathbf{x})$ but using the estimated posterior probabilities as outputted by the network instead of the true posterior probabilities.

If the bias vanishes, $\widetilde{\mathbf{F}_{\text{cat}}}$ becomes identical to $\mathbf{F}_{\text{cat}}$ and one gets
\beq
	\mathcal{\overline{C}} \,\geq \,
	\textstyle{\frac{1}{2}} \; \textstyle{\int} \operatorname{tr}
	[\mathbf{F}_{\text{cat}}^\mathsf{T}(\mathbf{x})
	\mathbf{F}_{\text{code}}(\mathbf{x})^{-1} ]\;P(\mathbf{x}) d^K\mathbf{x}.
	\label{eq:ineqK}
\eeq
The vanishing bias limit corresponds to the asymptotic limit discussed Sec. \ref{sec:geometry}, for which these inequalities (\ref{eq:ineq}) and (\ref{eq:ineqK}) actually become equalities (Eq.~(\ref{eq:Delta_general_1d}) and (\ref{eq:Delta_general})).

Can one show that (\ref{eq:ineq}) and (\ref{eq:ineqK}) are strict inequalities for small but nonzero bias? For nonzero bias the KL divergence is strictly positive, but it is not clear how the term in $\widetilde{\mathbf{F}_{\text{cat}}}$ behaves. 

In the case of a population code with a large number $N$ of cells, $1/\mathbf{F}_{\text{code}}(\mathbf{x})$ and the bias $b$ are of order $1/N$ (see Ref.~\cite{LBG_JPN_2008} for the precise hypothesis). Similarly, for the single cell with small noise variance $\sigma^2$ (see Appendix \ref{app:singlecell}), $1/F_{\text{code}}(x)$ and the bias $b$ are of order $\sigma^2$. Then in that cases one can check that the inequalities are strict at leading order in $1/N$ or $\sigma^2$.

Finally we note that the result Eq.~(\ref{eq:tildeineqK}), with the definition (\ref{eq:tildeFcat}), gives a different interpretation of the asymptotic formulas of the mutual information. The categorical information as defined by (\ref{eq:tildeFcat}) characterizes the relationship between category membership and feature space as seen by the network, that is from the output of the network, and not as given by the ``true'' data structure. It is only in the asymptotic limit of efficient learning that the two coincide. 

\subsection{Bounds from relationships between the Jensen gap and the variance}
\label{app:jensengap}

As mentioned Sec. \ref{sec:biasvar} following Eq. (\ref{eq:cbarx2_main}), the term within $\left\{...\right\} $ in the variance part of the decomposition, Eq. (\ref{eq:cbarx3-c_main}) -- recalled in this appendix, Eq. (\ref{eq:cbarx3-c_app}) above --, is a Jensen gap associated with the function $- \ln(.)$. Here we derive bounds on this quantity, making use of known bounds on the Jensen gap for convex functions, some of them obtained quite recently, see, e.g., Refs.~\cite{Becker2012,LiaoBerg_Sharpening2019,Lee_etal_FurtherSharpening2021,MerhavNeri_JensenLike2023}. For a convex function $\phi$ and a random variable $Z$ of distribution $P(Z)$, authors have obtained bounds on the Jensen gap 
\beq
\mathcal{J}(\phi;P) \equiv \textstyle{\int} \phi(z) P(z)dz - \phi(\textstyle{\int} z P(z) dz)
\eeq
of the form
\beq
G_{min} \; Var(Z)\; \leq\; \mathcal{J}(\phi;P)\; \leq\; G_{max} \; Var(Z),
\label{eq:gapbound}
\eeq
with $Var(Z)=\textstyle{\int} (z- \overline {z})^2 \,P(z) dz$, and $G_{min}$ and $G_{max}$ are some quantities depending on the function $\phi$ and on the distribution $P(.)$ of the random variable. If $0< G_{min}$ and $G_{max}<\infty$, these bounds characterize how the Jensen gap is constrained by the variance. In the case $0< G_{min}$, within our statistical inference context the lower bound will allow us to further make use of the Cramér-Rao bound as in the previous Sec. \ref{app:bound}. In addition, it would be interesting to get tight bounds, that is $G_{min} \lessapprox G_{max}$.

Let us consider the term specific to a given category $y$ and a given $x$, in the case $K=1$. Here and in the following, for ease of reading we omit to note the dependencies in $y$, $x$, $\mathbf{r}$ except when necessary. Hence we simply denote by $g$ the random variable $g_{y}(\mathbf{r})$ with distribution induced by $P(\mathbf{r}|x)$. We will denote by a bar the average of any quantity of the considered random variable. 

\subsubsection{A useful remark}
\label{app:biasvar1}

In the variance-type part of the decomposition, Eq. (\ref{eq:cbarx3-c_app}), the Jensen gap is the term within $\left\{...\right\} $. It is positive or zero although $ \ln g_{y}(\mathbf{r})/\overline{g_{y}}(\mathbf{x}) $ has not a constant sign. Denoting 
\beq
u(\mathbf{r})=(g_{y}(\mathbf{r})-\overline{g_{y}}(\mathbf{x}) )/\overline{g_{y}}(\mathbf{x}), 
\label{eq:u}
\eeq
since $\textstyle{\int} \,u(\mathbf{r})\,P(\mathbf{r}|\mathbf{x})\; d^{N}\mathbf{r}=0$, we can add $u(\mathbf{r})$ to the integrand, and write 
\beq 
\left\{...\right\} = \textstyle{\int} \,(u(\mathbf{r})-\ln(1+u(\mathbf{r})))\,P(\mathbf{r}|\mathbf{x})\,d^{N}\mathbf{r}.
\eeq
One can check that the function 
\beq
\phi(u) \equiv u-\ln(1+u),
\label{eq:phi}
\eeq
is always positive or zero (zero at $u=0$), convex, and the Jensen gap $\mathcal{J}$ for the convex function ($- \ln(.)$) is equivalently the one for the function $\phi$:
\beq
\mathcal{J} \;=\; \ln \overline{g} - \overline {\ln g} \;=\; \overline{\phi},
\label{eq:jensengap2}
\eeq
with
\beq
\overline{\phi}=\textstyle{\int} \phi(u) P(u)du.
\label{eq:phibar}
\eeq
Here $P(u)$ is the distribution induced by the one of $g_{y}$ given $\mathbf{r}$, $u$ being defined as in (\ref{eq:u}),
\beq
u=(g-\overline{g})/\overline{g}.
\eeq

\subsubsection{Bounds for the Jensen gap}
\label{app:jensengap-bounds}
We consider the bounds derived in Ref.~\citep{LiaoBerg_Sharpening2019} applied to the ($-\ln$) function, but which we derive here in a quite straightforward way. In the integrand in (\ref{eq:phibar}), we write $\phi(u) = u^2 \, h(u)$, with 
\beq
h(u) \equiv \frac{\phi(u)}{u^2},
\label{eq:h}
\eeq 
and we bound $h$:
\begin{eqnarray}
\mathcal{J} & \geq & Var(U) \;\inf \left\{ h(u) \right\}, \\
\mathcal{J} &\leq & Var(U) \;\sup \left\{ h(u) \right\},
\label{eq:sharpJensen}
\end{eqnarray}
with
\beq
Var(U) =\overline{u^2}.
\label{eq:varu}
\eeq
The interest of working with $\phi$ instead of ($-\ln$) should be clear: (1) $\phi$ is a non negative function, and (2), since $\overline{u}=0$, $\phi(0)=0$ and $\phi'(0)=0$, the Taylor expansion of $\phi$ near the mean $\overline{u}=0$ starts at second order. In particular $h(0)=1/2$. It remains to see if the lower and upper values that $h$ can take are, respectively, strictly positive and finite. 

The minimum value for $h$ is reached at the maximum value that the random variable $U$ can take. $u$ takes values in the range $[-1, \frac{1-\overline{g}}{\overline{g}}]$. If all this range has to be considered, then $\inf h = h(u_{\max})$ with $u_{\max}= \frac{1-\overline{g}}{\overline{g}}$. We want to lower bound $h(u_{\max})$ uniformly over $y$ and $x$. If $\overline{g}$ is close to $1$, then $h(u_{\max})$ is close to $1$. If $\overline{g}$ is close to $0$ ($x$ values for which the considered category is very unlikely), $h(u_{\max})$ is of order $\overline{g}$, hence small. For the upper bound, the maximum of $h$ is reached at $u=u_{\min}=-1$, but $h(-1)=+\infty$. At this point it is not clear how to get general bounds avoiding $0$ as lower bound and $+\infty$ as upper bound. We thus now consider more restrictive hypotheses.

\subsubsection{Case of residual ambiguity}
Let us assume that there is always some minimum ambiguity, that is, there is some small $\epsilon >0$, such that for any category $y$ and any $x$,
\beq
 \overline{g_{y}}(x) \geq \epsilon.
\eeq
(reintroducing the index $y$ to specify that we consider one particular category). Then $\inf h(u_y) \geq h(\frac{1-\epsilon}{\epsilon})$, and one gets
\beq
\mathcal{J}(g_{y};x) \;\geq\; \epsilon \; Var(U_{y})(x).
\label{eq:bound1}
\eeq
To get a finite upper bound, we need a stronger hypothesis. If we assume $g\geq \epsilon$ for any $\mathbf{r}$, then $\max h$ is of order $-\ln \epsilon$. One has thus the loose bounds, 
\beq
\epsilon \; Var(U_{y})(x) \; \leq \; \mathcal{J}(g_{y};x) \; \leq \; (-\ln \epsilon)\;Var(U_{y})(x).
\label{eq:nound1bis}
\eeq

We note that, if $\overline{g_{y}}$ is close to zero, we expect $g_{y}$ to be close to zero as well, this for almost every $\mathbf{r}$, so that actually the typical $u_y$ value should be close to zero, and then $h(u_y)$ close to $1/2$. The analysis (likely the bound) should be reconsidered to take into account that, integrating over $\mathbf{r}$ and $x$, under reasonable smoothness hypothesis rare events (such as $g_{y}(\mathbf{r})=1$ when $\overline{g_{y}} \sim 0$) should not matter.

\subsubsection{Case of small fluctuations}
In the spirit of the previous Sec. \ref{app:biasvarefficient}, we consider the hypothesis of small fluctuations around the mean. More precisely, we assume here that for any category $y$, for (almost all) $\mathbf{r}$ and $x$,
\beq
 |u_y| \leq \epsilon.
\eeq
Then we have $h(u_y)\geq h(\epsilon)\geq \textstyle{\frac{1}{2}}-\frac{\epsilon}{3}$, leading to
\beq
\mathcal{J}(g_{y};x) \;\geq\; \textstyle{\frac{1}{2}} (1-\frac{2\epsilon}{3}) \; Var(U_{y})(x).
\label{eq:bound3}
\eeq
This bound gives the $1/2$ factor as $\epsilon$ goes to zero, in agreement with Sec. \ref{app:biasvarefficient}.\\
For the upper bound, we have $h(u_y)\leq h(-\epsilon)=\frac{-\epsilon-\ln(1-\epsilon)}{\epsilon^2}$, leading to
\beq
\mathcal{J}(g_{y};x) \;\leq\; \left(\textstyle{\frac{1}{2}} + \frac{\epsilon}{3}+\frac{\epsilon^2}{4}+...\right) \; Var(U_{y})(x)
\label{eq:bound4}
\eeq
which, with the lower bound, proves that, for any $y$, $\mathcal{J}(g_{y};x) \rightarrow \textstyle{\frac{1}{2}}\, Var(U_{y})(x)$ when $\epsilon \rightarrow 0$.\\

Obtaining a more general tight bound would be of interest in statistical inference, since it would give a bound for the Bayes cost of the same nature as the Cramér-Rao bound for the quadratic cost.

\section{Asymptotic expression of the mutual information}
\label{app:eqDelta_gen}
\setcounter{equation}{0}
In this Appendix we assume the neural Fisher information matrix to be well defined (finite everywhere except possibly on a set of locations $\mathbf{x}$ of zero measure), and invertible.
 
\subsection{Derivation of the formula}
\label{app:eqDelta_gen_mainsteps}
\subsubsection{Main steps}
We give here the main steps leading to Eq. (\ref{eq:Delta_general}), Sec. \ref{sec:geometry}, that is, 
\beq
I[Y,\mathbf{R}] \;=\; I[Y,\mathbf{X}] - \textstyle{\frac{1}{2}} \textstyle{\int} \operatorname{tr} \left( \mathbf{F}_{\text{cat}}^\mathsf{T} (\mathbf{x})\,\mathbf{F}_{\text{code}}^{\,-1}(\mathbf{x}) \right)\; P(\mathbf{x}) \,d^K\mathbf{x}.
\label{eq:Delta_general_app}
\eeq
When $N$ goes to $\infty$, we expect the mutual information $I[Y, \mathbf{R}]$ to converge toward $I[Y, \mathbf{X}]$, and we are interested in the first non trivial correction to this asymptotic
limit. The main hypotheses are that, for $N$ large, the probability of $\mathbf{x}$ given $\mathbf{r}$ is sharply picked at its most probable value, the neural Fisher information matrix is invertible, and scales with the size of the neural layer. This last hypothesis corresponds to typical cases of neural noise correlations, but excludes particular types of noise correlations, see, e.g., Refs.~\cite{yoon1998correlations,abbott1999correlatedvariability,franke2016structures}. 

We thus compute for large $N$ the difference 
\beq
\Delta \equiv I[Y, \mathbf{R}] \;-\; I[Y, \mathbf{X}] \; \leq 0.
\label{eq:delta_def}
\eeq
First, as we show below, subSec. \ref{app:eqdeltaphi}, one can write
\beq
\Delta = \textstyle{\iint} P(\mathbf{r}|\mathbf{x}) \, \phi(\mathbf{x}) \; d^N\mathbf{r} \, d^K\mathbf{x},
\label{eq:delta_phi}
\eeq
where
\beq
\phi(\mathbf{x}) \equiv \sum_{y=1}^M P(\mathbf{x}) P(y|\mathbf{x}) \ln \frac{P(y|\mathbf{r})}{P(y|\mathbf{x})}.
\eeq
Then the computation is identical to the one in Ref.~\citep{LBG_JPN_2008}. We do not reproduce here this computation, but mention the main steps. The first step consists in integrating over $\mathbf{x}$. Taking the large $N$ limit, we show that the leading order is zero. We then seek for the first correction of order $1/N$, using Laplace, or steepest descent, method. The last step eventually consists in integrating over $\mathbf{r}$. 

\subsubsection{Derivation of Equation (\ref{eq:delta_phi})}
\label{app:eqdeltaphi}
The difference $\Delta$, defined above, Eq. (\ref{eq:delta_def}), can be written as
\beq
\Delta = - H[Y |\mathbf{R}] \,+\,H[Y | X],
\eeq
that is
\beq
\Delta \,=\,\textstyle{\int} \left(\textstyle{\sum}_{y} P(y|\mathbf{r}) \ln P(y|\mathbf{r})\right) P(\mathbf{r})d^N\mathbf{r}\,-\,\textstyle{\int} \left(\textstyle{\sum}_{y} P(y|\mathbf{x}) \ln P(y|\mathbf{x})\right) P(\mathbf{x})\, d^K\mathbf{x}.
\eeq
In the second term we can introduce $\textstyle{\int} P(\mathbf{r}|\mathbf{x})d^N\mathbf{r}$ (which is identically equal to $1$). For the first term, since we have the Markov chain (\ref{eq:markov-chain-small}), that is $y \rightarrow \mathbf{s} \rightarrow \mathbf{x} \rightarrow \mathbf{r}$, we can write $P(\mathbf{r}|y)= \textstyle{\int} P(\mathbf{r}|\mathbf{x}) \, P(\mathbf{x}|y )\,d^K\mathbf{x}$. Note that, $\mathbf{x}$ being a deterministic function of $\mathbf{s}$, $P(\mathbf{x})$ and $P(\mathbf{r}|\mathbf{x})$ are the distribution induced on $\mathbf{x}$ by the one of $\mathbf{s}$.
Hence,
\beqa
 P(\mathbf{r}) P(y|\mathbf{r}) &= & P(\mathbf{r}|y)\, P_{y} \nonumber\\
& =& \textstyle{\int} P(\mathbf{r}|\mathbf{x}) \, P(\mathbf{x}|y )\, P_{y}\,d^K\mathbf{x} \nonumber\\
& =& \textstyle{\int} P(\mathbf{r}|\mathbf{x}) \,P(y|\mathbf{x})\, P(\mathbf{x})\,d^K\mathbf{x}.
\eeqa
Gathering all the terms we get Eq.(\ref{eq:delta_phi}).

\subsection{Invariance by change of representation}
\label{app:invariance}
The mutual information $I[Y,\mathbf{X}]$ is invariant under any reversible transformation on $\mathbf{x}$. Thus, the asymptotic expression (\ref{eq:Delta_general}) should also be invariant under such a transformation. Let us check that this is the case. To see this, first consider the expression of the Fisher information matrix in terms of first order partial derivatives:
\beq
\big[\mathbf{F}_{\text{code}}(\mathbf{x})\big]_{ij} \; = \int \, \frac{\partial \ln P(\mathbf{r}|\mathbf{x}) }{\partial x_i}\; \frac{\partial \ln P(\mathbf{r}|\mathbf{x}) }{\partial x_j} \; P(\mathbf{r}|\mathbf{x}) \,d^N\mathbf{r}.
\label{eq:fisher_code_matrix_bis}
\eeq
Consider an arbitrary reversible function $Z$ from $\mathbb{R}^K$ to $\mathbb{R}^K$, and the change of variable $\mathbf{x} \rightarrow \mathbf{z}=Z(\mathbf{x})$ in (\ref{eq:Delta_general}). The probability density function for $\mathbf{x}$ induces a density $P_{_\mathbf{Z}}$ for $\mathbf{z}$. Obviously $I[Y,\mathbf{X}]=I[Y,\mathbf{Z}]$. Denoting by $\mathbf{J}$ the Jacobian matrix of this transformation,
\beq
\big[\mathbf{J}\big]_{ij}(\mathbf{x}) \; = \frac{\partial Z_j(\mathbf{x})}{\partial x_i},
\eeq
from (\ref{eq:fisher_code_matrix_bis}) we have
\beq
\mathbf{F}_{\text{code}}(\mathbf{x}) \; = \mathbf{J}^\mathsf{T} \, \mathbf{F}_{\text{code}}(\mathbf{z})\,\mathbf{J},
\eeq
and thus
\beq
\mathbf{F}_{\text{code}}^{\,-1}(\mathbf{x}) \; = \mathbf{J}^{-1} \, \mathbf{F}_{\text{code}}^{\,-1}(\mathbf{z})\, \mathbf{J}^{-1\;\mathsf{T}}.
\eeq
Similarly $ \mathbf{F}_{\text{cat}}(\mathbf{x}) = \mathbf{J}^\mathsf{T} \,\mathbf{F}_{\text{cat}}(\mathbf{z})\, \mathbf{J} $, hence
\beqa
\operatorname{tr}(\mathbf{F}_{\text{cat}}^\mathsf{T} (\mathbf{x})\mathbf{F}_{\text{code}}^{\,-1}(\mathbf{x})) &= & \operatorname{tr}(\mathbf{J}^\mathsf{T} \;\mathbf{F}_{\text{cat}}^\mathsf{T} (\mathbf{z})\mathbf{J}\, \mathbf{J}^{-1} \mathbf{F}_{\text{code}}^{\,-1}(\mathbf{z})\,\mathbf{J}^{-1\;\mathsf{T}}) \nonumber \\
&=& \operatorname{tr}(\mathbf{F}_{\text{cat}}^\mathsf{T} (\mathbf{z})\mathbf{F}_{\text{code}}^{\,-1}(\mathbf{z})).
\eeqa
As a result, (\ref{eq:Delta_general}) can also be written
\beq
I[Y,\mathbf{R}] = I[Y,\mathbf{Z}] - \frac{1}{2} \int\, \operatorname{tr}\left(\mathbf{F}_{\text{cat}}^\mathsf{T} (\mathbf{z}) \mathbf{F}_{\text{code}}^{\,-1}(\mathbf{z})\right)\; P_{_\mathbf{Z}}(\mathbf{z}) \,d^K\mathbf{z},
\label{eq:Delta_general_y}
\eeq
which is the same expression in terms of $\mathbf{z}$ instead of $\mathbf{x}$.

\section{Single coding cell in the low noise limit with non Gaussian distribution}
\label{app:singlecell}
\setcounter{equation}{0}
The main analysis of the mutual information, Sec. \ref{sec:geometry}, is based on a large size limit, corresponding to a large signal-to-noise limit in which one has both the noise strength going to zero (large number of cells), and the noise distribution becoming Gaussian. Considering the case of the coding of a 1D stimulus (which would correspond here to the neural coding of $x$ or $s$ instead of the category), \citet{WeiStocker2016} have shown that, with additive noise of arbitrary shape, additional terms appear as compared with the Gaussian case. To see the role of the shape of the noise distribution in the present context, we discuss here the case of a single coding cell with multiplicative noise of small amplitude. 

\subsection{Single cell model}
We assume given a discrete set of categories, $y=1,...,M$ with probabilities of occurrence $P_{y} \geq 0$, so that $\textstyle{\sum}_{y} P_{y} = 1$. Each category is characterized by a density distribution $P(\mathbf{s}|y)$ over the input (sensory) space. A sensory input $\mathbf{s} \in \mathbb{R}^N$ elicits a response $r\in \mathbb{R}$ defined as a noisy function of a scalar feature $x=X(\mathbf{s})$. Given a category $y$, the neural activity distribution is thus given by
\beq 
P(r| y) = \textstyle{\int} \, P(r|x) \, P(x|y) \, dx,
\label{eq:prmu}
\eeq
with 
\beq
P(x|y)=\textstyle{\int} \, \delta(x-X(\mathbf{s}))\, P(\mathbf{s}|y)\,d^{N_s}\mathbf{s}.
\eeq

The activity $r$ might be continuous or discrete. We consider two particular cases:\\
(i) a Poisson neuron: $r$ is the number of spikes that the cell generates during a certain time interval $t$, with a Poisson statistics with mean rate $f(x)$:
\beq
P(r|x) = \frac{\big( t f(x)\big)^{r}}{r!} \exp\big(- t f(x) \big).
\label{eq:prx_poisson}
\eeq
(ii) a continuous case,
\beq
r = f(x) + \sigma \sqrt{g(x)} \; z,
\label{eq:r_cont}
\eeq
where $f$ is a smooth invertible transfer function (e.g., $f$ increases smoothly from $0$ to $1$ as $x$ goes from $-\infty$ to $+\infty$), $g(x) \geq 0$, and $z$ a random noise of pdf $Q(z)$ having zero mean and unit variance (with $Q$ sufficiently regular and decreasing smoothly toward zero at $\pm \infty$). $\sigma$ gives the noise scale. We thus can write
\beq
P(r|x) = \frac{1}{\sigma \sqrt{g(x)} } \, Q\left(\frac{r-f(x)}{\sigma \sqrt{g(x)} }\right).
\label{eq:prx_Q}
\eeq
A particular case is the one of Gaussian noise:
\beq
P(r|x) = \frac{1}{\sqrt{2\pi \sigma^2 g(x)}} \exp\left(-\frac{(r-f(x))^2}{2 \sigma^2 g(x)} \right).
\label{eq:prx_Gauss}
\eeq
For large times, the Poisson neuron gives such a Gaussian statistics for $r/t$ with $g \equiv f$ and $\sigma=1/t$.

\subsection{The neural Fisher information for the single cell model} 
\label{app:singleFisher}
The Fisher information $F_{\text{code}}(x)$ associated with the above model (\ref{eq:r_cont}) is
\beqa
F_{\text{code}}(x) &=& - \textstyle{\int} \partial^2_{x^2} [\ln P(r|x)] \; P(r|x)\,dr\nonumber \\
&=& - \int \partial^2_{x^2} \left[\ln Q\left(Z(r,x)\right)
- \frac{1}{2} \ln g(x) \right]\; 
Q\left(Z(r,x)\right)
\,\frac{dr}{\sigma \sqrt{g(x)}}.
\eeqa
with
\beq
Z(r,x)\equiv \frac{r-f(x)}{\sigma \sqrt{g(x)}}.
\eeq
Now
\beq
\frac{\partial}{\partial x} \ln Q\left(Z(r,x)\right) =\frac{\partial}{\partial x} \left[Z(r,x)\right]\; \frac{d}{d z} \ln Q(z)\Big|_{z=Z(x)}
\eeq
with
\beq
\frac{\partial}{\partial x} \left[Z(r,x)\right] = - \left( \frac{f'(x)}{\sigma \sqrt{g(x)}} + Z(r,x) \,\frac{g'(x)}{2g(x)} \right).
\eeq
Then
\beqa
\frac{\partial^2}{\partial x^2} \ln Q\left(Z(r,x)\right) &=& 
- \left(\frac{\partial}{\partial x} \,\left( \frac{f'(x)}{\sigma \sqrt{g(x)}} + Z(r,x) \,\frac{g'(x)}{2g(x)} \right)\right)\,\, \frac{d}{d z} \ln Q(z)\Big|_{z=Z(r,x)} \nonumber \\
&+& \left[ \frac{f'(x)}{\sigma \sqrt{g(x)}} + Z(r,x) \,\frac{g'(x)}{2g(x)} \right]^2\;\frac{d^2}{d z^2} \ln Q(z)\Big|_{z=Z(r,x)}.
\eeqa 
Then we can compute $F_{\text{code}}(x)$ making the change of variable $r\rightarrow z=\frac{r-f(x)}{\sigma \sqrt{g(x)}}$, and making use of $\textstyle{\int} Q(z)\frac{d}{d z} \ln Q(z) \; dz=0$, $\textstyle{\int} z\,Q(z) \frac{d}{d z}\ln Q(z) \; dz=-1$, 
\beqa
F_{\text{code}}(x) &=& \frac{f'^2(x)}{\sigma^2 g(x)} F_Q \nonumber \\
&-&\frac{f'(x)}{\sigma \sqrt{g(x)} }\frac{g'(x)}{g(x)}\,\int Q(z)\,z\, \frac{d^2}{dz^2} \ln Q(z)\;dz \nonumber \\
&+& \left(\frac{g'(x)}{2g(x)}\right)^2\; (1- \int Q(z)\,z^2\, \frac{d^2}{d z^2} \ln Q(z)\;dz).
\eeqa
where
\beq
F_Q\equiv - \int Q(z) \frac{d^2}{dz^2} \ln Q(z) \;dz.
\label{eq:single_FQ}
\eeq
If the noise distribution is symmetric, $Q(-z)=Q(z)$, then the second term (of order $1/\sigma$) is zero. In the limit of small noise, the Fisher information is given by the first term, of order $1/\sigma^2$. $F_Q$ is the Fisher information for the model ``$\text{output}=x+z$'' (Fisher information which is in fact independent of $x$). Notice that $F_Q$ is independent of the noise strength $\sigma$. The quantity $F_Q/\sigma^2 $ is the same as the quantity noted $J[\delta]$ in Ref.~\cite{WeiStocker2016}. For the Gaussian case, $F_Q=1$. Since the Gaussian distribution minimizes the Fisher information (see, e.g., Refs.~\cite{Stoica_Babu_GaussianData_2011,Park_etal_Gaussian2013}), for an arbitrary distribution $Q$ of unit variance one has
\beq
F_Q \;\geq\; 1.
\label{eq:FQ>1}
\eeq
As a side remark, we note that from Stam inequality \cite{Stam1959} (which is central in the derivation of the results in Ref.~\cite{WeiStocker2016} mentioned above), one can also deduce that $F_Q$ is always greater than or equal to $1$. In our notation, this inequality can be written as
\beq
\frac{1}{2} \ln F_Q \;\geq\; \frac{1}{2} \ln 2\pi e \;-\;H_Q,
\label{eq:stam}
\eeq
where $H_Q$ is the entropy of the noise distribution,
\beq
H_Q = - \textstyle{\int} \, \ln Q(z) \;Q(z)\,dz.
\label{eq:single_HQ}
\eeq
The right-hand side in (\ref{eq:stam}) is the difference between the entropy of the Gaussian of unit variance and that of the $Q$ distribution, also with unit variance. Since the Gaussian distribution maximizes the entropy among the distributions of identical variance, this difference is positive or zero. Hence $\ln F_Q \;\geq\;0$, which implies $F_Q \;\geq\; 1$.

In the case of small noise limit, one can write
\beq 
F_{\text{code}}(x) = F_Q \,F_{\text{code}}^G(x),
\label{eq:single_FGQ}
\eeq
where
\beq 
F_{\text{code}}^G(x) \equiv \frac{f'^2(x)}{\sigma^2 g(x)}
\label{eq:single_FG}
\eeq
is the neural Fisher information in the Gaussian case.

\subsection{Mutual Information for the single coding cell: Asymptotic expression}
\subsubsection{Main result}
In the limit of small noise, for the mutual information we obtain
\beq
I[Y,R] = I[Y,X] - \frac{1}{2} \int \frac{F_{\text{cat}}(x)}{F_{\text{code}}^G(x)}\,P(x)\,dx,
\label{eq:single_result}
\eeq
with $F_{\text{code}}^G(x)$ given by (\ref{eq:single_FG}). We detail the proof in the next section, \ref{app:singlecell-proof}.

In the case of a large number of coding cells discussed in the main text, the asymptotic noise distribution is Gaussian. Here, for a small noise but a non Gaussian distribution $Q$, we see that the factor $F_Q$, given by Eq. (\ref{eq:single_FQ}), appears in the neural Fisher information (as shown above), but not in the mutual information, as if the noise had a Gaussian distribution. Since $F_{\text{code}}(x)=F_Q\,F_{\text{code}}^G(x)$, we can write (\ref{eq:single_result}) as
\beq
I[Y,R] = I[Y,X] - \frac{F_Q}{2} \int \frac{F_{\text{cat}}(x)}{F_{\text{code}}(x)}\,P(x)\,dx.
\label{eq:single_result_bis}
\eeq
Since $F_Q \geq 1$, we can write that, in the limit of vanishing noise, whatever the noise distribution, 
\beq
I[Y,R] \;\leq\; I[Y,X] \,-\, \frac{1}{2} \int \frac{F_{\text{cat}}(x)}{F_{\text{code}}(x)}\,P(x)\,dx.
\label{eq:single_inequ}
\eeq
with strict inequality if the noise distribution is not Gaussian, and equality for the Gaussian case. 
Note that, however, the presence or absence of the factor $F_Q$ in the asymptotic formula is of little importance for the optimization problem, both quantitatively and qualitatively. It would be interesting to get the next order term in the small noise expansion (or a more general result), to see if/when the asymptotic formula (\ref{eq:single_result}) is approached from above or from below.

In this high-efficiency regime, the bias-variance decomposition introduced Sec. \ref{sec:biasvar} allows to get a lower bound for the mean cost, 
\beq
 \overline{\mathcal{C}} \;\geq \; \frac{1}{2} \int \; \frac{F_{\text{cat}}(x)}{F_{\text{code}}(x)}\,P(x)\,dx,
\label{eq:single_inequ_biasvar}
\eeq
in agreement with the above inequality (\ref{eq:single_inequ}). 

Qualitatively, one can conclude here that maximizing the mutual information implies finding the transformation $\mathbf{s}\rightarrow x=X(\mathbf{s})$ which maximizes the mutual information $I[Y,X]$ and to fix the geometry in the $x$ space so that $F_{\text{code}}^G$ follows $F_{\text{cat}}$.

\subsubsection{Proof of the asymptotic formula}
\label{app:singlecell-proof}

Here we derive the expression (\ref{eq:single_result}) of the mutual information for the single cell model. We recall the hypotheses. All functions and pdf are as regular as needed. The transfer function $f$ is strictly monotonic ($f'(x)>0$), hence invertible. The pdf $Q$ has zero mean and unit variance, and is monotonically decreasing to zero as its argument goes to $\pm \infty$. For simplicity we further assume $Q(z)=Q(-z)$.

\noindent {\em Leading term.}\\
By definition the mutual information between the category membership and the neural activity is given by
\beq I(Y,R)= \sum_{y} P_y \int \, P(r|y) \ln \frac{P(r|y)}{P(r)}\,dr. \eeq 
In the limit of vanishing noise, $r=f(x)$, with $f$ invertible. The mutual information being invariant under any invertible change of variable,
\beq \lim_{\sigma\rightarrow 0} I(Y,R)=I(Y,X).
\eeq

\noindent {\em First non trivial order in the noise amplitude.}\\
We now consider the first non trivial order in $\sigma$ in the expansion of the mutual information. Thus we consider the expansion at small $\sigma$ of the difference
\beq 
\Delta = I(Y,R)-I(Y,X) \leq 0. 
\eeq
We can write $\Delta$ as:
\beq \Delta = \sum_{y} P_y \int \int \, P(r|x) P(x|y) \left(\ln \frac{P(r|y)}{P(x|y)}- \ln \frac{P(r)}{P(x)} \right)\,dr\,dx.
\label{eq:Delta}
\eeq 
From this expression and the structure of the model, one can anticipate which terms in the small noise expansion may contribute to the final result. In the following, we will denote by $\{\dots\}$ any term that we will not have to compute explicitly, as shown later.

Given the model (\ref{eq:prx_Q}), the difference $\Delta$ can be written as
\beq
\Delta= \sum_{y} P_y \int \int \, \frac{1}{\sigma \sqrt{g(x)}} Q\left(\frac{r-f(x)}{\sigma\sqrt{g(x)}}\right) P(x|y) 
\left(\ln \frac{P(r|y)}{P(x|y)}- \ln \frac{P(r)}{P(x)} \right)\,dr\,dx. \eeq 
Making the change of variable $r \rightarrow z=(r-f(x))/\sigma \sqrt{g(x)}$, 
\beq 
\Delta= 
\sum_{y} P_y \textstyle{\int} \textstyle{\int} \, Q\left(z \right) P(x|y)
\left( \ln A(x,z|y) - \ln B(x,z) \right)
\, dz \,dx, 
\label{eq:log-log}
\eeq 
with
\beq 
A(x,z|y) =\int Q\left(\frac{f(x)-f(x')}{\sigma \sqrt{g(x')}} \;+\; \sqrt{\frac{g(x)}{g(x')}} z \right)\,
\frac{P(x'|y)}{P(x|y)} \,\frac{dx'}{\sigma \sqrt{g(x')}},
\eeq
and similarly for $B(x,z)$, with $P(x)$ instead of $P(x|y)$:
\beq 
B(x,z) = \int Q\left(\frac{f(x)-f(x')}{\sigma \sqrt{g(x')}} \;+\; \sqrt{\frac{g(x)}{g(x')}} z \right)\,
\frac{P(x')}{P(x)}\, \frac{dx'}{\sigma \sqrt{g(x')}}.
\eeq
For $\sigma$ small, the integration over $x'$ is dominated by the vicinity of $x'=x$. 
From the Cramér-Rao inequality, one would expect the relevant domain of $(x'-x)^2$ to scale with the (inverse of the) neural Fisher information, $F_{\text{code}}(x)$, but actually it is only the Gaussian part, $F_{\text{code}}^G(x)$, which appears. We have $\frac{f(x)-f(x')}{\sigma \sqrt{g(x')}} =- (x'-x)\frac{f'(x)}{\sigma \sqrt{g(x)}}+...$, and we recognize that $\sqrt{F_{\text{code}}^G(x)}=\frac{f'(x)}{\sigma \sqrt{g(x)}}$. 
In both $A$ and $B$, we thus make the change of variable $x'\rightarrow u$ with
\beq
x'=x-\sigma \frac{\sqrt{g(x)}}{f'(x)}\,u = x - \frac{1}{\sqrt{F_{\text{code}}^G(x)}}\,u.
\eeq
We expand to second order in $\sigma$:
\beq
\frac{P(x'|y)}{P(x|y)} = 1- \frac{u}{\sqrt{F_{\text{code}}^G(x)}}
\,\frac{P'(x|y)}{P(x|y)} + 
\frac{u^2}{2\,F_{\text{code}}^G(x)} 
\,\frac{P''(x|y)}{P(x|y)},
\eeq
and similarly for $P(x)$, and the argument of the pdf $Q$ is 
\beq
\frac{f(x)-f(x')}{\sigma \sqrt{g(x')}} + z \sqrt{\frac{g(x)}{g(x')}}= u+z +\sigma \{\dots\} +\sigma^2 \{\dots\}.
\eeq
We expand $A$ and $B$ to second order in $\sigma$, and we perform the integration over $u$, giving terms that may depend on the variable $z$. The integration over the variable $u$ is easily performed. In particular, for the terms which may contribute, one make use of $\int Q(u+z)\, du=1$, $\int Q(u+z)\, u\, du=-z$, and $\int Q(u+z)\, u^2\, du=1+z^2$, leading to
\beqa
A(x,z|y) &=& 1 + \frac{z}{\sqrt{F_{\text{code}}^G(x)}}
\,\frac{P'(x|y)}{P(x|y)} + \sigma \{\dots\} \nonumber \\
&+& \frac{1+z^2}{F_{\text{code}}^G(x)}
\,\frac{P''(x|y)}{P(x|y)}\, 
 + \frac{\sigma \{\dots\}}{\sqrt{F_{\text{code}}^G(x)}}
\,\frac{P'(x|y)}{P(x|y)} + \sigma^2 \{\dots\}.
\eeqa
In the above expression, the terms $\{\dots\}$ do not depend on $y$. Note that, from the expansion of $Q(...)$, terms in $Q'(u+z)$ contribute to the terms of order at least $\sigma$ which do not depend on $y$, and to the one of order $\sigma^2$ proportional to $P'(x|y)/P(x|y)$, second line of the above equation -- and as we will see this term in $P'(x|y)/P(x|y)$ eventually do not contribute. The terms in $Q''(u+z)$ contribute to the very last term of order $\sigma^2$ in this equation. For the quantity $B(x,z)$, again we have exactly the same terms replacing $P(x|y)$ by $P(x)$. 

We can now expand the logarithms in (\ref{eq:log-log}) to second order in $\sigma$ -- making use of $\ln(1+ \{\dots\})= \{\dots\}-\frac{1}{2} (\{\dots\})^2$ + higher-order terms. The terms which, in the part coming from $A(x,z|y)$, do not depend on $y$, cancel with the corresponding terms in the part coming from $B(x,z)$. Hence in the difference $\Delta$, the only remaining terms are
\beqa
& \; & (\sigma \{\dots\} + \sigma^2 \{\dots\}) \left( \frac{P'(x|y)}{P(x|y)} - \frac{P'(x)}{P(x)} \right) 
\label{eq:p'}\\
& + & \sigma^2 \{\dots\} \left( \frac{P''(x|y)}{P(x|y)} - \frac{P''(x)}{P(x)} \right) \label{eq:p''}\\
& - & \frac{z^2}{2} \frac{1}{F_{\text{code}}^G(x)} \left( \frac{P'^2(x|y)}{P^2(x|y)} - \frac{P'^2(x)}{P^2(x)} \right),
\label{eq:p'2}
\eeqa
in which the last one, Eq. (\ref{eq:p'2}), comes from the square of the first order of the argument of the logarithms.\\
All these terms have to be integrated over the variable $z$ with pdf $Q$, multiplied by $P_y P(x|y)$, summed over $y$ and integrated over the variable $x$. Let us first show that the terms (\ref{eq:p'}) and (\ref{eq:p''}) give zero. Multiplying by $P(x|y)$, one gets $\sum_y P_y P'(x|y)-\sum_y P_y P(x|y)\frac{P'(x)}{P(x)}=P'(x)-P'(x)=0$. Similarly, $\sum_y P_y P''(x|y)-\sum_y P_y P(x|y)\frac{P''(x)}{P(x)}=P''(x)-P''(x)=0$.\\ Now consider the last term, (\ref{eq:p'2}). The integration over the variable $z$ gives $\int Q(z) z^2 dz=1$. We have then
\beq
\Delta = - \frac{1}{2} \int \frac{1}{F_{\text{code}}^G(x)} \,\sum_y P_y P(x|y) \left( \frac{P'^2(x|y)}{P^2(x|y)} - \frac{P'^2(x)}{P^2(x)} \right)\;dx.
\eeq
From Bayes, $P_y P(x|y)=P(y|x) P(x)$, and $ \frac{P'^2(x|y)}{P^2(x|y)} - \frac{P'^2(x)}{P^2(x)}= \left(\frac{d \ln P(y|x)}{dx}\right)^2$, so that the correction to the leading term is
\beq
\Delta = - \frac{1}{2} \int \, \frac{F_{\text{cat}}(x)}{F_{\text{code}}^G(x)}\;P(x) \,dx,
\eeq
which is the announced result.

\section{Neural Fisher information: multidimensional case with non-Gaussian additive noise}
\label{app:Fisher_KN}
\setcounter{equation}{0}
To supplement the discussion on the role of the noise, end of Sec. \ref{sec:geometry} and Appendix \ref{app:singlecell}, we derive here the expression of the Fisher information matrix $\mathbf{F}_{\text{code}}(\mathbf{x})$, for $\mathbf{x} \in \mathbb{R}^K$ and $\mathbf{r} \in \mathbb{R}^N$, in the case of an arbitrary noise distribution (but see also Ref.~\cite{LBG_These} for the Gaussian case). For simplicity we only consider additive noise. More precisely, we consider the model
\beq
 \mathbf{s}\rightarrow \mathbf{x}=\mathbf{X}(\mathbf{s}) \in \mathbb{R}^K \rightarrow \mathbf{r} \in \mathbb{R}^N, \mathbf{r}=\{r_i=f_i(\mathbf{x})+\sigma \,z_i\}_{i=1}^N,
\label{eq:modelnN}
\eeq
where the $f_i$ are arbitrary transfer functions (or ``tuning curves'') -- assumed smooth and differentiable, but not necessarily invertible --, and with the noise $\mathbf{z}=\{z_i\}_{i=1}^N$ of arbitrary distribution $Q$ with zero mean, $\textstyle{\int} \mathbf{z}\, Q(\mathbf{z})\,d^N\mathbf{z} =0$, and covariance matrix $\mathbf{C}$,
\beq
\left[\mathbf{C}\right]_{i,i'}= \textstyle{\int} \,z_i z_{i'}\,Q(\mathbf{z})\,d^N\mathbf{z}.
\eeq
We assume $\left[\mathbf{C}\right]_{i,i}=1$ so that $\sigma$ is the noise strength (not necessarily small) common to all coding cells.

The $K\times K$ Fisher information matrix components are then
\beqa
\left[\mathbf{F}_{\text{code}}(\mathbf{x})\right]_{j,j'} &=& - \int P(\mathbf{r}|\mathbf{x})\,\frac{\partial}{\partial x_j}\frac{\partial}{\partial x_{j'}}\,\ln P(\mathbf{r}|\mathbf{x})\; d^N\mathbf{r},
\eeqa 
with
\beq
P(\mathbf{r}|\mathbf{x}) = \frac{1}{\sigma^N}\, Q\left(\Bigl \{\frac{r_i-f_i(\mathbf{x})}{\sigma}\Bigl \}_{i=1}^N\right).
\eeq
With $z_i=(r-f_i(\mathbf{x}))/\sigma$, 
\beq
\frac{\partial}{\partial x_j} = -\sum_i \frac{1}{\sigma }\frac{\partial f_i(\mathbf{x})}{\partial x_j}\, \frac{\partial}{\partial z_i}.
\eeq
Then
\beq
\left[\mathbf{F}_{\text{code}}(\mathbf{x})\right]_{j,j'} = - \frac{1}{\sigma^2 }\int Q(\mathbf{z})
\sum_{i, i'} \frac{\partial f_i}{\partial x_j}\, \frac{\partial}{\partial z_i}\;\frac{\partial f_{i'}}{\partial x_{j'}}\, \frac{\partial}{\partial z_{i'}}\,\ln Q(\mathbf{z})\,d^N\mathbf{z}.
\eeq
Introducing the Fisher information matrix associated with the distribution $Q$,
\beq
\left[\mathbf{F}_{Q}\right]_{i,i'} = - \int Q(\mathbf{z})\,\frac{\partial}{\partial z_{i}} \frac{\partial}{\partial z_{i'}}\, \ln Q(\mathbf{z})\;d^N\mathbf{z},
\eeq
one finally gets
\beq
\left[\mathbf{F}_{\text{code}}(\mathbf{x})\right]_{j,j'} = \frac{1}{\sigma^2 } \sum_{i, i'} \frac{\partial f_i}{\partial x_j}(\mathbf{x}) \left[\mathbf{F}_{Q}\right]_{i,i'} \frac{\partial f_{i'}}{\partial x_{j'}}(\mathbf{x}),
\label{eq:FnN}
\eeq
or equivalently,
\beq
\mathbf{F}_{\text{code}}(\mathbf{x}) = \frac{1}{\sigma^2 } \grad f^\mathsf{T}(\mathbf{x})\, \mathbf{F}_{Q} \,\grad f(\mathbf{x}).
\label{eq:FnN2}
\eeq
Note that for the Gaussian case, 
\beq
\mathbf{F}_{Q} = \mathbf{C}^{-1}.
\eeq

From (\ref{eq:FnN2}), one sees that the neural Fisher information combines three components: the noise amplitude, $\sigma$, the shape of the noise distribution through $\mathbf{F}_{Q}$, and the local changes of metric due to the transfer functions (or tuning curves) $f_i$. 

For uncorrelated noise, that is $Q(\mathbf{z})=\Pi_{i=1}^N Q_i(z_i)$, $\left[\mathbf{F}_{Q}\right]_{i,i'} =\delta_{i,i'}\, F_{Q_i}$, and 
\beq
\left[\mathbf{F}_{\text{code}}(\mathbf{x})\right]_{j,j'} = \frac{1}{\sigma^2 } \sum_{i} F_{Q_i} \,\frac{\partial f_i}{\partial x_j}\, \frac{\partial f_{i}}{\partial x_{j'}}.
\label{eq:FnNuncor}
\eeq
If in addition all the noise distributions are identical, $Q_i=Q$, $F_{Q_i}=F_{Q}$, then
\beq
\left[\mathbf{F}_{\text{code}}(\mathbf{x})\right]_{j,j'} = \frac{F_{Q}}{\sigma^2 } \sum_{i} \,\frac{\partial f_i}{\partial x_j}\, \frac{\partial f_{i}}{\partial x_{j'}},
\label{eq:FnNuncor2}
\eeq
so that, as in the 1D case (see above, Sec. \ref{app:singlecell}), the contribution of the noise distribution reduces to a global multiplicative factor, with thus little impact on the optimization issues. Actually, (\ref{eq:FnNuncor2}) can be interpreted by saying that the Fisher information is the one of the model with re-scaled transfer functions $\tilde{f}_i(\mathbf{x})= \sqrt{F_Q}\, f_i(\mathbf{x})$, and independent normal noises. It is only in the case of correlated noise that the structure of the noise distribution plays a role in the optimization of the neural code. 

\section{Optimization of the neural Fisher Information}
\label{app:optFisher}
\setcounter{equation}{0}
We comment here on the optimization of the neural Fisher information for a given projection space $X$, hence a given categorical Fisher information. As explained in the main text, Sec. \ref{sec:geometry}, the general result is that minimization of the coding cost requires that $F_{\text{code}}$ essentially follows the categorical Fisher information $F_{\text{cat}}$. The precise result will depend on the constraints on the neural system. The constraints may be on the parameters of the neurons, as in Ref.~\cite{KB_JPN_2020}, or directly on the Fisher information considered as a function, as in Ref.~\cite{LBG_JPN_2008}, which is what we consider here.

In Sec. \ref{app:galconstraint} below we first consider the minimization under a general constraint, making explicit the solution in a simple case. Then, in Sec. \ref{app:adopting-IB}, we consider the information-theoretic constraint adopting an information bottleneck view point. This allows us to further discuss, Secs. \ref{app:adopting-IB} and \ref{app:scaling}, the links between our approach and the IB one (see Sec.~\ref{sec:IB}). Finally, in Sec. \ref{app:opt-more-details}, we provide additional details on the optimization under a general constraint.

\subsection{Minimization of the coding cost under constraints}
\label{app:galconstraint}
For simplicity we only consider here the 1D case. We want to minimize the right-hand side of equation (\ref{eq:Delta_general_1d}) over the choice of the function $F_{\text{code}}$, under a chosen constraint $\Psi{}$, an increasing function of its argument. Introducing a Lagrange multiplier $\lambda$ for the constraint, the quantity to minimize becomes
\beq
\mathcal{E} = \frac{1}{2} \int \frac{F_{\text{cat}}(x)}{F_{\text{code}}(x)}\; P(x)\,dx 
\;+\; 
\lambda \left(\textstyle{\int} \Psi(F_{\text{code}}(x))\;P(x)\,dx \;-\;c \,\right),
\label{eq:optFishPsi}
\eeq
leading to $F_{\text{code}}(x)$ solution of 
\beq
F_{\text{code}}(x)^2\, \Psi'(F_{\text{code}}(x)) = \frac{1}{2\lambda}\,F_{\text{cat}}(x).
\eeq
For instance, if $\Psi(F)=F^{\alpha}$, one gets 
\beq
F_{\text{code}}(x) = \left(\frac{F_{\text{cat}}(x)}{2 \alpha \lambda}\right)^{\frac{1}{1+\alpha}},
\eeq 
which is meaningful for $\alpha>0$.
The second derivative at this solution is
\beq
\frac{\partial^2 \mathcal{E}}{\partial F_{\text{code}}(x)^2} = P(x)\, \frac{1}{2}\,(\alpha+1) (2\alpha \lambda)^{\frac{3}{\alpha+1}}\,F_{\text{cat}}(x)^{\frac{\alpha-2}{\alpha+1}}\,,
\eeq
which is strictly positive wherever $P(x)>0$ and $F_{\text{cat}}(x)>0$. The limit $\alpha \rightarrow 0$, with $\beta\equiv 1/(2\alpha \lambda)$ fixed as $\alpha \rightarrow 0$, corresponds to the IB-type information-theoretic constraint discussed below. 

If one wants to preserve the invariance by change of coordinates satisfied by the coding cost (see Appendix \ref{app:invariance}), then one may consider a constraint on $\Psi(F_{\text{code}}(x)/G(x))$ for some well behaved strictly positive function $G$ that one may want to choose as a reference:
\beq
\mathcal{E} = \frac{1}{2} \int \frac{F_{\text{cat}}(x)}{F_{\text{code}}(x)}\; P(x)\,dx 
\;+\; 
\lambda \left[\textstyle{\int} \Psi(\frac{F_{\text{code}}(x)}{G(x)})\;P(x)\,dx \;-\;c \,\right].
\label{eq:optFishPsiG}
\eeq
A natural choice for this function is here to take $G(x)\equiv F_{\text{cat}}(x)$. In that case, in the cost, $F_{\text{code}}$ only appears in the ratio $F_{\text{code}}/F_{\text{cat}}$. As a result, the optimum is always $F_{\text{code}}(x) \propto F_{\text{cat}}(x)$ (only the proportionality constant depends on the function $\Psi$). In Sec. \ref{app:opt-more-details} below, we give more details on the optimization for an arbitrary constraint.

\subsection{Adopting the Information Bottleneck viewpoint}
\label{app:adopting-IB}
As presented Sec. \ref{sec:IB}, adopting the viewpoint of the Information Bottleneck approach~\citep{tishby99information}, we may minimize the mutual information $I[X,\mathbf{R}]$ under the constraint that the information conveyed by the neural code about the categories is large enough:
\beqa
\mathcal{E} &=& I[X,\mathbf{R}] - \beta I[Y,\mathbf{R}].
\label{eq:bottleneck_1}
\eeqa
In the same asymptotic limit as the one considered here, \citet{Brunel_Nadal_1998} have shown that $I[X,\mathbf{R}] $ behaves as $\textstyle{\frac{1}{2}} \textstyle{\int} \ln F_{\text{code}}(x)\;P(x)\,dx$ (again here for $K=1$). Combining the results from Refs.~\citep{Brunel_Nadal_1998} and \citep{LBG_JPN_2008} we can thus write 
\beqa
\mathcal{E} &=& \textstyle{\frac{1}{2}}\, \textstyle{\int} \ln F_{\text{code}}(x)\;P(x)\,dx \nonumber \\
&-& \beta \left(I[Y,X] - \frac{1}{2} \int \frac{F_{\text{cat}}(x)}{F_{\text{code}}(x)}\,P(x)\,dx\right).
\label{eq:bottleneck_2}
\eeqa
Up to the (here constant) term $I[Y,X]$, this is equivalent to the cost (\ref{eq:optFishPsi}), in the case $\Psi(.)=\ln(.)$, taking the dual approach -- that is exchanging the roles of the cost and the constraint, $\beta=1/\lambda$. The optimal function is here $F_{\text{code}}(x) \propto F_{\text{cat}}(x)$. Note that this entropic or IB case is a particular example with a constraint which preserves the invariance by change of coordinate. Indeed, under a change of coordinates, the cost term give an additional constant term, hence not affecting the optimization.

The generalization of (\ref{eq:bottleneck_2}) to the multidimensional case is
\beqa
\mathcal{E} &=& \textstyle{\frac{1}{2}} \textstyle{\int} \ln \det \mathbf{F}_{\text{code}}(\mathbf{x})\;P(\mathbf{x}) \,d^K\mathbf{x} \nonumber \\
&-& \beta \left(I[Y,X] - 
\frac{1}{2} \int \operatorname{tr} \left(\, \mathbf{F}_{\text{cat}}^\mathsf{T} (\mathbf{x})\,\mathbf{F}_{\text{code}}^{\,-1}(\mathbf{x}) \right)\, P(\mathbf{x}) \,d^K\mathbf{x} \,\right),
\label{eq:bottleneck_3}
\eeqa
for which the optimum is again $\mathbf{F}_{\text{code}}(\mathbf{x}) \propto \mathbf{F}_{\text{cat}}(\mathbf{x})$. To see this, one can expand $\mathcal{E}(\mathbf{F}_{\text{code}}+\delta \mathbf{F})$ for a small perturbation $\delta \mathbf{F}$ such that $\mathbf{F}_{\text{code}}+\delta \mathbf{F}$ remains a symmetric positive-definite matrix. Making use of $\ln \det = \operatorname{tr} \ln$, and of the cyclic property of the trace, one gets 
\beq
\delta \mathcal{E}= \frac{1}{2} \textstyle{\int} \operatorname{tr} \left[ \left( - \beta \mathbf{F}_{\text{code}}^{\,-1}(\mathbf{x})\mathbf{F}_{\text{cat}}^\mathsf{T}(\mathbf{x}) \mathbf{F}_{\text{code}}^{\,-1}(\mathbf{x}) + \mathbf{F}_{\text{code}}^{\,-1}(\mathbf{x})\right) \delta \mathbf{F}\right] P(\mathbf{x}) \,d^K\mathbf{x}.
\eeq 
This perturbation is null for any $\delta \mathbf{F}$ if
\beq
\mathbf{F}_{\text{code}}=\beta \,\mathbf{F}_{\text{cat}}. 
\eeq
This solution does corresponds to a minimum of the cost: one finds that the next order in the expansion, taken at this solution (that is considering $\mathbf{F}_{\text{code}}=\beta \,\mathbf{F}_{\text{cat}} +\delta \mathbf{F}$), is $\frac{1}{4\beta^2}\,\textstyle{\int} \operatorname{tr} [(\mathbf{F}_{\text{cat}}(\mathbf{x})^{-1} \delta \mathbf{F})^2 ]\,P(\mathbf{x})\,d^K\mathbf{x}\;>0$. 

\subsection{A note on scaling and bifurcations} 
\label{app:scaling}
An important remark about scaling is in order. In the large signal-to-noise ratio limit considered here, $F_{\text{code}}(x)$ scales as the inverse of the variance $\sigma^2$ of the noise -- $\sigma^2 \sim 1/t$ if we consider a Poisson process for the neuron activity, $t$ being the observation time. Writing $F_{\text{code}}(x)=F_{\text{code}}^0(x)\, / \,\sigma^2$, the relevant terms for the optimization in (\ref{eq:bottleneck_2}) are
\beq
\frac{1}{2} \int \ln F_{\text{code}}^0(x)\;P(x)\,dx \,+\, \beta \, \frac{\sigma^2}{2} \int \frac{F_{\text{cat}}(x)}{F_{\text{code}}^0(x)}\; P(x)\,dx.
\label{eq:bottleneck_4}
\eeq
One sees that, if $\beta$ is of order $1$, the second term is negligible compared with the first one. A consistent solution requires that $\beta$ scales as $\beta=\beta_0/\sigma^2$. In such case, the derivative with respect to $F_{\text{code}}^0(x)$ gives (whenever $P(x) \neq 0$)
\beq
F_{\text{code}}^0(x)=\beta_0 F_{\text{cat}}(x).
\eeq
One can check that this is a minimum of the cost, its second derivative being $P(x)/(2\beta_0^2 F_{\text{cat}}^2) > 0$ (wherever $P(x)$ is not null). Thus the relevant IB regime here is the one of large $\beta$.\\

If one insists on working with a finite $\beta$, that is fixing a finite value as $\sigma\rightarrow 0$, the optimization of (\ref{eq:bottleneck_4}) has no solution. This suggests that a finite $\beta$ value would correspond to a regime where the asymptotic limit is not reached (hence in this case the asymptotic expression of the mutual information can no longer be used). Taking the example of a Poisson process, the correspondence is $\beta$ small $\approx$ short time limit, $\beta$ large $\approx$ large time limit. From the results obtained on the information conveyed by a Poisson neuron about a stimulus \citep{Stein_1967,Brunel_Nadal_1997,Bethge_etal_2003}, and on the coding of categories at short times \citep{LBG_These}, we expect to observe bifurcations in the optimal solution as one increases $\beta$, in line with the known IB bifurcations \citep{tishby-zaslavsky2015}.

\subsection{Optimization for an arbitrary constraint}
\label{app:opt-more-details}
We give here more details on the optimization for a general constraint. We consider the case of a general function $\Psi$ and an arbitrary function $G$ in the cost (\ref{eq:optFishPsiG}). We assume that the function $\Psi$ is positive, piecewise twice differentiable, and with a strictly positive derivative at any point $x$ for which $P(x)>0$. We denote by $\Omega$ this part of the space, and assume $G >0$ in $\Omega$. We rewrite the cost in terms of
\beq
u(x)\equiv \frac{F_{\text{code}}(x)}{G(x)},\;\; v(x) \equiv \frac{F_{\text{cat}}(x)}{G(x)},
\eeq
\beq
\mathcal{E} = \frac{1}{2} \int \frac{v(x)}{u(x)}\; P(x)\,dx 
\;+\; \lambda \left(\textstyle{\int} \Psi(u(x))\;P(x)\,dx \;-\;c \,\right).
\label{eq:cost-u}
\eeq
The first order equation gives, for each $x$ in $\Omega$, $u(x)^2\, \Psi'(u(x)) = v(x)/(2\lambda).$ Since the locations in $\Omega$ are only coupled through the global constraint, we can parametrize the solutions by the values of $v$:
\beq
u^2\, \Psi'(u) = v/(2\lambda).
\label{eq:uopt}
\eeq
There is a unique solution $u[v] $ if the function $u \rightarrow u^2\, \Psi'(u)$ is strictly monotonic (this is the case for the example of the power-law function given above, Sec. \ref{app:galconstraint}). Otherwise, there will be ranges of $v$ values for which there are multiple solutions. The one giving the smallest contribution to the cost should be selected.

Solutions $u[v] $ of this equation contribute to the minimum of the cost if the second derivative is positive, that is if
\beq
\frac{v}{u^3}\;+\;\lambda \Psi''(u) \;> 0
\eeq
at $u=u[v]$. 

Let us consider a range of values of $v$ for which the solution $u[v] $ is continuously differentiable function of $v$. Taking the derivative of (\ref{eq:uopt}) with respect to $v$, we have 
\beq
\left( 2u\Psi'(u)+ u^2 \Psi''(u) \right)\;\frac{du}{dv} \,=\, 1/(2\lambda).
\eeq
Making use of (\ref{eq:uopt}), this gives
\beq
\left( \frac{v}{u^3}\;+\;\lambda \Psi''(u) \right)\;\frac{du}{dv} \,=\,\frac{1}{2u^2}\;> 0.
\eeq
Thus the second derivative is positive provided $u[v]$ is an increasing function of $v$. 

\begin{figure}
\centering
\textbf{a.}\hspace{-0.1cm}
\includegraphics[height=3.4cm,valign=t]{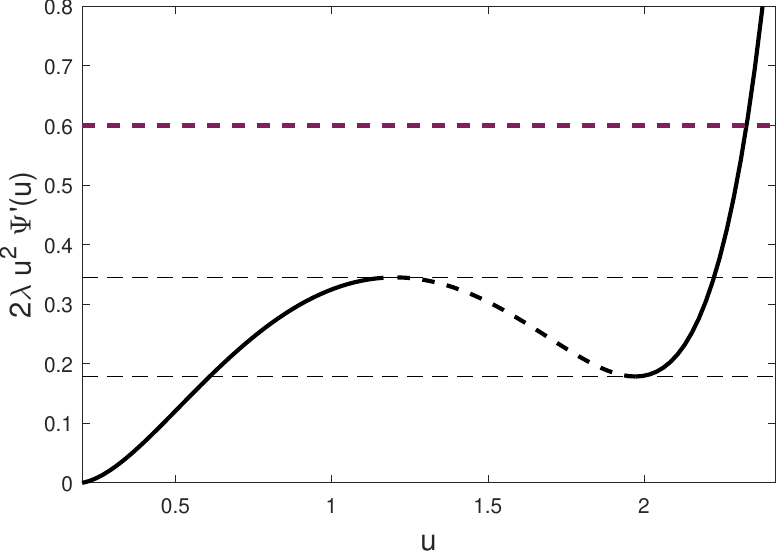}
\hfill
\textbf{b.}\hspace{-0.1cm}
\includegraphics[height=3.4cm, valign=t]{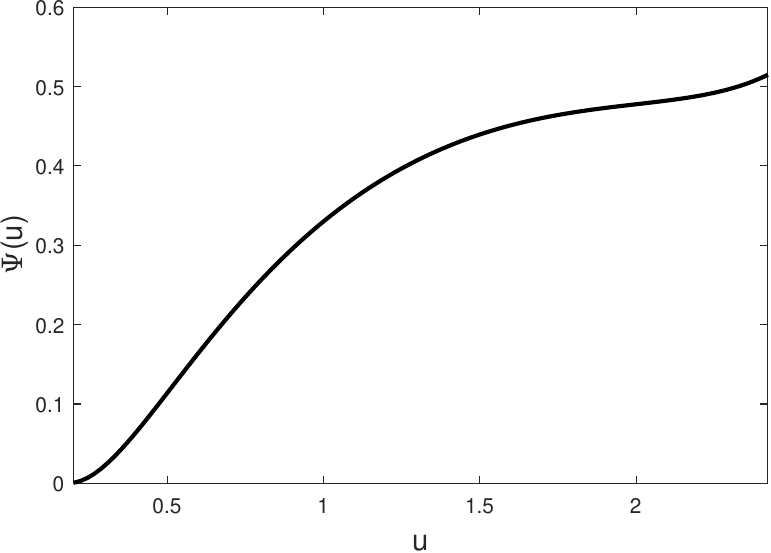}
\hfill 
\textbf{c.}\hspace{-0.1cm}
\includegraphics[height=3.4cm, valign=t]{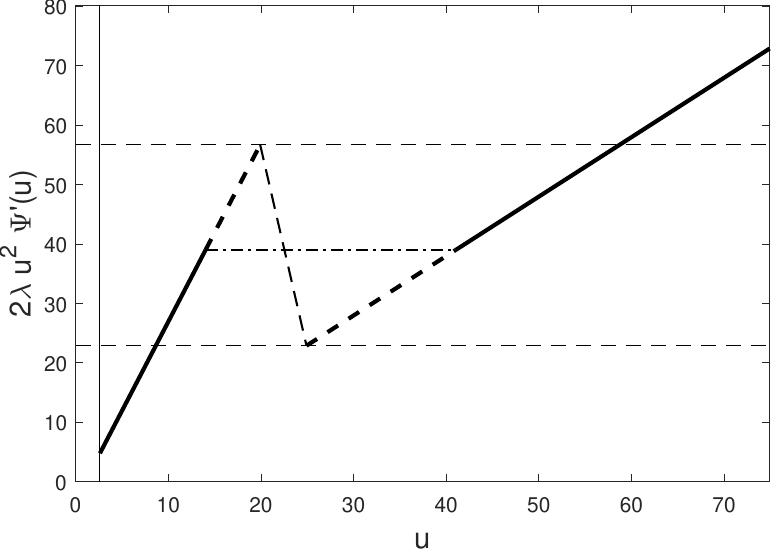}
\hfill
\caption{One-dimensional example illustrating the possibility of a discontinuity in $F_{\text{code}}(x)$. (a). Sketch of a curve $2\lambda u^2\Psi'(u)$ exhibiting a range with multiple solutions. The continuous parts of the curve are the ones leading to minima of the cost. The $v=F_{\text{cat}}$ values lie between zeros and its maximum value indicated by the dashed purple line. (b). The corresponding function $\Psi$, obtained by numerical integration of $\Psi'(u)$. Note the change of curvature in $\Psi$ which leads to the decreasing part of the function in panel (a). (c). A simple example for which one can compute the value at which the solution jumps (dot-dashed line) from one branch (left) to the other (right) as $v=F_{\text{cat}}$ increases.}
\label{fig:multival}
\end{figure}
We have thus the following picture. If the function $u^2\, \Psi'(u)$ is a strictly increasing function of $u$, then for any $x$ there is a unique solution $F_{\text{code}}(x)=G(x)\, u[v]$ for $v= F_{\text{cat}}(x)/G(x)$. For $G(x)=1$ for all $x$, one has then that $F_{\text{code}}(x)$ increases continuously with $F_{\text{cat}}(x)$. Note that, if $G$ is taken as $G(x)=F_{\text{cat}}(x)$, $v=1$ for any $x$, so that $F_{\text{code}}(x)=u[1]\, F_{\text{cat}}(x)$.

If the function $u^2\, \Psi'(u)$ is not monotonic, but is such that, for any $x$, there is at least one solution in a range where $u^2\, \Psi'(u)$ is increasing, then there might be locations $x$ for which several solutions exist. In such a case the one giving the smallest contribution to the cost has to be chosen. This opens the possibility to have constraints for which discontinuities in $F_{\text{code}}(x)$ appear. Such a case is unlikely to occur with typical constraints. However, constraints at the micro level (e.g., on weights) might be equivalent to a more or less complex constraint at the macro level (on the function $F_{\text{code}}$). In any case, it is worth considering the consequences of having a constraint leading to a nonmonotonic function $u^2\, \Psi'(u)$. 

Suppose for instance a behavior as sketched in Fig.~\ref{fig:multival}(a). The x axis gives the possible values $u$ of $F_{\text{code}}$, and the y axis the values $v$ of $F_{\text{cat}}$, which lie between zero and the maximum $v_{\text{max}}=F_{\text{cat}}(x_{\text{cat}})$. The location $x_{\text{cat}}$ is that of the maximum of $F_{\text{cat}}(x)$, typically identical or very close to the category boundary, see Sec. \ref{sec:Fcat}. The function $u^2\, \Psi'(u)$ is plotted in continuous lines for the parts where it is an increasing function of $u$, and with a dashed line for the decreasing part, which do not correspond to a minimum of the cost. As $x$ gets closer to $x_{\text{cat}}$, the value of $F_{\text{cat}}(x)$ increases, going through the range for which there is two solutions. At some location, there must be a jump from the left to the right branch. Hence, there will be an $x$ value at which $F_{\text{code}}$ jumps to a higher value -- from which the variation of $F_{\text{code}}$ is again continuous. In Fig.~\ref{fig:multival}(c), we show a simple example where each branch is a linear segment. In that case, one can find parameters values so that the jump occurs in the middle of the range with multiple solutions (in that example, the data probability distribution is taken uniform on some interval, and the categorical Fisher information is assumed to grow linearly up to its maximum). 

We have not explored the possibility of jumps back and forth between the two branches. The qualitative result that $F_{\text{code}}$ follows $F_{\text{cat}}$ is maintained, except possibly at backward jumps.

\section{Fisher information matrices: From stimulus to feature space}
\label{app:fisher-s-x}
\setcounter{equation}{0}
We have assumed in Sec.~\ref{sec:featurespaces} that the stimuli/data are associated with an underlying feature space $\mathbf{x}^*$, specific to the categories, that is $P(y | \mathbf{s})=P(y | \mathbf{x}^*)$, with typically $K^*=\text{dim}(\mathbf{x}^*) \ll N=\text{dim}(\mathbf{s})$. As discussed Sec. \ref{sec:underlyingman}, in machine learning, the data scientist has only access to the data, not to the underlying feature space. It is thus of interest to consider the links between Fisher information matrices in stimulus/data space, $\mathbf{F}_{\text{cat}}(\mathbf{s})$ and in feature space, $\mathbf{F}_{\text{cat}}(\mathbf{x}^*)$. Furthermore, we have also assumed that through learning a feature space $\mathbf{x} \in \mathbb{R}^K$ is found by the network, for which, in case of efficient learning, 
\beq
P(y | \mathbf{s})=P(y | \mathbf{x}^*) = P(y | \mathbf{x}).
\eeq
In that case, for a neural layer under consideration, one is also interested in the links between $\mathbf{F}_{\text{code}}(\mathbf{s})$ and $\mathbf{F}_{\text{code}}(\mathbf{x}^*)$ or $\mathbf{F}_{\text{code}}(\mathbf{x})$. The analysis below thus corresponds to either $\mathbf{x}^*$, or to $\mathbf{x}$ in case of efficient learning. For simplicity, in the following we omit the $^*$ in order to lighten the notation. 

We first consider the categorical Fisher information. Given that $P(y | \mathbf{s})=P(y | \mathbf{x})$, we have for any component $j$ of the input,
\beq
\frac{\partial \ln P(y | \mathbf{s})}{\partial s_j}= \sum_i \frac{\partial x_i}{\partial s_j}\; \frac{\partial \ln P(y | \mathbf{x})}{\partial x_i},
\eeq
and 
\beq
-\;\frac{\partial^2 \ln P(y | \mathbf{s})}{\partial s_j \partial s_{j'}}=\; -\;\sum_{i,i'} \frac{\partial x_i}{\partial s_j}\,\frac{\partial x_{i'}}{\partial s_{j'}}\; \frac{\partial^2 \ln P(y | \mathbf{x})}{\partial x_i \partial x_{i'}} \;-\; \sum_i \frac{\partial^2 x_i}{\partial s_j \partial s_{j'}}\, \frac{\partial \ln P(y | \mathbf{x})}{\partial x_i}.
\eeq
Multiplying by $P(y | \mathbf{s})=P(y | \mathbf{x})$, and summing over $y$, the second term gives zero, and one gets
\beq
[\mathbf{F}_{\text{cat}}(\mathbf{s})]_{j,j'} = \sum_{i,i'} \frac{\partial x_i}{\partial s_j}\, [\mathbf{F}_{\text{cat}}(\mathbf{x})]_{i,i'}\;\frac{\partial x_{i'}}{\partial s_{j'}},
\eeq
that is
\beq
\mathbf{F}_{\text{cat}}(\mathbf{s}) = \left[\mathbf{J}(\mathbf{s}) \right]^\mathsf{T} \mathbf{F}_{\text{cat}}(\mathbf{x}) \;\mathbf{J}(\mathbf{s}),
\eeq
where $\mathbf{J}(\mathbf{s})$ is the $K\times N$ Jacobian matrix:
\beq
 \left[\mathbf{J}(\mathbf{s}) \right]_{i,j}= \frac{\partial x_i}{\partial s_j}.
 \label{eq:Fcat-s-x}
 \eeq
 Similarly, one has
 \beq
\mathbf{F}_{\text{code}}(\mathbf{s}) = \left[\mathbf{J}(\mathbf{s}) \right]^\mathsf{T} \mathbf{F}_{\text{code}}(\mathbf{x}) \;\mathbf{J}(\mathbf{s}).
\label{eq:Fcode-s-x}
\eeq
Here the Jacobian matrix is not invertible (in contrast with the case of a change of variable, Appendix \ref{app:invariance}). The rank of $\mathbf{F}_{\text{cat}}(\mathbf{s})$ and $\mathbf{F}_{\text{code}}(\mathbf{s})$ are equal to, respectively, those of $\mathbf{F}_{\text{cat}}(\mathbf{x})$ and $\mathbf{F}_{\text{code}}(\mathbf{x})$, which are both at most equal to $K$, the dimension of $\mathbf{x}$ (see Sec. \ref{sec:Fcat} for the rank of $\mathbf{F}_{\text{cat}}$).

What happens to the Cramér-Rao bound? The Fisher information matrix, being a real symmetric matrix, can be diagonalized in an orthogonal basis. If we call $K_{\text{code}}(\mathbf{s})$ the rank of $\mathbf{F}_{\text{code}}(\mathbf{s})$, there is no Cramér-Rao bound for the projection of $\mathbf{s}$ onto the null space of dimension $N-K_{\text{code}}(\mathbf{s})$. For the projection of $\mathbf{s}$ onto the subspace of nonzero eigenvalues, the Cramér-Rao bound applies with the Fisher information matrix restricted to this space of dimension $K_{\text{code}}(\mathbf{s})$. Given the neural activity, only these $K_{\text{code}}(\mathbf{s})$ components of the data can be reconstructed with some quadratic quality measured by the Cramér-Rao bound. Conversely, in an adversarial attack, a perturbation of the data may affect the network output only through its impact onto these components.

\paragraph*{Case $K=1$ with a single coding cell.}
We illustrate the above relations on the simple model of a single cell discussed 
in Appendix \ref{app:singlecell}: 
\beq
y \rightarrow \mathbf{s} \in \mathbb{R}^N \rightarrow x=X(\mathbf{s}) \in \mathbb{R} \rightarrow r \in \mathbb{R}. 
\eeq
The Fisher information matrix associated with the neural activity $r$ with respect to the input $\mathbf{s}$ is here:
\beq
[\mathbf{F}_{\text{code}}(\mathbf{s})]_{j,j'} = \frac{\partial x}{\partial s_j}\, F_{\text{code}}({x})\;\frac{\partial x}{\partial s_{j'}},
\eeq
where $F_{\text{code}}({x})$ is a scalar. Denoting $\grad x$ the $N$-dimensional vector of the derivatives $\frac{\partial x}{\partial s_j}$, one sees that, (i) $\grad x$ is eigenvector of $\mathbf{F}_{\text{code}}(\mathbf{s})$ associated with the unique nonzero eigenvalue, $\lambda_{\text{code}}(\mathbf{s})=(\grad x)^2\,F_{\text{code}}({x})$, and (ii), there are $N-1$ zero eigenvalues, with eigenspace the space orthogonal to $\grad x$.\\
Similarly we have
\beq
[\mathbf{F}_{\text{cat}}(\mathbf{s})]_{j,j'} = \frac{\partial x}{\partial s_j}\, F_{\text{cat}}({x})\;\frac{\partial x}{\partial s_{j'}},
\eeq
and the unique nonzero eigenvalue of $\mathbf{F}_{\text{cat}}(\mathbf{s})$, associated with the eigenvector $\grad x$, is $\lambda_{\text{cat}}(\mathbf{s})=(\grad x)^2\,F_{\text{cat}}({x})$.

\section{Categorical Fisher information: Location of the maxima and Principal Discriminant Curves}
\label{app:maxfcat}
\setcounter{equation}{0}
To supplement Sec. \ref{sec:maxfcat}, the goal of this appendix is to get some insight on how the location of the maximum of $f_{\text{cat}}(\mathbf{x})$ is displaced with respect to the class boundary depending on the differences between the category distributions, and to provide more examples of PDCs.

\subsection{Gaussian distributions with diagonal covariance matrices}
\label{app:diagcov}
We here consider the simplest example with diagonal covariance matrices. We consider equiprobable categories with Gaussian distributions, centered at $\mathbf{c}_{\pm}=\pm \mathbf{c}$, with covariance matrices proportional to the identity matrices:
\beq
\mathbf{\mathbf{\Sigma}}_{-}= \sigma^2 \,\mathbb{I}, \;\;\;\mathbf{\mathbf{\Sigma}}_{+}= a^2 \;\sigma^2 \,\mathbb{I}.
\eeq
Without loss of generality we assume a larger variance for the $+$ category, that is $a\geq 1$. Hence, the maxima of the categorical information are located in the domains where the $+$ category, the one with the largest variance, is the most probable. 

We have 
\beq
L(\mathbf{x}) = \frac{\eta}{2} \left( \,\|\mathbf{x}\|^2 
\,+\, 2\, \rho\, \mathbf{c}.\mathbf{x} 
\,+\, \|\mathbf{c}\|^2 \,-\, K\,\gamma\, \right),
\label{eq:LGaussDiag}
\eeq
and
\beq
\grad L(\mathbf{x} ) \; = \; \eta \; \left( \mathbf{x} \,
+\,\rho \, \mathbf{c} \right),
\eeq
\beq
H=\eta \,\mathbb{I},
\eeq
where
\beq 
\eta 
=\frac{a^2-1}{a^2\, \sigma^2}, \;\;\;\; \eta \geq 0,
\eeq 
\beq 
\rho 
=\frac{a^2 +1}{a^2-1}, \;\;\;\; \rho\geq 1,
\eeq 
and
\beq 
\gamma= \frac{2}{\eta}\; \ln a, \;\;\;\;\gamma\geq 0.
\eeq 
Note that, as $a\rightarrow 1$, $\rho \rightarrow \infty, \eta \rightarrow 0$ and $\gamma \rightarrow \,\sigma^2, \eta\,\rho \rightarrow 2/\sigma^2$. On each axis we take $\|\mathbf{c}\|$ as unit, that is we can set $\|\mathbf{c}\|=1$, and we have two parameters, $\sigma$ and $a$.

The category boundaries are given by $L(\mathbf{x})=0$, that is
\beq
\|\mathbf{x} + \rho\,\mathbf{c}\|^2 \,=\, \rho^2 \,-\, 1\,+\,K \, \gamma.
\eeq
The boundary is a $(K-1)$-sphere (a circle in $2$ dimension) centered at $-\rho\,\mathbf{c}$ and of radius $z_B$,
\beq 
z_B= \sqrt{\rho^2 \,-\, 1\,+\,K \, \gamma}.
\eeq
For $a\rightarrow 1$, the sphere center goes to infinity, the radius diverges, the boundary becomes a hyperplane orthogonal to the line joining the two category centers, crossing at the origin.

The level sets are as well spheres centered at $-\rho \mathbf{c}$, and the PDCs are the rays originating from this center.

Equation (\ref{eq:maxfcat}) for the location of the extrema of the categorical Fisher information can be written
\beq 
\|\mathbf{x} + \rho\,\mathbf{c}\|^2 
\;=\; \frac{2}{\eta}\; \frac{e^{L(\mathbf{x})} + 1}{e^{L(\mathbf{x})} -1}.
\label{eq:maxfcatDiag}
\eeq 
Writing
\beq 
\mathbf{x} + \rho\,\mathbf{c} = z\, \mathbf{u},
\eeq
where $\mathbf{u}$ is an arbitrary unit vector of $R^K$, and $z>0$, we have
\beq 
z^2 \;=\; \frac{2}{\eta}\; \frac{e^{l(z)} + 1}{e^{l(z)} -1}.
\label{eq:zz}
\eeq 
with
\beq 
l(z) \equiv \frac{\eta}{2} \,(z^2-\rho^2+1-K\gamma) = \frac{\eta}{2} \,(z^2-z_B^2).
\label{eq:lz}
\eeq
Thus the location of the maxima of the categorical Fisher information is a $(K-1)$-sphere centered at the same location as the one of the category boundary, with radius $z>z_B$ given by the (unique) solution of the above equation (\ref{eq:zz}). For $a \rightarrow 1$, the two spheres become identical, with $z-z_B \sim (a-1)\, \sigma^4$.

\subsection{Gaussian distributions with non diagonal covariance matrices}
\label{app:eigen_inequ}
For covariance matrices which do not commute, one can at least state the intuitive result that, if the smallest eigenvalue of, say, the covariance matrix $\mathbf{\Sigma}_{+}$, is greater than the largest eigenvalue of the other covariance matrix, $\mathbf{\Sigma}_{-}$, then $H \succeq 0$. Then the maxima of the categorical Fisher information are located in the domain where the $+$ category, the one with the largest variances, is the most probable.

Let us give some details. Let $\mathbf{A}$ and $\mathbf{B}$ be two real symmetric matrices in $K$ dimension with eigenvalues $\{a_i, i=1,..,K\}$ and $\{b_i, i=1,..,K\}$ respectively, listed in decreasing order ($a_1 \geq a_2...$ and $b_1 \geq b_2...$). The sum $\mathbf{C}=\mathbf{A}+\mathbf{B}$ is as well a real symmetric matrix, and we denote by $\{c_i, i=1,..,K\}$ its eigenvalues (also listed in decreasing order). 

Let $\mathbf{u}$ be eigenvector of $\mathbf{C}$ with unit norm ($\mathbf{u}^2=1$) for eigenvalue $c$, that is $\mathbf{C}.\mathbf{u} = c\,\mathbf{u}$. There exists orthogonal transformations $\mathbf{P}$ and $\mathbf{Q}$ which diagonalize $\mathbf{A}$ and $\mathbf{B}$, respectively, so that $\mathbf{A}=\mathbf{P}^\mathsf{T} (\text{diag a})\,\mathbf{P}$ and $\mathbf{B}=\mathbf{Q}^\mathsf{T}(\text{diag b})\, \mathbf{Q}$. Then $\mathbf{u}^\mathsf{T}\mathbf{C}\mathbf{u}= c$ and $\mathbf{u}^\mathsf{T}\mathbf{C}\mathbf{u}= \textstyle{\sum}_k a_k [(\mathbf{P}\mathbf{u})_k]^2 + \textstyle{\sum}_k b_k [(\mathbf{Q}\mathbf{u})_k]^2$. Since for any $k$, $a_k\geq a_K$ and $b_k \geq b_K$, we have $c\geq a_K [\mathbf{P}\mathbf{u}]^2+ b_K [\mathbf{Q}\mathbf{u}]^2 = a_K+b_K$. Hence $c_K=\min_k c_k\geq a_K + b_K$.

This inequality can also easily be derived from known inequalities for the eigenvalues of the sum (here difference) of real symmetric matrices (see, e.g., Ref. \cite{Fulton_1998}). The inequality of interest here is:
\beq
\textstyle{\sum}_{i=1}^{K-1} c_i \leq \textstyle{\sum}_{i=1}^{K-1} a_i + \textstyle{\sum}_{i=1}^{K-1} b_i.
\label{eq:ineq_sum}
\eeq
Taking the trace of the sum $\mathbf{C}$ we have $\textstyle{\sum}_{i=1}^K c_i=\textstyle{\sum}_{i=1}^K a_i + \textstyle{\sum}_{i=1}^K b_i$, and making use of the above inequality (\ref{eq:ineq_sum}) we get
\beq
c_K \geq a_K + b_K.
\label{eq:eigen_inequ}
\eeq

We apply this inequality (\ref{eq:eigen_inequ}) to $\mathbf{A}=\mathbf{\Sigma}_{-}^{-1}, \mathbf{B}= -\mathbf{\Sigma}_{+}^{-1}$ (so that $\mathbf{C}$ is the Hessian $\mathbf{H}$), given the eigenvalues of $\mathbf{\Sigma}_{-}$, $\{(\sigma_i^{-})^2, i=1,..,K\}$, and of $\mathbf{\Sigma}_{+}$, $\{(\sigma_i^{+})^2, i=1,..,K\}$, again listed in decreasing order. These eigenvalues are the variance along the principal axis of the category distributions. We have $a_K=(1/\sigma_1^{-})^2, b_K=(1/\sigma_K^{+})^2$, and thus
\beq 
c_K \geq \frac{1}{(\sigma_1^{-})^2} - \frac{1}{(\sigma_K^{+})^2},
\eeq
which is positive if
\beq 
\sigma_K^{+} \geq \sigma_1^{-}.
\eeq
This is a sufficient (but not necessary) condition for having $\mathbf{H} \succeq 0$.

\subsection{Location of the maxima for similar data distributions}
\label{sec:maxfactapprox}
One can see that the location of the maxima of $f_{\text{cat}}$ is close to the class boundary for covariance matrices not too different. Indeed, in such case the Hessian is small. We can expand Eq.~(\ref{eq:maxfcat}) for $\mathbf{x}$ at the vicinity of the class boundary $\mathbf{x}_b$ on a PDC. Since $L(\mathbf{x}_b )=0$, at first order the distance along the PDC from the class boundary of the maximum of the categorical Fisher information is given by
\beq
\mathbf{v}(\mathbf{x}_b) . (\mathbf{x}-\mathbf{x}_b) = 4\, \frac{\mathbf{v}(\mathbf{x}_b)^\mathsf{T}\,\mathbf{H} \,\mathbf{v}(\mathbf{x}_b) }{ \|\grad L(\mathbf{x}_b)\|^3},
\label{eq:approx}
\eeq
where $\mathbf{v}(\mathbf{x}_b) \equiv \grad L(\mathbf{x}_b)/ \|\grad L(\mathbf{x}_b)\|$ is the unit vector orthogonal to the class boundary (tangent to the PDC) at $\mathbf{x}_b$. In addition, for sharply peaked distributions, we expect $\|\grad L\|$ to be large at the class boundary, so that the distance from the class boundary of the maximum of the categorical Fisher information is also of order $1/\|\grad L(\mathbf{x}_b)\|^3$. More generally, that is for non Gaussian cases but for distributions sufficiently smooth with similar shapes, and essentially differing by their centers, the Hessian will be small and we expect qualitatively similar results.

\subsection{Numerical illustrations: 1D case}
We illustrate the above results, Sec. \ref{app:diagcov}, in the 1D (hence scalar) case. The Gaussian distributions are centered at $c_{\pm}=\pm c$, and $c=1$, with standard deviations $\sigma_{-}=\sigma, \sigma_{+}=a \sigma$. This is essentially equivalent to considering, in $K$ dimensions, properties along the axis joining the two centers, except for the factor $K$ which is here equal to $1$. The zero-sphere consists in two points on this axis, located at $x_b^{\pm}$, 
\beq 
x_b^{\pm} = - \rho\,\pm\sqrt{\rho^2 -1 + \gamma }.
\eeq
There are indeed two boundaries. One, $x_b^{+}$, is in between the two centers for $a$ not too large. The other one is at a value $x_b^{-}$ more negative than the center of the $-$ category: at large negative values of $x$, the $+$ category becomes the most probable. This boundary is in general not relevant, concerning very rare events. However, if $a$ is large and $\sigma$ small enough, the $-$ category appears as lying within the $+$ category, and both boundaries are relevant. In any case, we essentially focus on the boundary $x_b^{+}$ which corresponds to the meaningful boundary in real applications.

The categorical Fisher information,
\beq 
f_{\text{cat}}(x)= \frac{1}{1+e^{L(x)}} \frac{1}{1+e^{- L(x)}} \; \left(\frac{dL(x)}{dx}\right)^2,
\eeq
where $dL/dx=\eta(x+\rho)$, has two maxima, located at $x_{\text{cat}}^{\pm} =-\rho \pm z$, $z>0$, with $z$ being a solution of (\ref{eq:zz}), (\ref{eq:lz}) (with $K=1$).

\begin{figure}
\centering
\textbf{a.}\hspace{-0.1cm}
\includegraphics[height=4.9cm, valign=t]{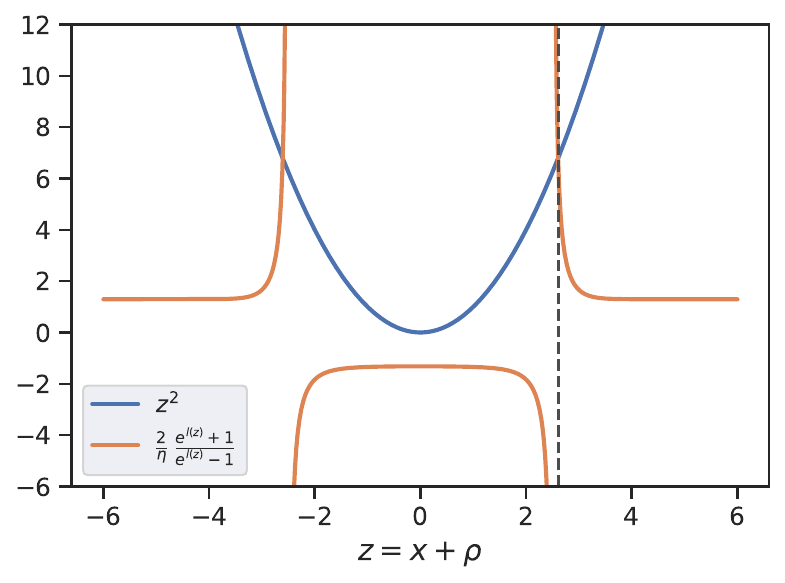}
\hfill
\textbf{b.}\hspace{-0.1cm}
\includegraphics[height=4.9cm, valign=t]{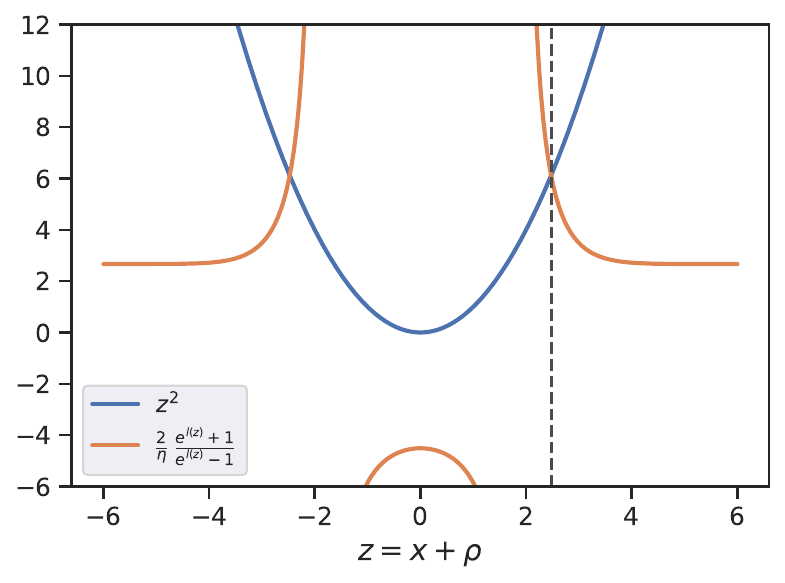}
\hfill
\caption{\textbf{One-dimensional example with two Gaussian categories: Visualization of Eq.~(\ref{eq:zz}) giving the location of the maxima of $f_{\text{cat}}$}. (a). $a=1.5$, $\sigma=0.6$, corresponding to the case shown in Fig.~\ref{fig:gaussian1d_xb_xcat_example} (top panels). (b). $a=2.0$, $\sigma=1.0$, corresponding to the case shown in Fig.~\ref{fig:gaussian1d_xb_xcat_example} (bottom panels).}
\label{fig:gaussian1d_xcat}
\end{figure}

In Fig.~\ref{fig:gaussian1d_xcat} we illustrate the numerical solution of Eq.~(\ref{eq:zz}) for the parameter values corresponding to the results we present in the main text, Fig.~\ref{fig:gaussian1d_xb_xcat_example}. 

\subsection{Numerical illustrations: Principal discrimination curves in 2D} 
\label{app:eigen_inequ_details}

\begin{figure}[ht]
\centering
\textbf{a.}
\includegraphics[height=4.9cm, valign=t]{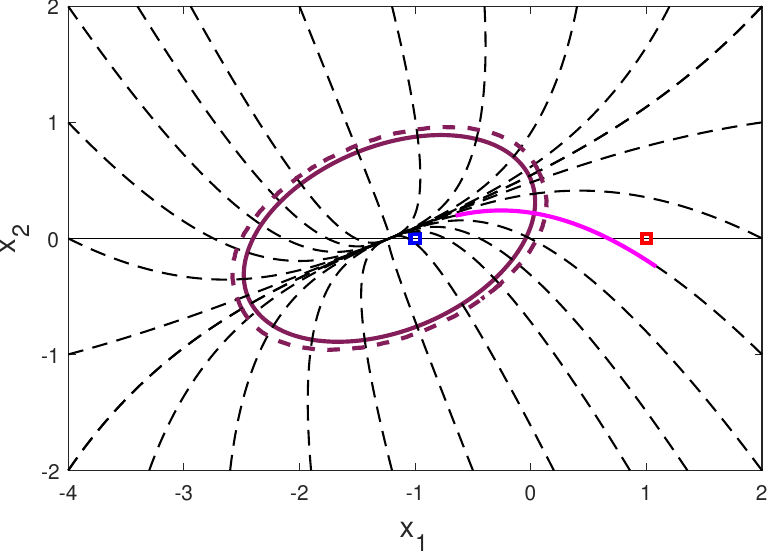}%
\hfill
\textbf{b.}\hspace{0.1cm}
\includegraphics[height=4.9cm, valign=t]{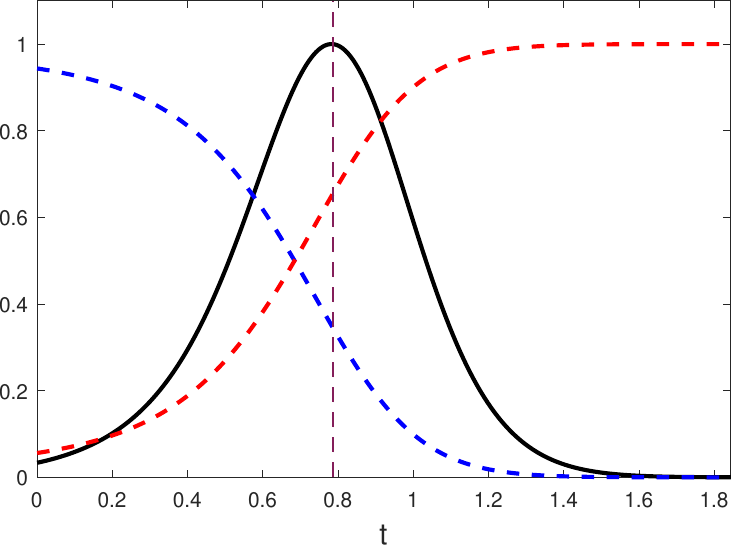}\\ 
\textbf{c.}
\includegraphics[height=4.9cm, valign=t]{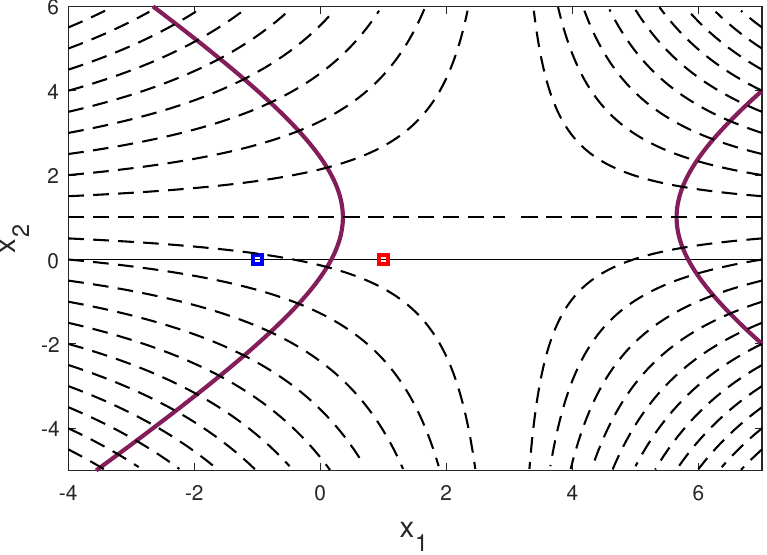} 
\hfill
\textbf{d.}
\includegraphics[height=4.9cm, valign=t]{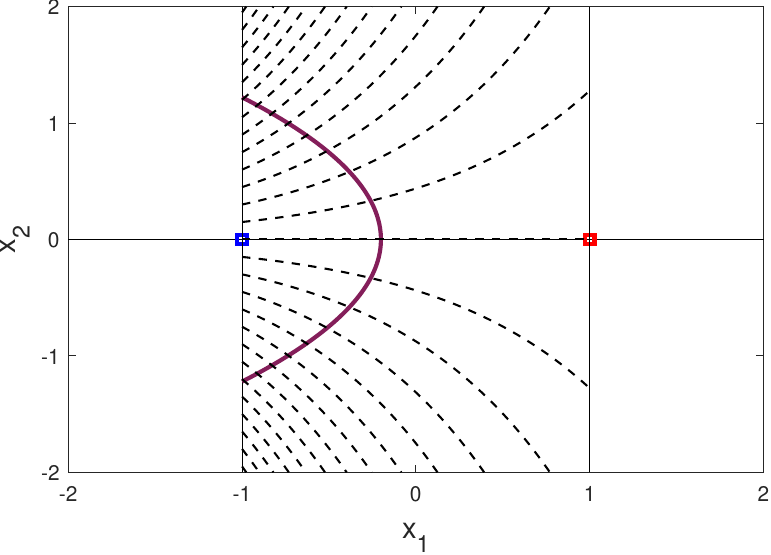} 
\caption{\textbf{Two-dimensional examples with two categories: Category boundary and Principal discriminant curves.} For each one of the panels (a), (c), (d), the abscissa is chosen as the line going through the category centers. The center of the $-$ category is on the left (blue square), the one of the $+$ category on the right (red square). The category boundary is the continuous purple thick line. A sample of PDCs is plotted with thin dashed lines. (a), (c). 2D Gaussian categories with different covariance matrices. 
(a). Elliptic boundary. The location of the maxima of the categorical information, in dashed purple thick line, is very close to the boundary. 
Along the segment of PDC in thick magenta, we plot in panel (b) the categorical Fisher information eigenvalue (divided by its maximum value), together with the posterior probabilities of each category, in blue and red. The purple dashed vertical line gives the location of the maximum of the categorical information. The abscissa for this panel is the curvilinear abscissa along the segment, with origin taken at the beginning of the segment inside the ellipse. 
(c). Hyperbolic boundary. The right branch of the class boundary is in fact not relevant (the density of data, not shown, is extremely small in this part of the plane). 
(d). An example with non Gaussian categories. The domain is bounded on the $x_1$ axis. For each category, independent $x_1$ and $x_2$ distributions, with for $x_1$ an exponential decrease from the domain boundary toward the inside of the domain, and for $x_2$ Gaussian distributions.}
\label{fig:ppdcurves}
\end{figure}

Here we focus on the principal discriminant curves (PDC), considering in Fig.~\ref{fig:ppdcurves} different 2-dimensional cases. We show two Gaussian cases with different covariance matrices for the two categories, and also one example with non-Gaussian categories. In the case of Fig.~\ref{fig:ppdcurves}(a), we also show the location of the maxima of the categorical Fisher information, as we did for the case presented in the main text, Fig.~\ref{fig:gaussian2d_pdc}.

Fig.~\ref{fig:ppdcurves}(a): 2D Gaussian categories with covariance matrices that can be diagonalized in a same basis: 
\beq
\mathbf{\Sigma}_{-} =\begin{pmatrix}
 .2 & .05\\
 .05 & .1
\end{pmatrix}, \; \mathbf{\Sigma}_{+} = 10 \; \mathbf{\Sigma}_{-}.
\eeq
This case is analogous to the circular one shown in Fig.~\ref{fig:gaussian2d_pdc}, but with an elliptic boundary. The category with smallest variances is an island within the sea of the other category. All PDCs end at a same point within the ellipse. The maxima of the categorical information lies on an ellipse slightly larger than the one of the boundary. The difference between the two covariance matrices is chosen large enough so that one can distinguish the two ellipses.\\
Figure~\ref{fig:ppdcurves}(c): 2D Gaussian categories with covariance matrices which do not commute:
\beq
\mathbf{\Sigma}_{-} =\begin{pmatrix}
 .4 & .1\\
 .1 & .2
\end{pmatrix}, \; \mathbf{\Sigma}_{+} =\begin{pmatrix}
 .2 & .1\\
 .1 & .4
\end{pmatrix}.
\eeq
This parameter choice leads to hyperbolic boundaries.

With Fig.~\ref{fig:ppdcurves}(d) we extend the numerical illustration to non Gaussian categories. In this example, the domain is bounded along the $x_1$ axis. For each category, the distributions of $x_1$ and $x_2$ are independent. For $x_1$, we consider an exponential decrease from the domain boundary toward the inside of the domain: for $x_1 \in [-1, 1]$, 
\beq
P(x_1|\pm)= \frac{1}{Z_{\pm}}\; \exp{-|x_1-c_{\pm}|/\tau_{\pm} }, 
\eeq
where $Z_{\pm}$ is the normalization constant, $Z_{\pm}= 1- \exp{-2c/\tau_{\pm}}$, with $c_{\pm}=\pm c, \;c=1$, $\tau_{-}=.2, \tau_{+}=.5$. For $x_2$, we consider Gaussian distributions with different variances: 
\beq
P(x_2|\pm)=\frac{1}{\sqrt{2\pi \sigma_{\pm}^2}}\;\exp{-\frac{x_2^2}{2\sigma_{\pm}^2}},
\eeq
with $\sigma_{-}^2=.1, \sigma_{+}^2=.4$.

\clearpage
\section{Categorical and neural Fisher information matrices: 2D illustration}
\label{app:2d_fcat_fcod}
\begin{figure}[h]
\centering
\includegraphics[width=0.85\linewidth, valign=t]{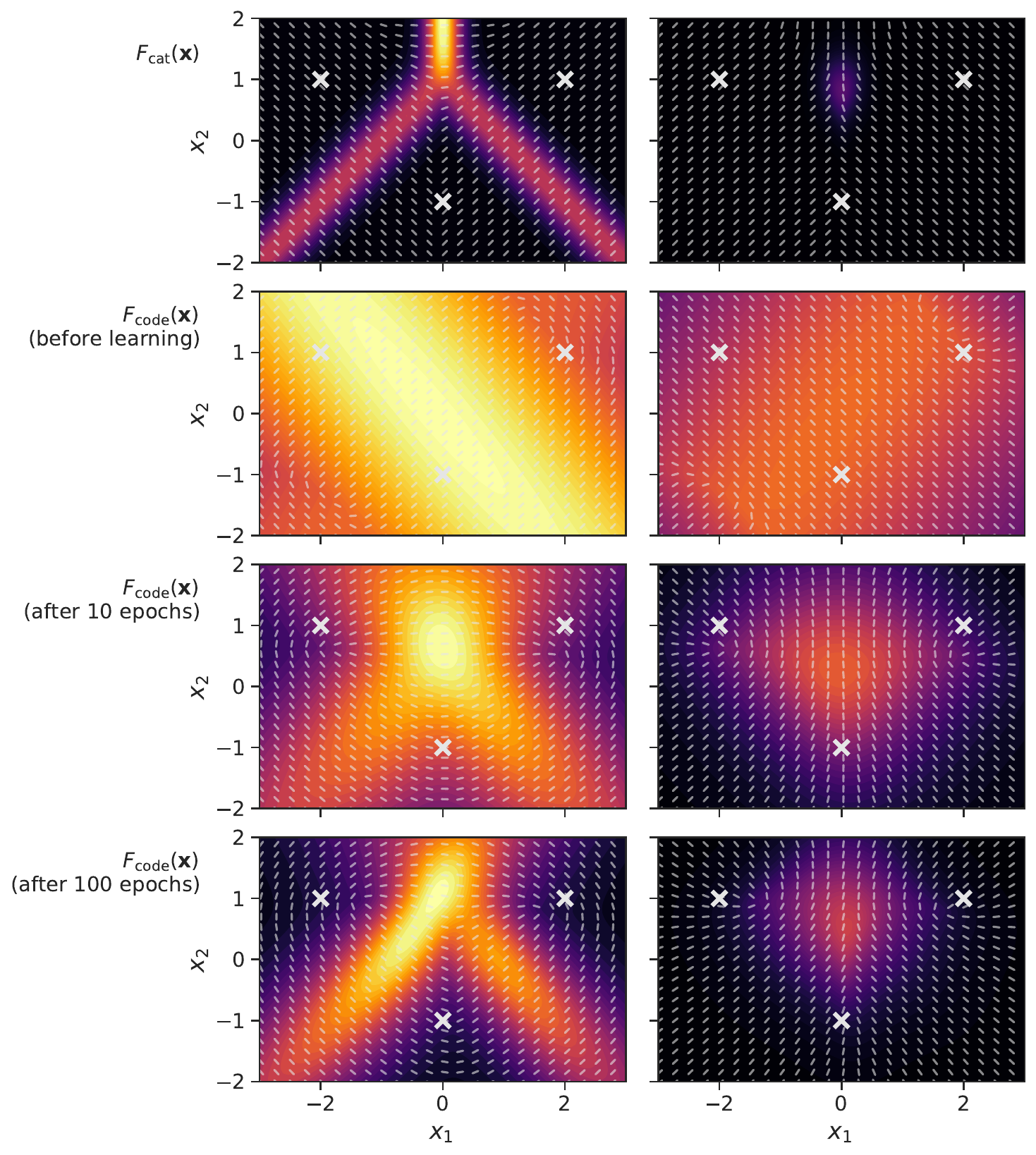}\\
\caption{\textbf{Two-dimensional example with three Gaussian categories: categorical and neural Fisher information quantities}. 
The small line represents the direction at each point on the $(x_1, x_2)$ plane of the eigenvector of the Fisher information matrices associated with: (left) the largest eigenvalue, (right) the smallest eigenvalue. 
The top row represents the categorical Fisher information matrix $\mathbf{F}_{\text{cat}}(\mathbf{x})$. The three following rows represent the neural Fisher information matrix $\mathbf{F}_{\text{code}}(\mathbf{x})$ at various stages of training, namely before training, after 10 epochs, and after 100 epochs. In each plot, the magnitude of the considered eigenvalue is represented by the color, the lighter the greater. The magnitude values are normalized for each row, so as to compare the respective magnitudes of the largest and smallest eigenvalues at each point.
}
\label{fig:gaussian2d_si}
\end{figure}

\setcounter{equation}{0}
For the numerical example with three categories in two dimensions discussed Sec. \ref{sec:2d_gauss}, as a supplement to Fig.~\ref{fig:gaussian2d}, we compare here the categorical and neural Fisher information matrices during learning. To do so, we provide in Fig~\ref{fig:gaussian2d_si} a full visualization of these Fisher information matrices in the $(x_1, x_2)$ plane. At each point in the plane, we look at both the largest and the smallest eigenvalues, and at the associated eigenvectors. Note that the top- and bottom-left panels, corresponding to the largest eigenvalues, are the same as, respectively, Figs.~\ref{fig:gaussian2d}(b) and \ref{fig:gaussian2d}(d).

\clearpage
\section{Additional numerical experiments with MNIST}
\label{app:additionalMNIST}
\setcounter{equation}{0}

\begin{figure}[!hb]
\centering
\textbf{a.}\hspace{-0.1cm}
\includegraphics[width=0.47\linewidth,valign=t]{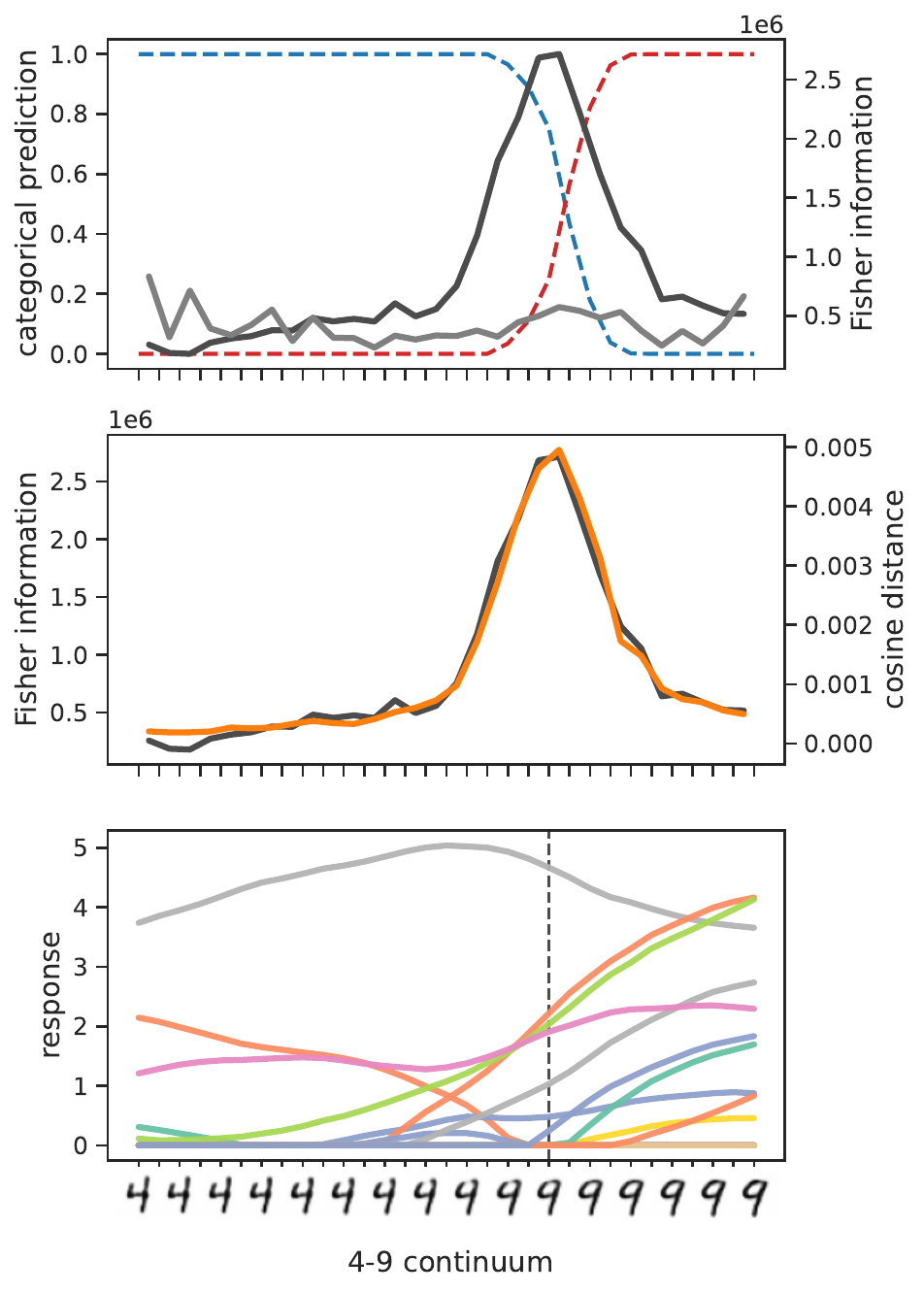}
\hfill
\textbf{b.}\hspace{-0.1cm}
\includegraphics[width=0.47\linewidth,valign=t]{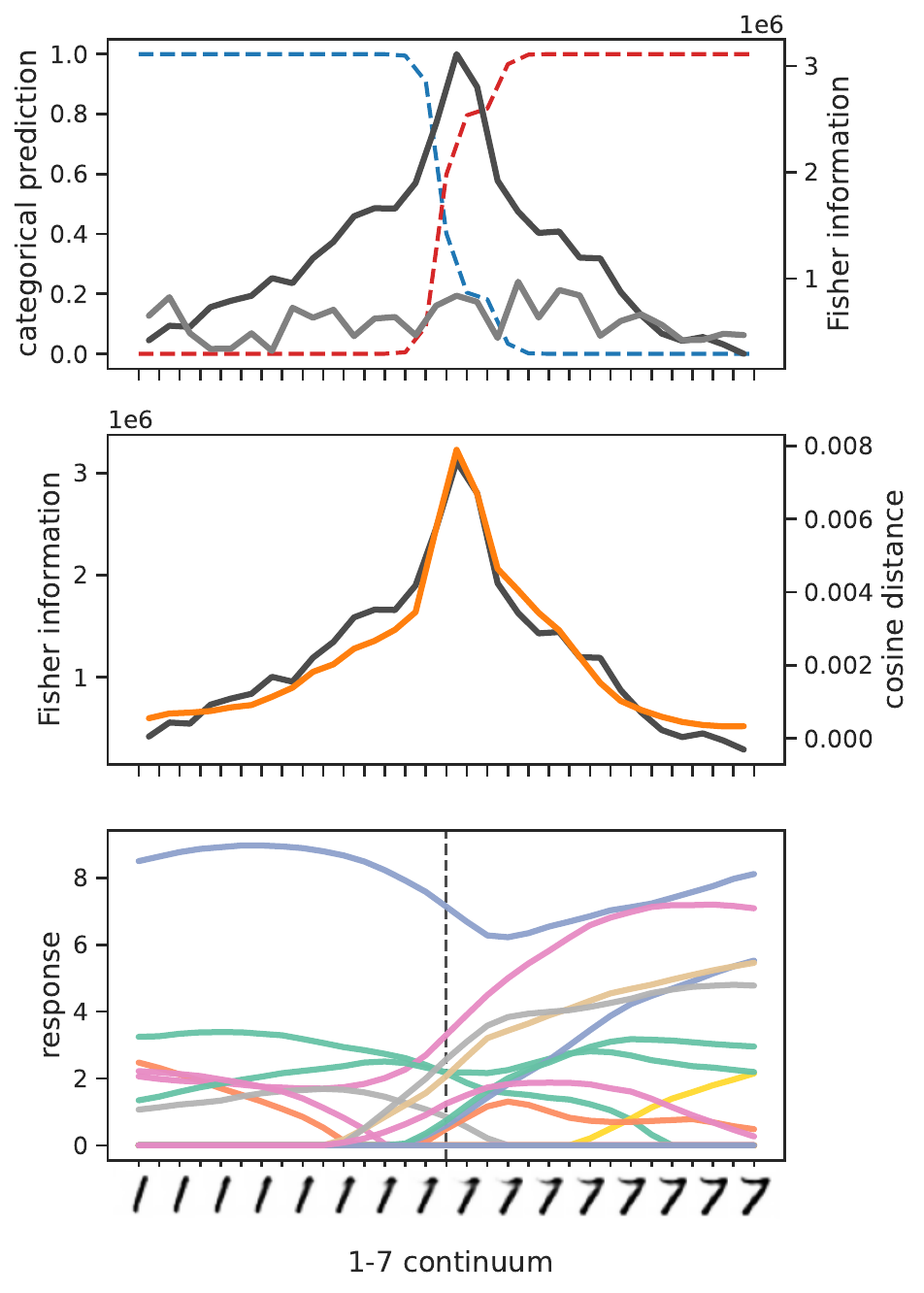}
\caption{\textbf{Categorical perception along a 4-9 continuum (left) and  a 1-7 continuum (right)}. The neural network is the exact same network as in Fig.~\ref{fig:mnist}.
(top) Scalar neural Fisher information $F_{\text{code}}$ along the continua (averaged over the same 10 training runs as in Fig.~\ref{fig:mnist}), before (light gray) and after (dark gray) learning. The dashed colored lines indicate the posterior probabilities, as found by the network, blue corresponding to category on the left and red to category on the right. 
(middle) Comparison between Fisher information (dark gray, left y axis) and cosine distance (orange, right y axis) between neural activities evoked by contiguous items along each continuum.
(bottom) Tuning curves of the 20 first neurons of the last hidden layer of one of the model trained. The vertical dotted lines locate the corresponding maximum of the neural Fisher information. The top and middle sub-panels of panel (a) are identical to panels (a) and (b) of Fig.~\ref{fig:mnist}. 
 }
\label{fig:mnist_extra}
\end{figure}

\begin{figure}[tb]
	\centering
	\textbf{a.}\hspace{-0.1cm}
	\includegraphics[width=0.47\linewidth,valign=t]{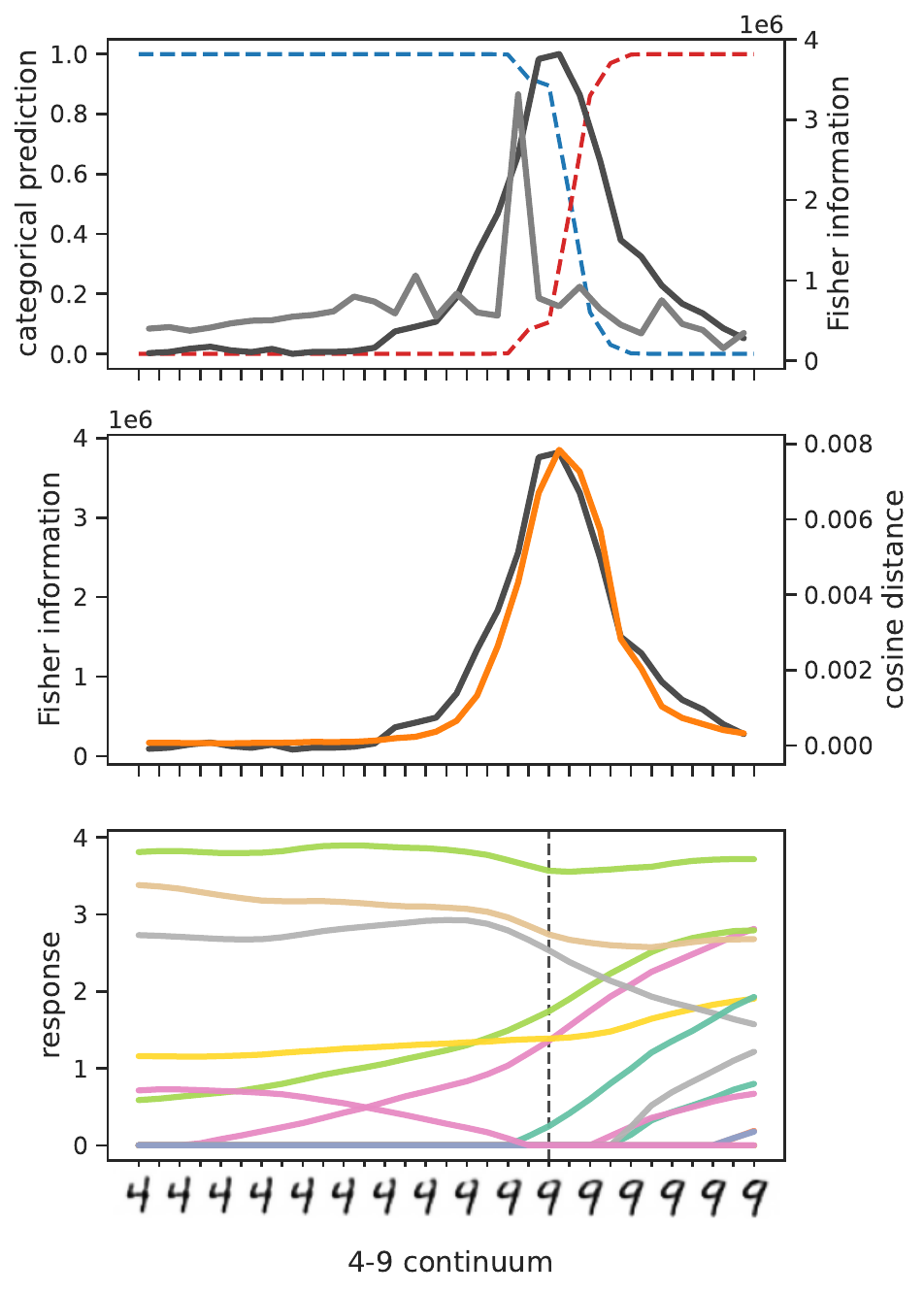}
	\hfill
	\textbf{b.}\hspace{-0.1cm}
	\includegraphics[width=0.47\linewidth,valign=t]{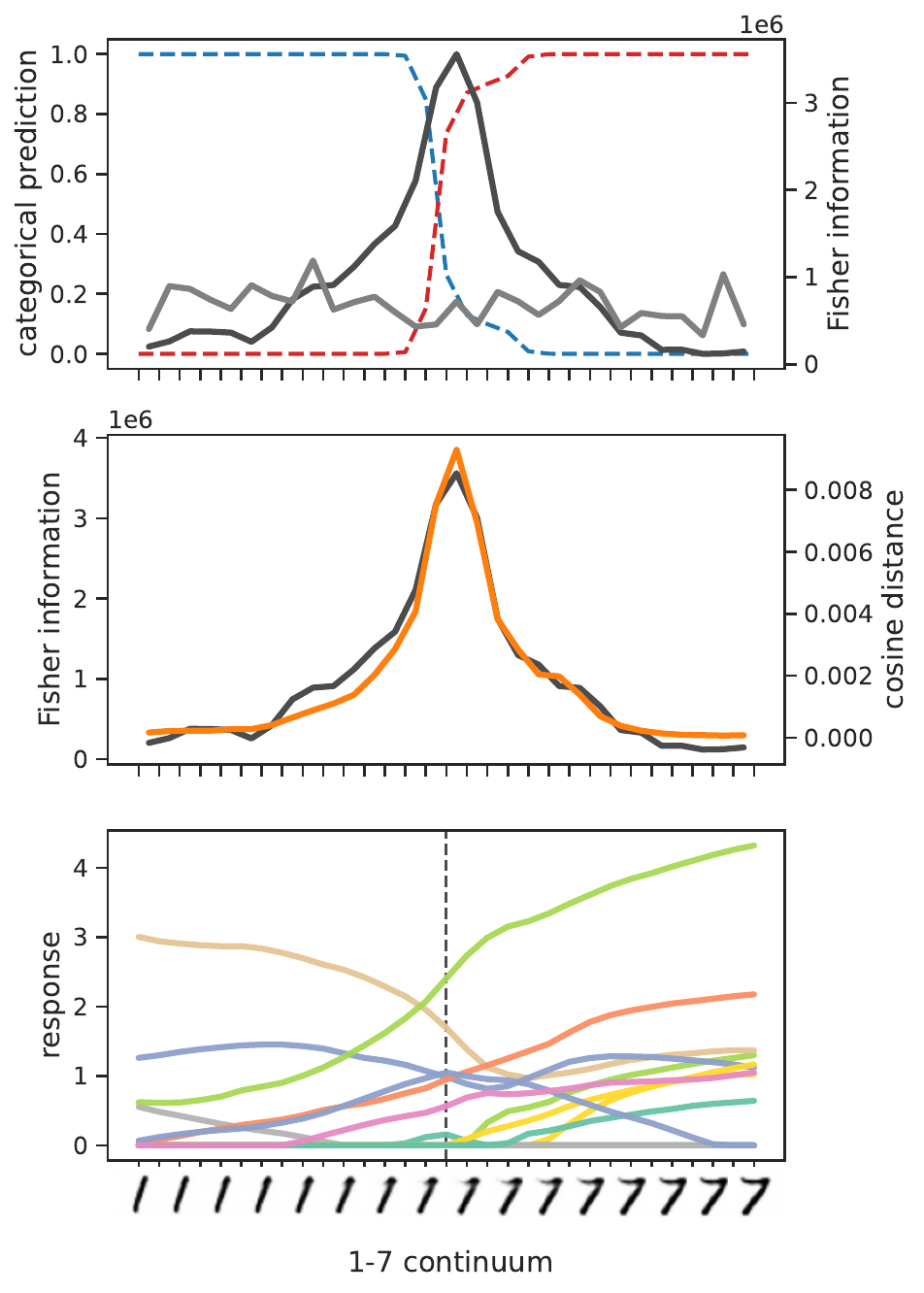}
	\caption{\textbf{Additional experiments with MNIST: Deeper network}. Same as in Fig.~\ref{fig:mnist_extra} but considering a multilayer network made of four hidden layers.
     }
	\label{fig:mnist_extra_deeper}
\end{figure}

In this appendix, we provide additional results with the MNIST database. In Fig.~\ref{fig:mnist} we considered a 4 to 9 continuum. Here we also present the results for a 1 to 7 continuum, in Fig.~\ref{fig:mnist_extra} for the very same neural network as in Fig.~\ref{fig:mnist} 
(averaging over of the same 10 training runs on the full MNIST database), and in Fig.~\ref{fig:mnist_extra_deeper} for a deeper network. 

In addition we plot the tuning curves of a set of neurons in the last hidden layer of one of the model trained, as observed in response to contiguous items along each continuum. See the discussion in the main text, end of Sec.~\ref{sec:mnist}.

\clearpage
\addcontentsline{toc}{section}{References} 
\section*{} 
\bibliographystyle{apalike}
\bibliography{refs} 

\end{document}